\documentclass{article}
\usepackage[final]{gpsmdms_2022}
\usepackage[utf8]{inputenc}
\usepackage{xcolor}
\usepackage{siunitx}
\usepackage{tabularx}
\usepackage{booktabs}
\usepackage{graphicx}
\usepackage[nolist]{acronym}
\usepackage{hyperref}
\usepackage{appendix}
\usepackage[authoryear, square]{natbib}
\usepackage[position=b]{subcaption}
\usepackage{grffile}
\usepackage{authblk}
\usepackage{floatrow}
\newfloatcommand{capbtabbox}{table}[][\FBwidth]
\usepackage{url}
\usepackage{todonotes}
\setuptodonotes{inline}
\usepackage[inline]{enumitem}
\usepackage{amsmath}
\usepackage[capitalise,noabbrev]{cleveref}

\title{PI is back! Switching Acquisition Functions\\ in Bayesian Optimization}
\author[1]{Carolin Benjamins}
\author[2,3]{Elena Raponi}
\author[2]{Anja Jankovic}
\author[2]{Koen van der Blom}
\author[4]{Maria Laura Santoni}
\author[1]{Marius Lindauer}
\author[2]{Carola Doerr}
\affil[1]{\small{Institute of AI, Leibniz University Hannover, Germany}}
\affil[2]{\small{Sorbonne Universit\'e, CNRS, LIP6, Paris, France}}
\affil[3]{\small{TU München, Germany}}
\affil[4]{\small{University of Camerino, Italy}}

\begin{document}

\begin{acronym}
\acro{DoE}[DoE]{\emph{Design of Experiment}}
\acro{BO}{Bayesian Optimization}
\acro{AF}{acquisition function}
\acro{EI}{Expected Improvement}
\acro{PI}{Probability of Improvement}
\acro{UCB}{Upper Confidence Bound}
\acro{ELA}{Exploratory Landscape Analysis}
\acro{GP}{Gaussian Process}
\acro{TTEI}{Top-Two Expected Improvement}
\acro{TS}{Thompson Sampling}
\acro{DAC}{Dynamic Algorithm Configuration}
\acro{AC}{Algorithm Configuration}
\acro{CMA-ES}{CMA-ES}
\end{acronym}

\maketitle

\begin{abstract}
Bayesian Optimization (BO) is a powerful, sample-efficient technique to optimize expensive-to-evaluate functions. 
Each of the BO components, such as the surrogate model, the acquisition function (AF), or the initial design, is subject to a wide range of design choices.
Selecting the right components for a given optimization task is a challenging task, which can have significant impact on the quality of the obtained results.\\
In this work, we initiate the analysis of which AF to favor for which optimization scenarios. To this end, we benchmark SMAC3 using Expected Improvement (EI) and Probability of Improvement (PI) as acquisition functions on the 24 BBOB functions of the COCO environment.
We compare their results with those of schedules switching between AFs.
One schedule aims to use EI's explorative behavior in the early optimization steps, and then switches to PI for a better exploitation in the final steps.
We also compare this to a random schedule and round-robin selection of EI and PI. \\
We observe that dynamic schedules oftentimes outperform any single static one.
Our results suggest that a schedule that allocates the first $\SI{25}{\percent}$ of the optimization budget to EI and the last $\SI{75}{\percent}$ to PI is a reliable default. However, we also observe considerable performance differences for the 24 functions, suggesting that a per-instance allocation, possibly learned on the fly, could offer significant improvement over the state-of-the-art BO designs.
\end{abstract}

\section{Introduction}


\acf{BO} \citep{mockus_bayesian_2012} is a technique to find the global optimum of black-box problems when each evaluation of the objective function is expensive or otherwise resource-intensive, e.g., problems requiring elaborated physical experiments or computationally-heavy simulations. 
\ac{BO} is widely known as one of the most efficient black-box approaches in terms of number of function evaluations required, hence a promising tool for the optimization of low-budget problems. 

\ac{BO} is defined as a modular framework which consists of \begin{enumerate*}[label=(\roman*)]
\item defining an initial set of points where the true (unknown, expensive) objective function is evaluated,
\item approximating the objective function based on the initial design by placing a prior probability distribution on it (i.e., building a \emph{surrogate model}),
\item selecting where to sample next by optimizing an \emph{acquisition function} (also known as \emph{infill criterion}) that takes care of the trade-off between exploration and exploitation, 
\item evaluating new samples on the true objective function, and finally
\item computing the posterior probability distribution on the true objective function based on new knowledge, hence updating the surrogate model.
\end{enumerate*}
Steps (iii) to (v) are repeated iteratively until the evaluation budget is exhausted.

\begin{figure}[h]
\begin{floatrow}
\ffigbox{%
  \includegraphics[width=0.48\textwidth]{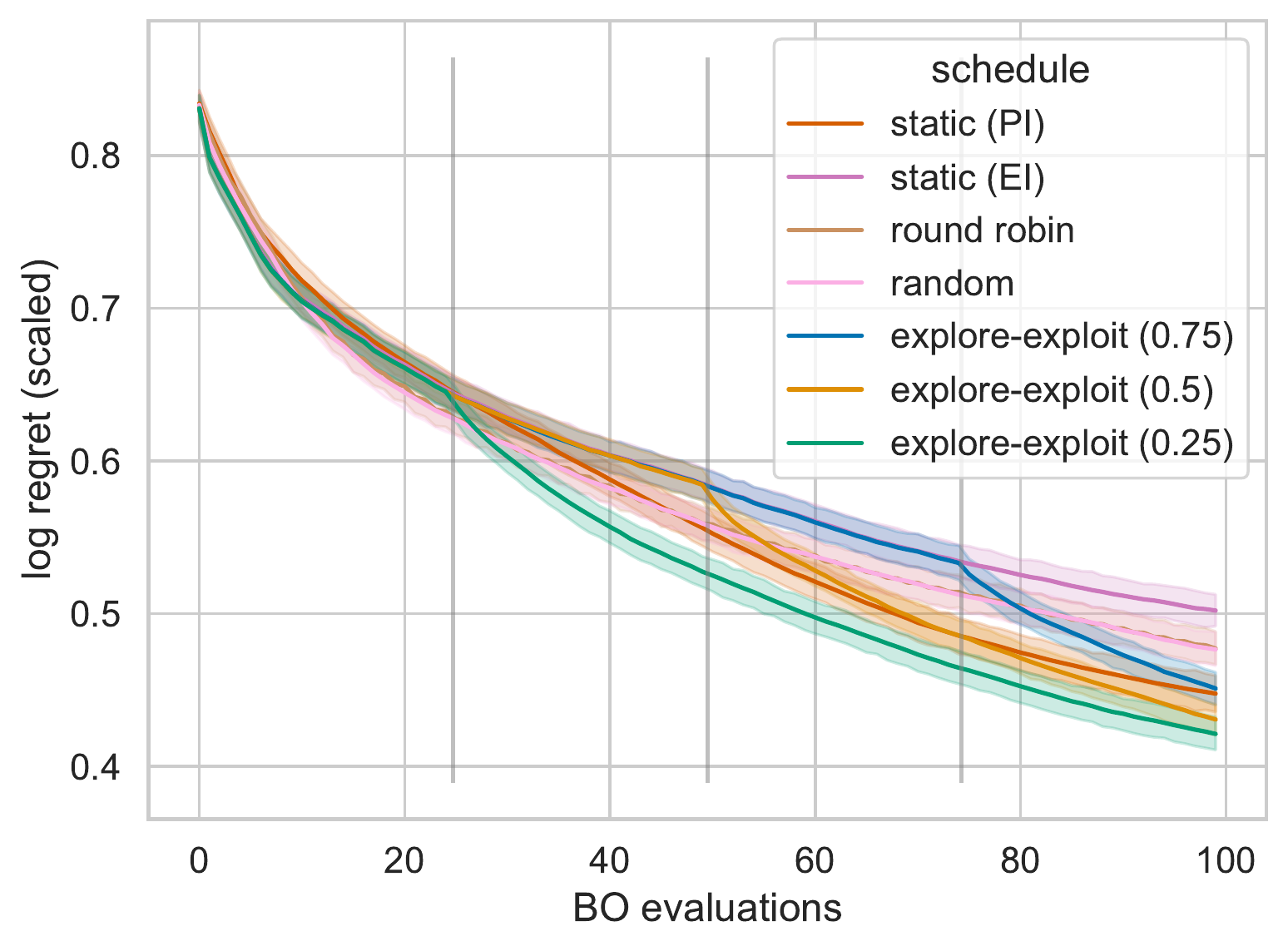}
}{%
  \caption{Switching \ac{AF}s is beneficial. Scaled log-regret averaged over all \num{24} BBOB functions for \num{5} dimensions with the \SI{95}{\percent} confidence interval for the schedules from~\cref{tab:schedules}.}
    \label{fig:convergence_all_ci}
} \quad
\capbtabbox{%
\resizebox{0.48\textwidth}{!}{
    \begin{tabular}{ll}
        \toprule
         Name & Schedule \\
         \midrule
         static (EI) & \includegraphics[width=3cm]{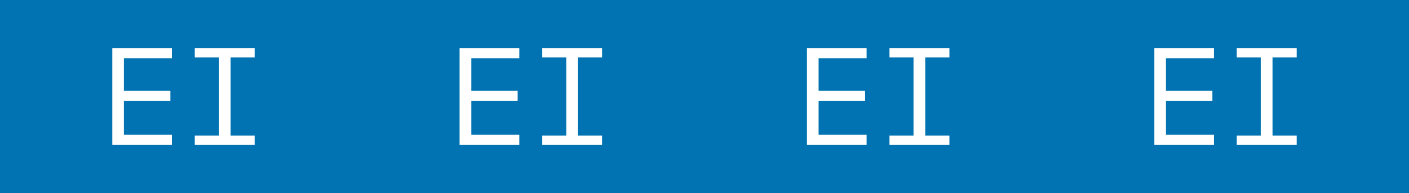}\\
         static (PI) & \includegraphics[width=3cm]{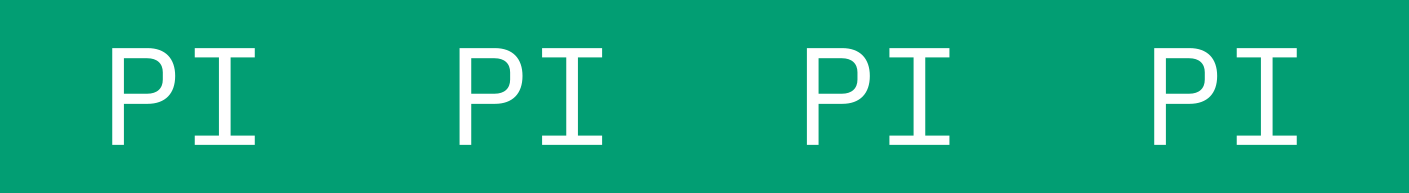}\\
         random & \includegraphics[width=3cm]{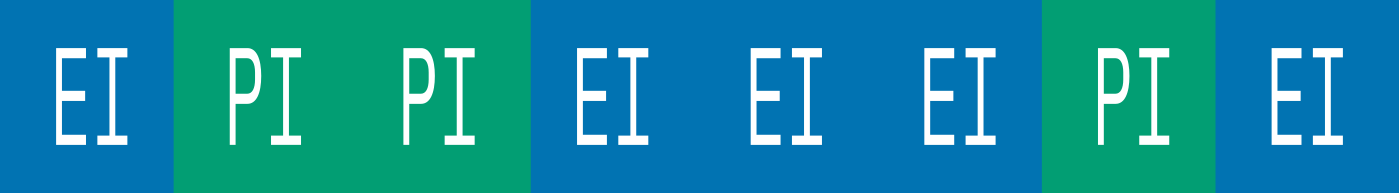}\\
         round robin & \includegraphics[width=3cm]{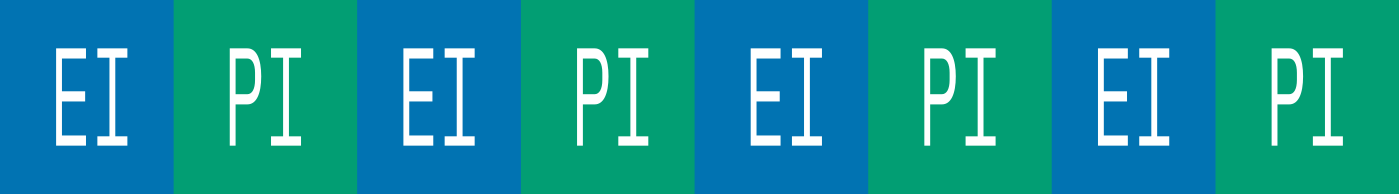}\\
         explore-exploit (\num{0.25}) & \includegraphics[width=3cm]{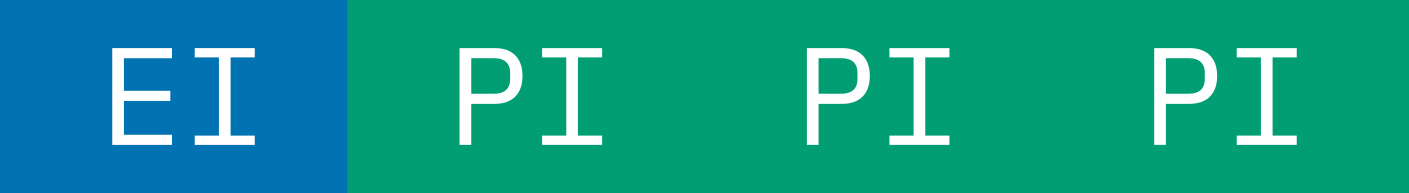}\\
         explore-exploit (\num{0.5}) & \includegraphics[width=3cm]{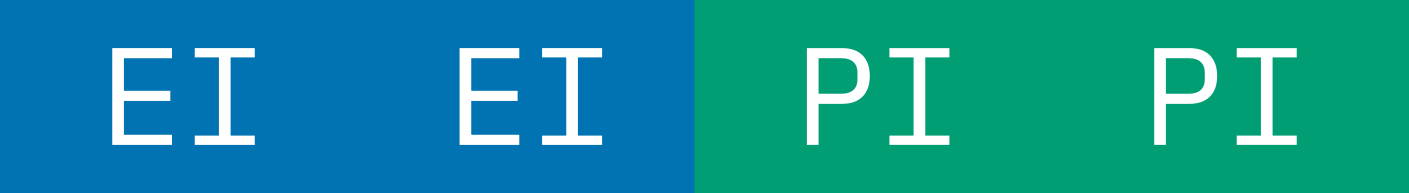}\\
         explore-exploit (\num{0.75}) & \includegraphics[width=3cm]{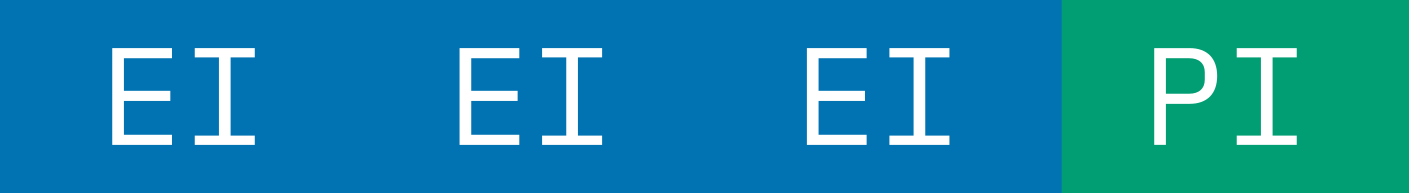}\\
         \bottomrule
    \end{tabular}
    }
}{%
  \caption{Schedule portfolio of PI, EI and their combinations.}%
  \label{tab:schedules}
}
\end{floatrow}
\end{figure}

Each component of the sequential \ac{BO} pipeline is subject to a wide range of design choices, which in turn affects the overall performance of the \ac{BO} procedure~\citep{lindauer-dso19a,cowenrivers-arxiv21a,BossekDK20}.
For example, each \acf{AF} exhibits different behavior, with some being more explorative and others more exploitative.
To date, design choices have been predominantly \emph{static} during \ac{BO} execution.
However, other areas of optimization indicate that \emph{dynamic} choices can lead to performance gains~\cite{karafotias-tec15a,DoerrD18chapter,adriaensen-arxiv22a}.

In this work, we investigate whether the same applies to \ac{BO}. Specifically, we 
propose a \emph{dynamic schedule} that switches between two \ac{AF}s, namely \ac{PI} and \ac{EI}~\cite{forrester_engineering_2008}, \emph{during} the \ac{BO} procedure.
This simple approach is already quite effective and outperforms any single static schedule on most problems.
We assess our method on the BBOB test suite from the COCO library~\citep{hansen-joms21}, which is a standard benchmark in black-box optimization.
We purposely opt for such a benchmark in order to shift focus from a purely performance-oriented view towards \emph{understanding} the problem characteristics that favor one or the other \ac{AF}, which is a stark difference between this paper and previous works on dynamic \ac{AF} selection.
We thus observe that the choice of schedule is largely problem-dependent and that further investigation in this direction is of utmost importance to fully understand and exploit the untapped potential of adaptive \ac{AF} selection.

\textbf{Background }
Since our approach is focused on \ac{PI} and  \ac{EI}, we briefly recall their formulations:
\begin{equation*}
\text{PI}(x)=\Phi\left(\frac{\Delta(x)}{\sigma(x)}\right)
\quad \text{and} \quad 
\text{EI}(x)=\begin{cases}
\Delta(x)\ \Phi\left(\frac{\Delta(x)}{\sigma(x)}\right)+\sigma(x)\ \phi\left(\frac{\Delta(x)}{\sigma(x)}\right) &\mbox{if } \sigma(x)>0 \\

0 &\mbox{if } \sigma(x)=0
\end{cases}
\end{equation*}
with $\Delta(x) = y_{\text{min}} - \mu(x)$, where $ y_{\text{min}}$ is the best observed value so far, and the mean $\mu(x)$ and the variance $\sigma^2(x)$ are queried from the posterior of the surrogate model -- in our work a \ac{GP}. $\Phi$ and $\phi$ are the Gaussian cumulative distribution function and probability density function, respectively.
By definition, in a minimization problem, the maximization of \ac{PI} drives the selection of new candidates for evaluation towards areas of the domain where the mean value of the \ac{GP} model is low.
On the contrary, \ac{EI} balances exploitation (thanks to the first summation term) and exploration (second summation term) by alternating sampling in promising areas of the domain according to the mean value and unexplored ones, where the variance of the \ac{GP} model is high.
Comparing both, EI is more explorative and PI more exploitative, see~\cref{sec:explore_vs_exploit} for an example.

\textbf{Related Work }
We first present existing work on \ac{AF}s and focus on different approaches considering \ac{AF}s.
Important findings are reported when the main target is to improve and develop better \ac{AF}s, from enhancing \ac{EI}~\citep{qin-nips17,balandat-neurips20a} to meta-learning neural \ac{AF}s~\citep{volpp-iclr20a}.
A different line of work is concerned with combining different \ac{AF}s, e.g.~by building a portfolio of \ac{AF}s (\ac{EI}, \c{PI}, \ac{UCB} with different hyperparameter settings) and then using an online multi-armed bandit strategy to assign probabilities of which \ac{AF} to use at which step, called GP-Hedge~\citep{hoffman-uai11}.
Their work indicates that the performance of GP-Hedge highly varies with the number of arms and their respective hyperparameter settings.
Similarly to GP-Hedge, \cite{kandasamy-jmlr2020} update weights of their portfolio (\ac{UCB}, \ac{EI}, \ac{TS}~\citep{thompson-biomet33a}, \ac{TTEI}~\citep{qin-nips17}) in an online manner.
They do not include \ac{PI} as they observe it exhibits inferior performance compared to other single static \ac{AF}s.
In addition, robust versions of \ac{EI}, \ac{PI} and \ac{UCB} can be combined to a multi-objective \ac{AF} combining the strengths of the individual ones~\citep{cowenrivers-arxiv21a}.
In contrast to the existing literature, we here take a step back and ask ourselves what we could achieve by employing a simplistic approach of switching between \ac{AF}s through a dynamic schedule.

It has also been shown in other optimization-related areas that dynamic choices are beneficial in terms of performance, e.g.~in evolutionary computation~\cite{karafotias-tec15a,DoerrD18chapter}, planning~\citep{speck-icaps21local} and deep learning~\citep{adriaensen-arxiv22a}. 
Recently, the introduction of \ac{DAC}~\citep{biedenkapp-ecai20a} underlines the potential of training dynamic schedules (as opposed to selecting algorithm components \emph{on the fly}, as is usually done in evolutionary computation~\cite{hansen-ec03a}).

\section{Proposed Method}
To evaluate the potential of dynamic \ac{AF} schedules, we consider several different variants, summarized in~\cref{tab:schedules}.
As baselines, we use static schedules, namely \ac{EI} and \ac{PI}.
Two types of dynamic schedules are considered: alternating and switching.
The former go back and forth between \ac{EI} and \ac{PI}, whereas the latter make a single switch from \ac{EI} to \ac{PI} after a pre-defined percentage of the budget has been spent.
We introduce two alternating schedules as baselines for a dynamic approach: random and round robin.
For the random schedule the \ac{AF} is chosen uniformly at random before a new solution is sampled, while round robin simply alternates the \ac{AF}s for each sample.
Finally, three different switching schedules are considered, making a switch after \SI{25}{\percent}, \SI{50}{\percent} and \SI{75}{\percent} of the budget remaining after the initial design (i.e., surrogate-based function evaluations), respectively.
This makes it possible to investigate whether the conceptual idea of switching is beneficial and additionally gives insight into the effects of switching at different stages of the optimization process.

\section{Experiments}
We evaluate our schedules on the \num{24} single-objective noiseless BBOB functions of the COCO benchmark~\citep{hansen-joms21} in dimensions \num{5} with \num{60} seeds (i.e., we perform \num{60} independent runs on each problem and for each \ac{AF} schedule).
We optimize our functions with an initial design (or: \ac{DoE}) of $3d=15$ data points and \ac{BO} length of $20d=100$ function evaluations. 
Per seed, the initial design is the same for every schedule.
To create the dynamic \ac{BO}, SMAC3~\citep{lindauer-jmlr22a} is adapted accordingly and we use a \ac{GP} as surrogate model.
All experiments were conducted on a Slurm CPU cluster with \num{1592} CPUs available across nodes.
For visualization, we consider the log-regret of the incumbent (best evaluated  search point) and only visualize the evaluations proposed by \ac{BO}, omitting the initial design.
In the violin plots the log-regrets of the final incumbents of all \num{60} seeds are normalized to $[0,1]$ per BBOB function.
In the convergence plots the log-regret of the incumbents of all \num{60} seeds over all evaluations are normalized to $[0,1]$ per BBOB function and the means with the \SI{95}{\percent} confidence intervals are shown.
The plots for each function can be found in~\cref{sec:all_bbob_plots}.
The code can be found here: \url{https://github.com/automl/pi_is_back}.

\textbf{Observations }
\begin{figure}[ht]
\centering
    \begin{subfigure}[b]{0.45\textwidth}
        \centering
        \includegraphics[width=0.7\textwidth]{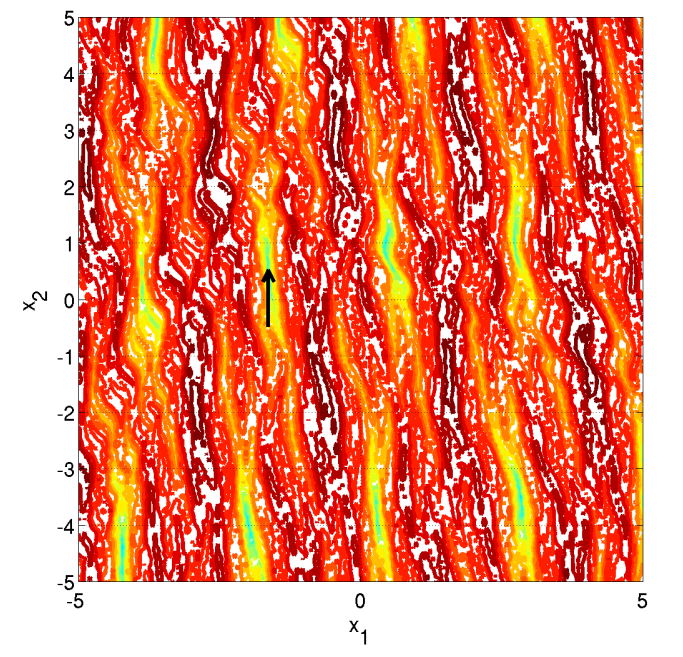}
        \caption{Landscape of Weierstrass (F16)~\citep{hansen-joms21}}
        \label{subfig:bbob_f16}
    \end{subfigure}
    \hfill
    \centering
    \begin{subfigure}[b]{0.45\textwidth}
        \centering
        \includegraphics[width=\textwidth]{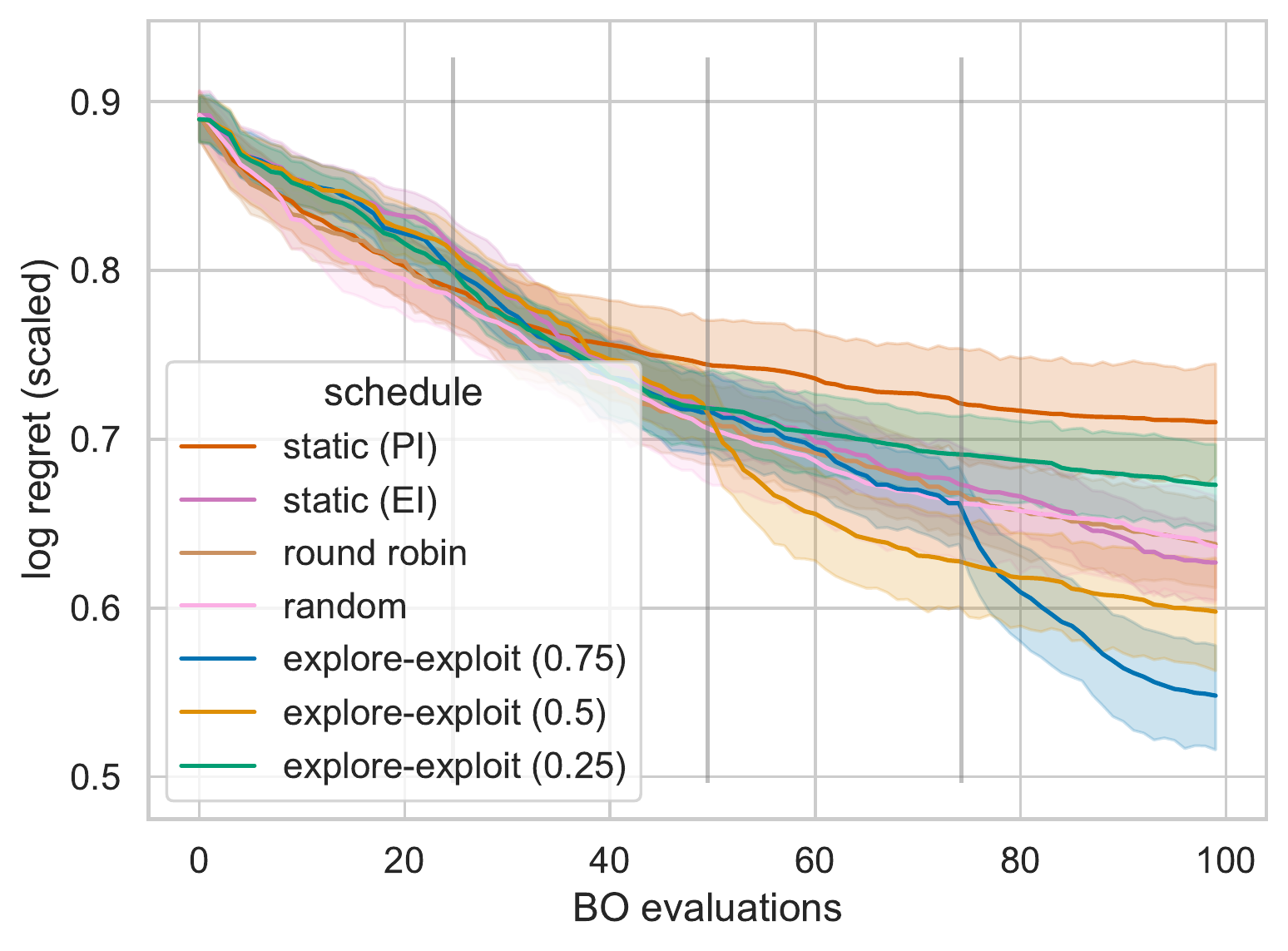}
        \caption{Log-Regret (Scaled) per Step}
        \label{subfig:convergence_16_main}
    \end{subfigure}
    \caption{BBOB Function 16: Highly rugged and moderately repetitive landscape with many local optima. Dynamic schedules outperform static ones and we can clearly observe the boost gained after switching from EI to PI.}
    \label{fig:bbob_function_1_main}
\end{figure}
A first observation is that switching from \ac{EI} to \ac{PI} is in general beneficial when the function landscape has an adequate global structure (F1-F19).
Here, the only exceptions are F5 and F19.
F5 is a purely linear function, where \ac{EI} performs best as it explores fast in the beginning and exploits sufficiently fast later on because of the simplicity of the landscape.
For F19, we hypothesize that only PI is able to exploit  a lucky initial design landing close to optima as indicated by the individual runs in~\cref{subfig:convergence_perseed_19}.


In addition, we can see that PI works well for uni-modal and quite smooth functions (F1-F14), and even better when used after the switch.
This observation is in line with \ac{PI}'s exploitative behavior.
In these cases exploiting does not miss any other optima further away in the landscape.


In contrast, \ac{PI} is in general worse than \ac{EI} and also not beneficial after the switch for multi-modal functions with weak global structure (F20-F24).
Again, this is in line with our intuition because for these functions we have many important basins of attraction that have to be discovered before starting exploitation.
The only exception is F23 (\cref{fig:bbob_function_23}), a rugged function with a high number of global optima, where the probability of starting in a basin of attraction is high and thus exploitation is a viable strategy.
Also, the flatter the landscape the worse PI performs which can be also seen in F7 (slope with step, \cref{fig:bbob_function_7}).


Besides the static and the switching schedules, round robin switches from \ac{EI} to \ac{PI} and vice versa for each new function evaluation.
The round robin schedule creates a step-wise progress, but is never the best strategy.
Apparently, less frequent switching is preferable in order to take advantage of the strengths of each \ac{AF}.
On average, the random schedule performs similarly to round robin, but with a smoother progression.
Most likely, the random schedule still switches too frequently for the \ac{AF}s to work effectively.


On F16 we observe that a late switch from \ac{EI} to \ac{PI} performs best, see~\cref{subfig:convergence_16_main}.
We can also nicely spot the general boosts after switching after $\SI{25}{\percent}$, $\SI{50}{\percent}$ and $\SI{75}{\percent}$.
F16 has a highly rugged and moderately repetitive landscape with many local optima with evident quality difference.
Therefore, PI can be trapped in a bad local optimum if applied too early.


\textbf{Key Insights }
To summarize our observations, the general optimal choice of static or dynamic \ac{AF} schedules depends on the problem at hand.
Plus, switching from \ac{EI} to \ac{PI} after a certain percentage of the budget of total number of function evaluations is beneficial and can boost performance.
However, the optimal switching point depends on the problem.
On average, switching after $\SI{25}{\percent}$ seems to be best and is a viable schedule if the problem's properties are unknown.

\section{Conclusions and Future Work}
In this work we acknowledge the modular structure of \acf{BO} and explore different \acl{AF} schedules composed of \acf{EI} and \acf{PI}, with the premise that a dynamic choice would boost \ac{BO} performance.
We evaluate our approach on a standard benchmark for black-box optimization, and observe that dynamic \acf{AF} schedules often outperform single static ones, although their performance depends on the problem at hand.
As a general recommendation, if the properties of the problem to optimize are unknown, switching from \ac{EI} to \ac{PI} after $\SI{25}{\percent}$ of the budget of surrogate-based evaluations should be favored.
We also highlight some key insights into the interplay of the nature of the problem and the performance of the proposed approach.

These preliminary findings suggest the potential of the dynamic choices in a \ac{BO} pipeline merit further investigation.
As future steps in this direction,  finding the most appropriate schedule of different \ac{AF}s is an open problem we could tackle with meta-learning.
One way would be to train models recommending which \ac{AF} to use at which step based on problem representation, i.e., typically \ac{ELA} features~\citep{mersmann-gecco11}.
In addition, finding meaningful state features could enable an online selection of the \ac{AF} via heuristics or Reinforcement Learning~\citep{adriaensen-arxiv22a}.
To this end, besides elapsed time and progress being default choices as state features, we could incorporate properties of the surrogate model to serve as a guide to select the \ac{AF}.
Furthermore, including more \ac{AF}s and their hyperparameters in the portfolio, as well as dynamically changing other components of \ac{BO}, like the surrogate model, the \ac{GP} kernel, or the optimizer of the \ac{AF}, might increase overall performance and sample-efficiency of \ac{BO}.

\subsection*{Acknowledgments}
Our work has been financially supported by the the ANR T-ERC project \emph{VARIATION} (ANR-22-ERCS-0003-01), by the CNRS INS2I project \emph{RandSearch}, and by the PRIME programme of the German Academic Exchange Service (DAAD) with funds from the German Federal Ministry of Education and Research (BMBF), and by RFBR and CNRS, project number 20-51-15009.
Carolin Benjamins and Marius Lindauer acknowledge funding by the German Research Foundation (DFG) under LI 2801/4-1.

\bibliography{lib,references,shortproc}

\begin{thebibliography}{20}
\providecommand{\natexlab}[1]{#1}
\providecommand{\url}[1]{\texttt{#1}}
\expandafter\ifx\csname urlstyle\endcsname\relax
  \providecommand{\doi}[1]{doi: #1}\else
  \providecommand{\doi}{doi: \begingroup \urlstyle{rm}\Url}\fi

\bibitem[Mockus(2012)]{mockus_bayesian_2012}
Jonas Mockus.
\newblock \emph{Bayesian {{Approach}} to {{Global Optimization}}: {{Theory}}
  and {{Applications}}}.
\newblock {Springer Science \& Business Media}, December 2012.
\newblock ISBN 978-94-009-0909-0.

\bibitem[Lindauer et~al.(2019)Lindauer, Feurer, Eggensperger, Biedenkapp, and
  Hutter]{lindauer-dso19a}
M.~Lindauer, M.~Feurer, K.~Eggensperger, A.~Biedenkapp, and F.~Hutter.
\newblock Towards assessing the impact of bayesian optimization's own
  hyperparameters.
\newblock In \emph{{IJCAI}'19 {DSO} {W}orkshop}, 2019.

\bibitem[Cowen-Rivers et~al.(2021)Cowen-Rivers, Lyu, Tutunov, Wang, Grosnit,
  Griffiths, Maraval, Jianye, Wang, Peters, and Ammar]{cowenrivers-arxiv21a}
A.~Cowen-Rivers, W.~Lyu, R.~Tutunov, Z.~Wang, A.~Grosnit, R.~Griffiths,
  A.~Maraval, H.~Jianye, J.~Wang, J.~Peters, and H.~Ammar.
\newblock An empirical study of assumptions in {Bayesian} optimisation.
\newblock \emph{arXiv:2012.03826 [cs.LG]}, 2021.

\bibitem[Bossek et~al.(2020)Bossek, Doerr, and Kerschke]{BossekDK20}
Jakob Bossek, Carola Doerr, and Pascal Kerschke.
\newblock Initial design strategies and their effects on sequential model-based
  optimization: an exploratory case study based on {BBOB}.
\newblock In \emph{Proc. of Genetic and Evolutionary Computation Conference
  (GECCO'20)}, pages 778--786. ACM, 2020.
\newblock \doi{10.1145/3377930.3390155}.
\newblock URL \url{https://doi.org/10.1145/3377930.3390155}.

\bibitem[Karafotias et~al.(2015)Karafotias, Hoogendoorn, and
  Eiben]{karafotias-tec15a}
G.~Karafotias, M.~Hoogendoorn, and {\'{A}}.~Eiben.
\newblock Parameter control in evolutionary algorithms: Trends and challenges.
\newblock \emph{{IEEE} Trans. Evolutionary Computation}, 19\penalty0
  (2):\penalty0 167--187, 2015.

\bibitem[Doerr and Doerr(2020)]{DoerrD18chapter}
Benjamin Doerr and Carola Doerr.
\newblock Theory of parameter control mechanisms for discrete black-box
  optimization: Provable performance gains through dynamic parameter choices.
\newblock In \emph{Theory of Evolutionary Computation: Recent Developments in
  Discrete Optimization}, pages 271--321. Springer, 2020.
\newblock \doi{10.1007/978-3-030-29414-4_6}.
\newblock Also available online at \url{https://arxiv.org/abs/1804.05650}.

\bibitem[Adriaensen et~al.(2022)Adriaensen, Biedenkapp, Shala, Awad, Eimer,
  Lindauer, and Hutter]{adriaensen-arxiv22a}
S.~Adriaensen, A.~Biedenkapp, G.~Shala, N.~Awad, T.~Eimer, M.~Lindauer, and
  F.~Hutter.
\newblock Automated dynamic algorithm configuration.
\newblock \emph{arXiv:2205.13881 [cs.AI]}, 2022.

\bibitem[Forrester et~al.(2008)Forrester, S{\'o}bester, and
  Keane]{forrester_engineering_2008}
Alexander I.~J. Forrester, Andr{\'a}s S{\'o}bester, and Andy~J. Keane.
\newblock \emph{Engineering {{Design}} via {{Surrogate Modelling}} - {{A
  Practical Guide}}}.
\newblock {John Wiley \& Sons Ltd.}, 2008.
\newblock ISBN 978-0-470-06068-1.

\bibitem[Hansen et~al.(2021)Hansen, Auger, Ros, Mersmann, Tu{\v s}ar, and
  Brockhoff]{hansen-joms21}
N.~Hansen, A.~Auger, R.~Ros, O.~Mersmann, T.~Tu{\v s}ar, and D.~Brockhoff.
\newblock {COCO}: A platform for comparing continuous optimizers in a black-box
  setting.
\newblock \emph{Optimization Methods and Software}, 36:\penalty0 114--144,
  2021.
\newblock \doi{10.1080/10556788.2020.1808977}.

\bibitem[Qin et~al.(2017)Qin, Klabjan, and Russo]{qin-nips17}
Chao Qin, Diego Klabjan, and Daniel Russo.
\newblock Improving the expected improvement algorithm.
\newblock In Isabelle Guyon, Ulrike von Luxburg, Samy Bengio, Hanna~M. Wallach,
  Rob Fergus, S.~V.~N. Vishwanathan, and Roman Garnett, editors, \emph{Advances
  in Neural Information Processing Systems 30: Annual Conference on Neural
  Information Processing Systems 2017, December 4-9, 2017, Long Beach, CA,
  {USA}}, pages 5381--5391, 2017.
\newblock URL
  \url{https://proceedings.neurips.cc/paper/2017/hash/b19aa25ff58940d974234b48391b9549-Abstract.html}.

\bibitem[Balandat et~al.(2020)Balandat, Karrer, Jiang, Daulton, Letham, Wilson,
  and Bakshy]{balandat-neurips20a}
M.~Balandat, B.~Karrer, D.~Jiang, S.~Daulton, B.~Letham, A.~Wilson, and
  E.~Bakshy.
\newblock Botorch: A framework for efficient monte-carlo {Bayesian}
  optimization.
\newblock In \emph{Proc. of {N}eur{IPS}'20}, 2020.

\bibitem[Volpp et~al.(2020)Volpp, Fröhlich, Fischer, Doerr, Falkner, Hutter,
  and Daniel]{volpp-iclr20a}
M.~Volpp, L.~Fröhlich, K.~Fischer, A.~Doerr, S.~Falkner, F.~Hutter, and
  C.~Daniel.
\newblock Meta-learning acquisition functions for transfer learning in bayesian
  optimization.
\newblock In \emph{Proc. of {ICLR}'20}, 2020.
\newblock URL \url{https://openreview.net/forum?id=ryeYpJSKwr}.

\bibitem[Hoffman et~al.(2011)Hoffman, Brochu, and de~Freitas]{hoffman-uai11}
Matthew Hoffman, Eric Brochu, and Nando de~Freitas.
\newblock Portfolio allocation for {Bayesian} optimization.
\newblock In \emph{Proceedings of the Twenty-Seventh Conference on Uncertainty
  in Artificial Intelligence}, UAI'11, pages 327--336, Arlington, Virginia,
  USA, 2011. AUAI Press.
\newblock ISBN 9780974903972.

\bibitem[Kandasamy et~al.(2020)Kandasamy, Vysyaraju, Neiswanger, Paria,
  Collins, Schneider, Poczos, and Xing]{kandasamy-jmlr2020}
Kirthevasan Kandasamy, Karun~Raju Vysyaraju, Willie Neiswanger, Biswajit Paria,
  Christopher~R Collins, Jeff Schneider, Barnabas Poczos, and Eric~P Xing.
\newblock Tuning hyperparameters without grad students: Scalable and robust
  {Bayesian} optimisation with dragonfly.
\newblock \emph{J. Mach. Learn. Res.}, 21\penalty0 (81):\penalty0 1--27, 2020.

\bibitem[Thompson(1933)]{thompson-biomet33a}
W.~Thompson.
\newblock On the likelihood that one unknown probability exceeds another in
  view of the evidence of two samples.
\newblock \emph{Biometrika}, 25\penalty0 (3/4):\penalty0 285--294, 1933.

\bibitem[Speck et~al.(2021)Speck, Biedenkapp, Hutter, Mattm{\"{u}}ller, and
  Lindauer]{speck-icaps21local}
David Speck, Andr{\'{e}} Biedenkapp, Frank Hutter, Robert Mattm{\"{u}}ller, and
  Marius Lindauer.
\newblock Learning heuristic selection with dynamic algorithm configuration.
\newblock In Susanne Biundo, Minh Do, Robert Goldman, Michael Katz, Qiang Yang,
  and Hankz~Hankui Zhuo, editors, \emph{Proceedings of the Thirty-First
  International Conference on Automated Planning and Scheduling, {ICAPS} 2021,
  Guangzhou, China (virtual), August 2-13, 2021}, pages 597--605. {AAAI} Press,
  2021.
\newblock URL \url{https://ojs.aaai.org/index.php/ICAPS/article/view/16008}.

\bibitem[Biedenkapp et~al.(2020)Biedenkapp, Bozkurt, Eimer, Hutter, and
  Lindauer]{biedenkapp-ecai20a}
A.~Biedenkapp, H.~F. Bozkurt, T.~Eimer, F.~Hutter, and M.~Lindauer.
\newblock Dynamic algorithm configuration: Foundation of a new meta-algorithmic
  framework.
\newblock In \emph{Proc. of {ECAI}'20}, pages 427--434, 2020.

\bibitem[Hansen et~al.(2003)Hansen, M{\"{u}}ller, and
  Koumoutsakos]{hansen-ec03a}
N.~Hansen, S.~D. M{\"{u}}ller, and P.~Koumoutsakos.
\newblock Reducing the time complexity of the derandomized evolution strategy
  with covariance matrix adaptation {(CMA-ES)}.
\newblock \emph{Evolutionary Computing}, 11\penalty0 (1):\penalty0 1--18, 2003.

\bibitem[Lindauer et~al.(2022)Lindauer, Eggensperger, Feurer, Biedenkapp, Deng,
  Benjamins, Ruhkopf, Sass, and Hutter]{lindauer-jmlr22a}
M.~Lindauer, K.~Eggensperger, M.~Feurer, A.~Biedenkapp, D.~Deng, C.~Benjamins,
  T.~Ruhkopf, R.~Sass, and F.~Hutter.
\newblock {SMAC3}: A versatile bayesian optimization package for
  {H}yperparameter {O}ptimization.
\newblock \emph{Journal of Machine Learning Research (JMLR) -- MLOSS},
  23\penalty0 (54):\penalty0 1--9, 2022.

\bibitem[Mersmann et~al.(2011)Mersmann, Bischl, Trautmann, Preuss, Weihs, and
  Rudolph]{mersmann-gecco11}
Olaf Mersmann, Bernd Bischl, Heike Trautmann, Mike Preuss, Claus Weihs, and
  G{\"{u}}nter Rudolph.
\newblock Exploratory landscape analysis.
\newblock In Natalio Krasnogor and Pier~Luca Lanzi, editors, \emph{13th Annual
  Genetic and Evolutionary Computation Conference, {GECCO} 2011, Proceedings,
  Dublin, Ireland, July 12-16, 2011}, pages 829--836. {ACM}, 2011.
\newblock \doi{10.1145/2001576.2001690}.
\newblock URL \url{https://doi.org/10.1145/2001576.2001690}.

\end{thebibliography}
\bibliographystyle{unsrtnat}

\appendix

\newpage
\section{Exploration vs. Exploitation}
\label{sec:explore_vs_exploit}
\begin{figure}[ht]
    \centering
    \includegraphics[width=0.8\textwidth]{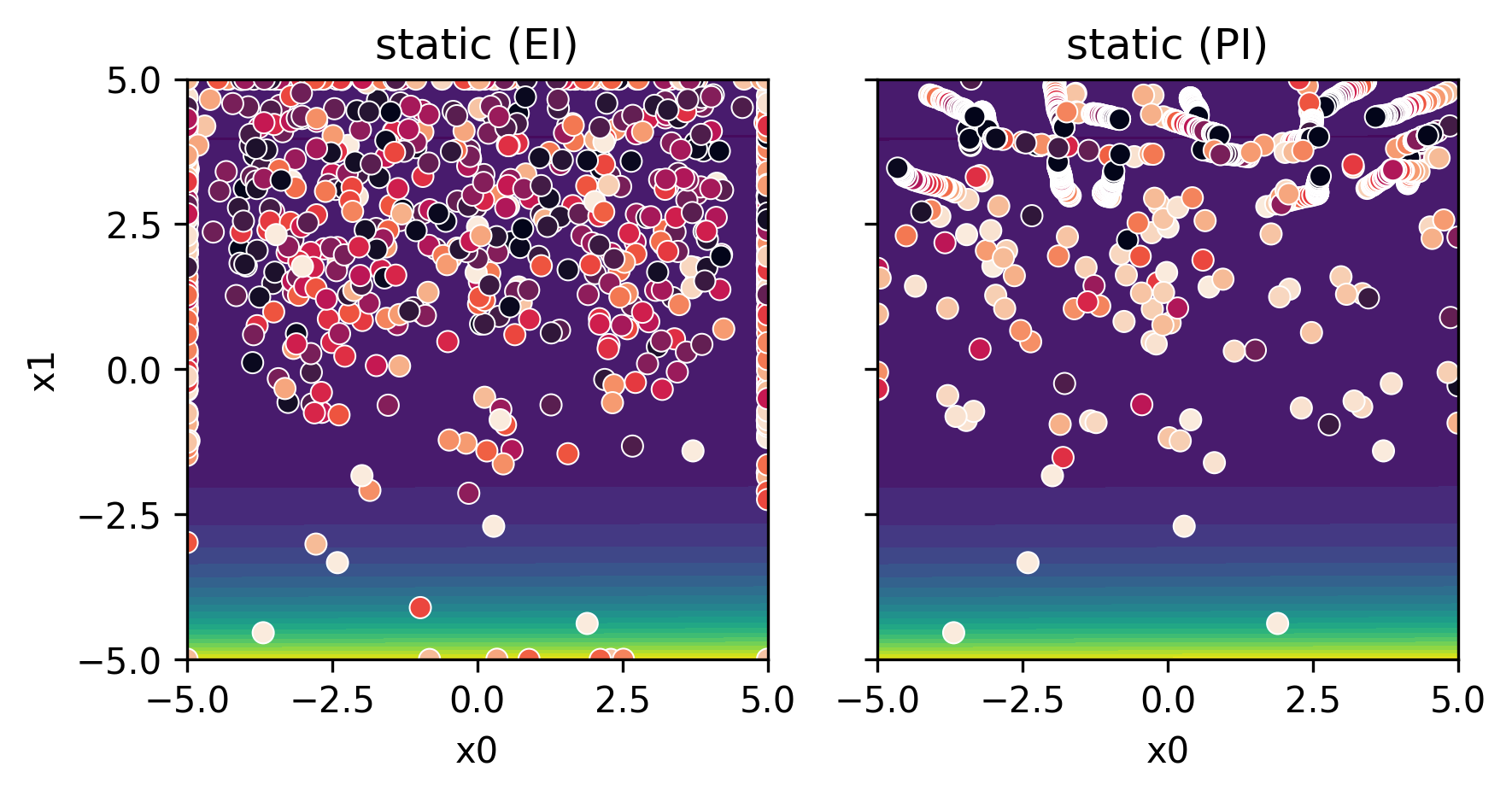}\\
    \includegraphics[width=0.8\textwidth]{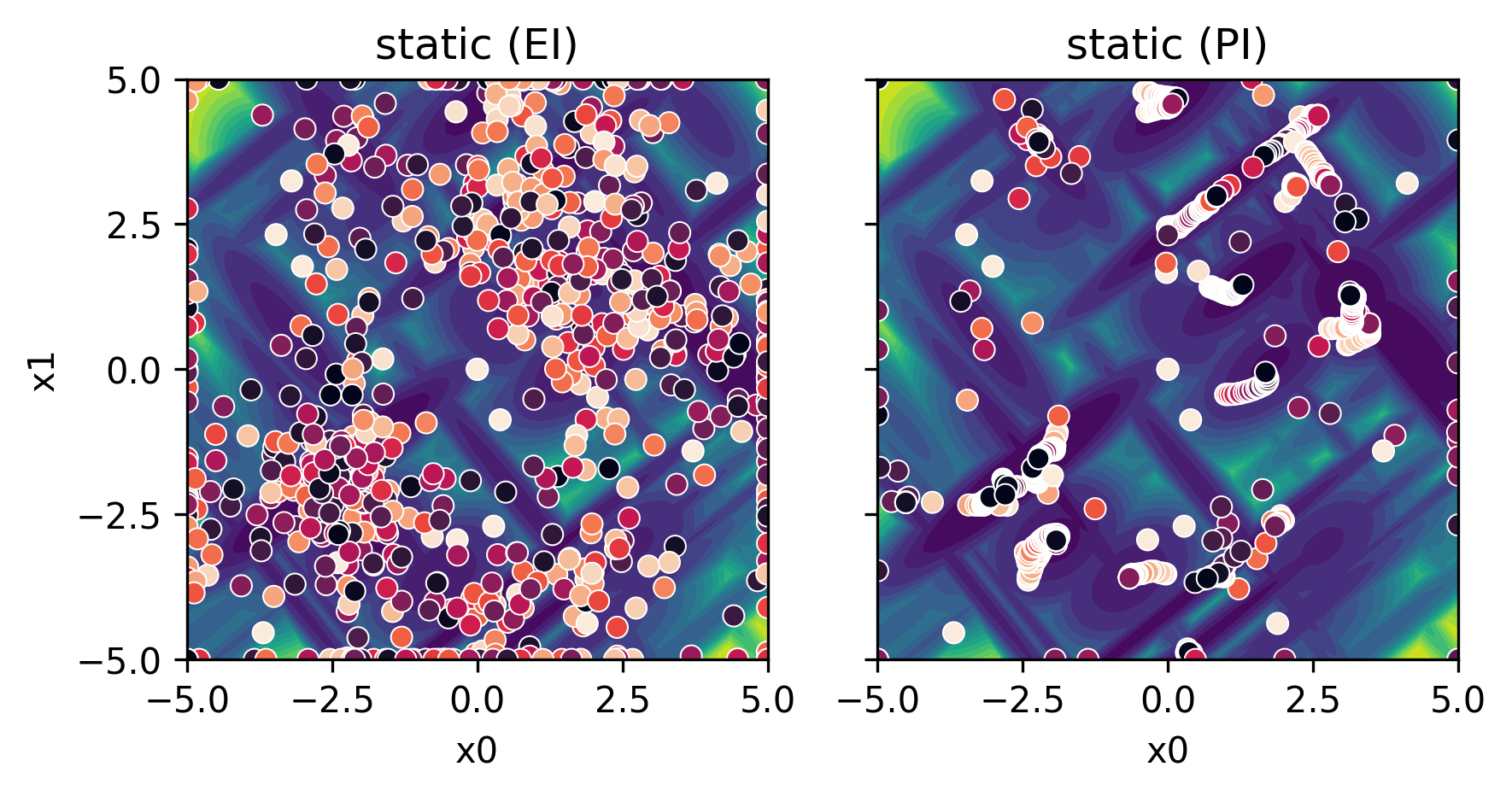}\\
    \includegraphics[height=1cm]{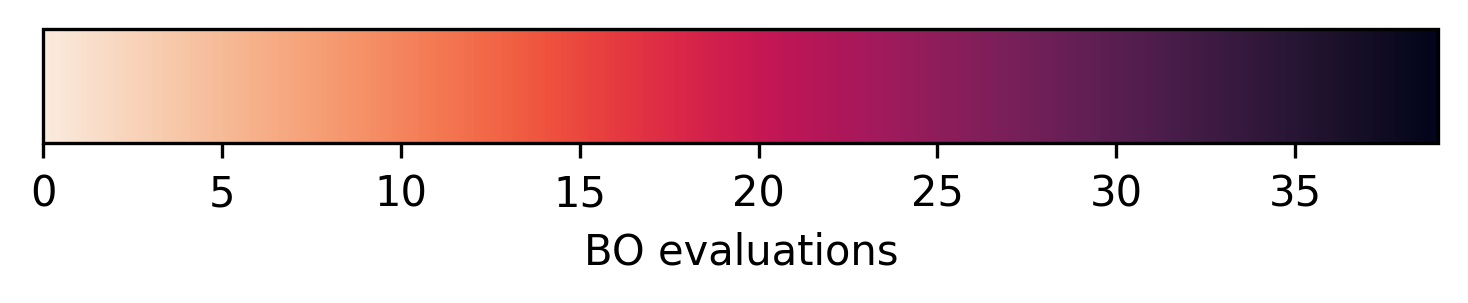}
    \caption{EI vs. PI on BBOB function 12 (upper) and 21 (lower). The dots indicate the visited configuration at a certain step. We visualize all \num{20} seeds with an initial design size of \num{10} (not plotted) and \num{40} BO evaluations.}
    \label{fig:ei_vs_pi_21}
\end{figure}

We choose EI and PI to build our \ac{AF} schedules because of their different search behavior.
In~\cref{fig:ei_vs_pi_21} we exemplarily show EI's explorative character and PI's exploitative character.
Both behaviors can be beneficial depending on the problem to optimize.

\newpage

\section{Additional Results}
\label{sec:all_bbob_plots}

\begin{figure}[h]
    \centering
    \begin{subfigure}[b]{0.45\textwidth}
        \centering
        \includegraphics[width=\textwidth]{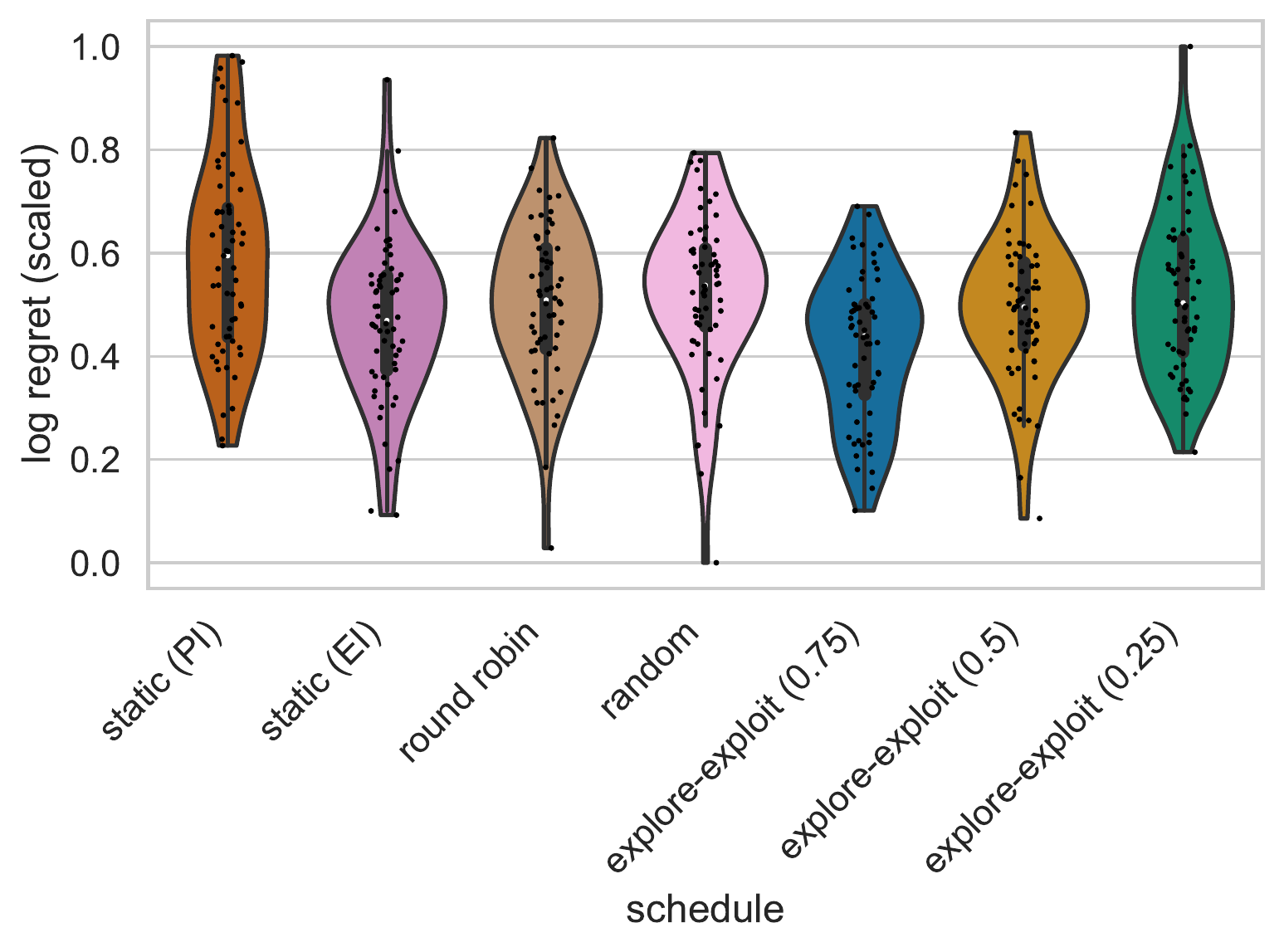}
        \caption{Final Log Regret (Scaled)}
        \label{subfig:boxplot_1}
    \end{subfigure}
    \hfill
    \begin{subfigure}[b]{0.45\textwidth}
        \centering
        \includegraphics[width=\textwidth]{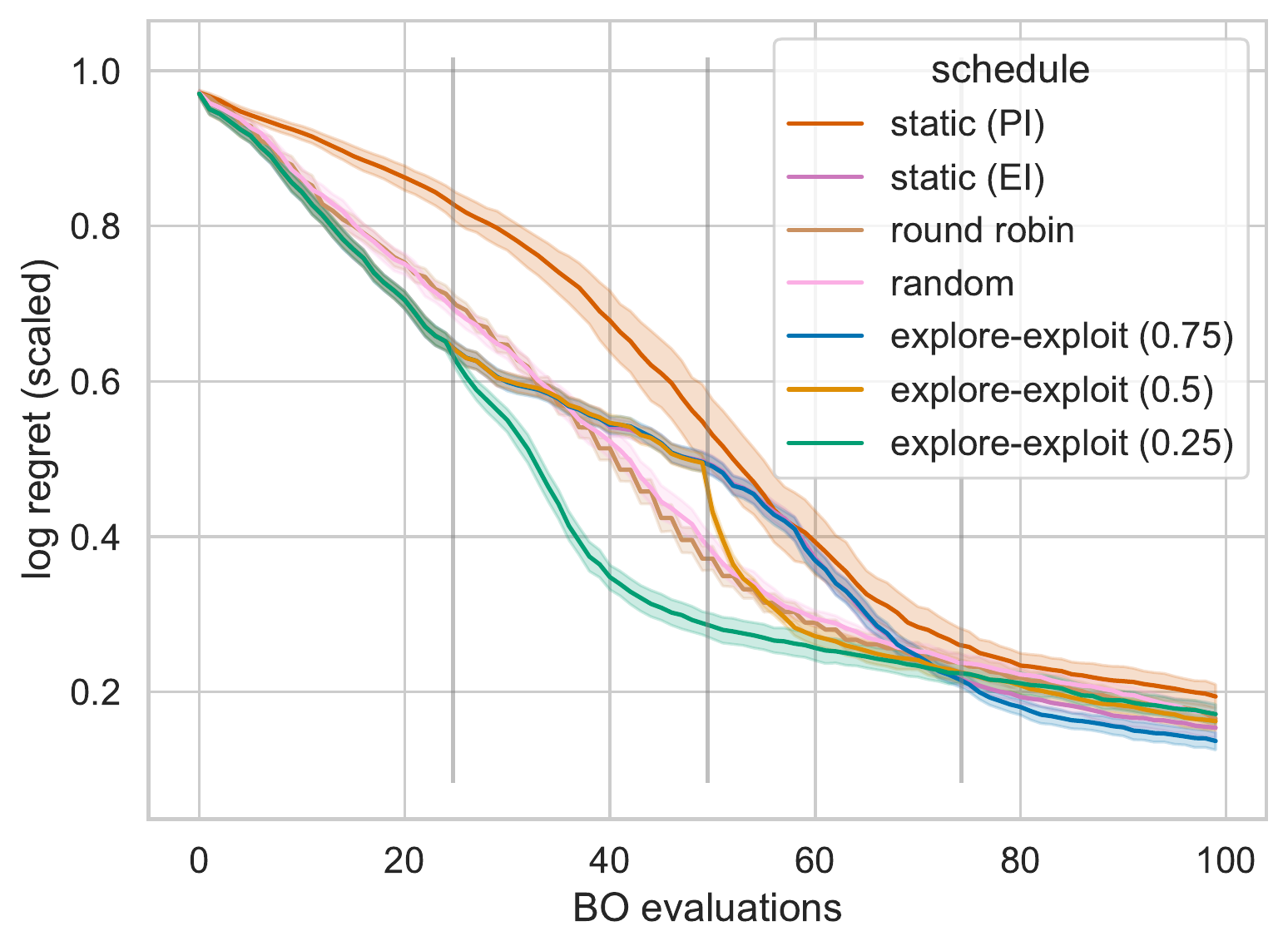}
        \caption{Log-Regret (Scaled) per Step}
        \label{subfig:convergence_1}
    \end{subfigure}\\
    \vspace*{3mm}
    \centering
    \begin{subfigure}[b]{\textwidth}
        \centering
        \includegraphics[width=\textwidth]{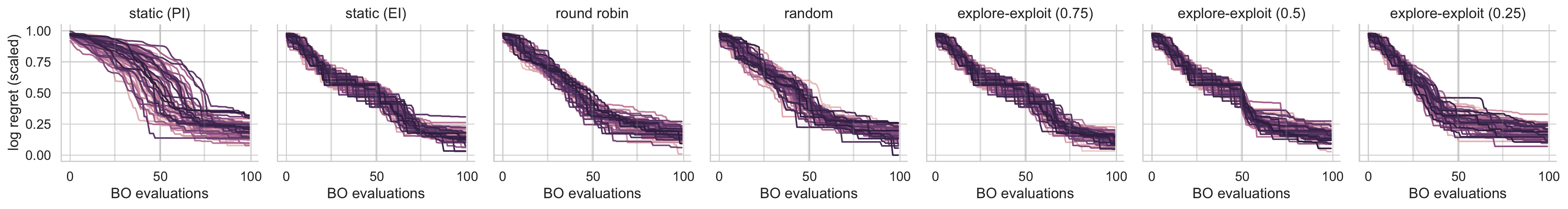}
        \caption{Log-Regret (Scaled) per Step and Seed}
        \label{subfig:convergence_perseed_1}
    \end{subfigure}
    \caption{BBOB Function 1}
    \label{fig:bbob_function_1}
\end{figure}

\begin{figure}[h]
    \centering
    \begin{subfigure}[b]{0.45\textwidth}
        \centering
        \includegraphics[width=\textwidth]{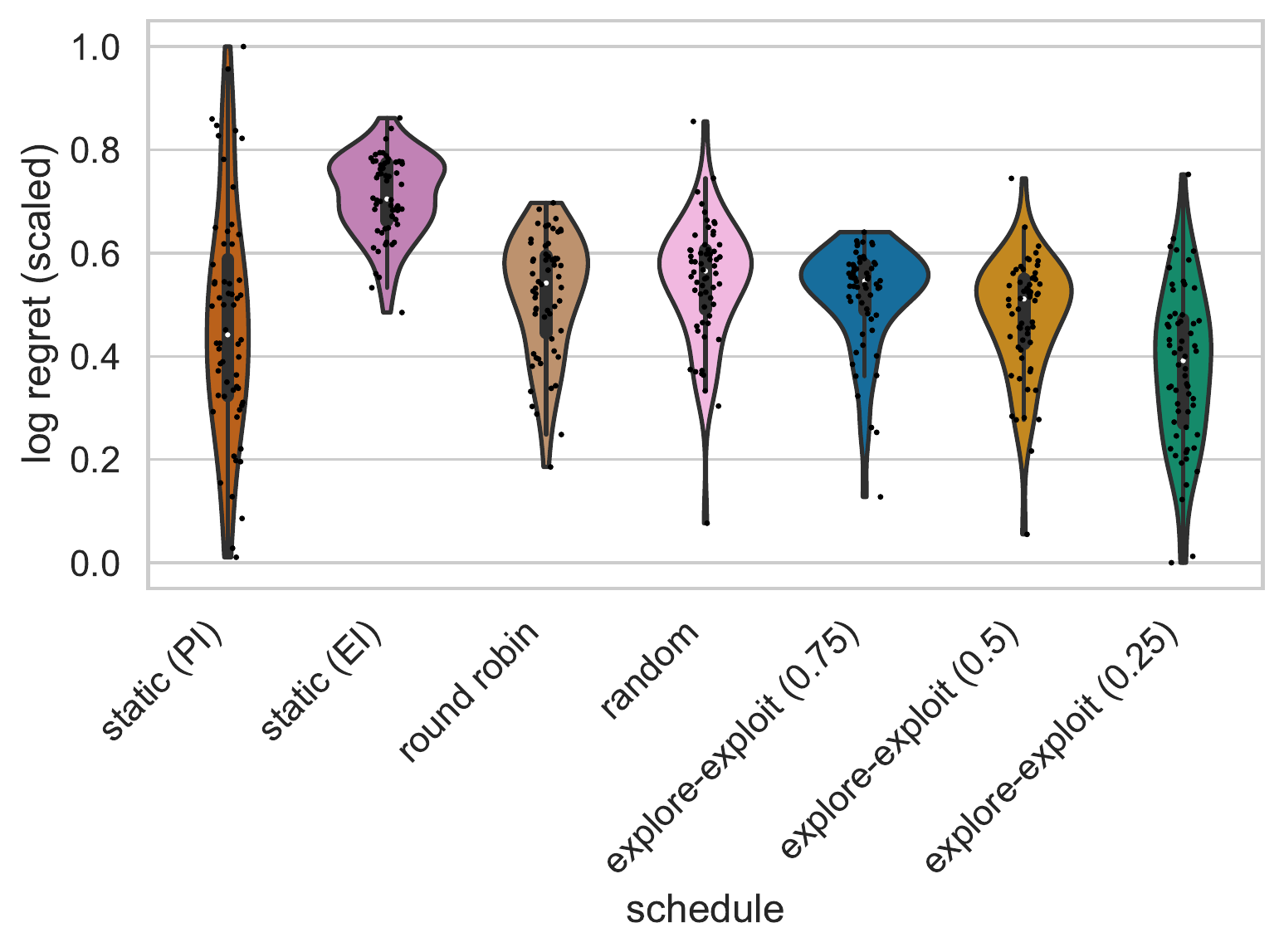}
        \caption{Final Log Regret (Scaled)}
        \label{subfig:boxplot_2}
    \end{subfigure}
    \hfill
    \begin{subfigure}[b]{0.45\textwidth}
        \centering
        \includegraphics[width=\textwidth]{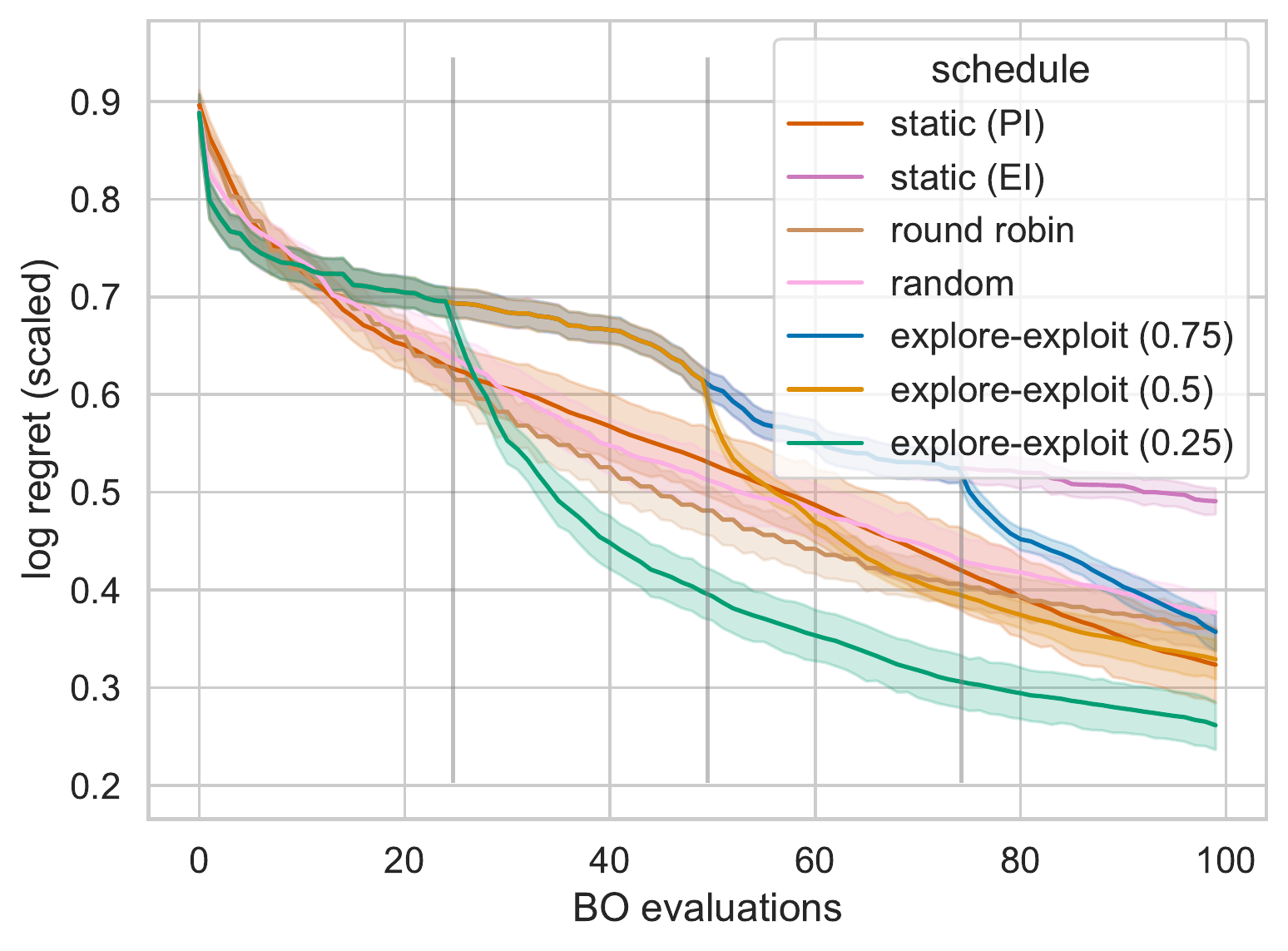}
        \caption{Log-Regret (Scaled) per Step}
        \label{subfig:convergence_2}
    \end{subfigure}\\
    \vspace*{3mm}
    \centering
    \begin{subfigure}[b]{\textwidth}
        \centering
        \includegraphics[width=\textwidth]{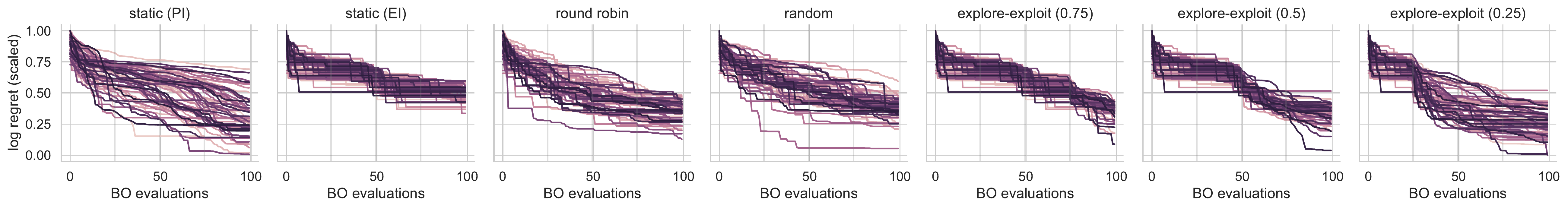}
        \caption{Log-Regret (Scaled) per Step and Seed}
        \label{subfig:convergence_perseed_2}
    \end{subfigure}
    \caption{BBOB Function 2}
    \label{fig:bbob_function_2}
\end{figure}

\begin{figure}[h]
    \centering
    \begin{subfigure}[b]{0.45\textwidth}
        \centering
        \includegraphics[width=\textwidth]{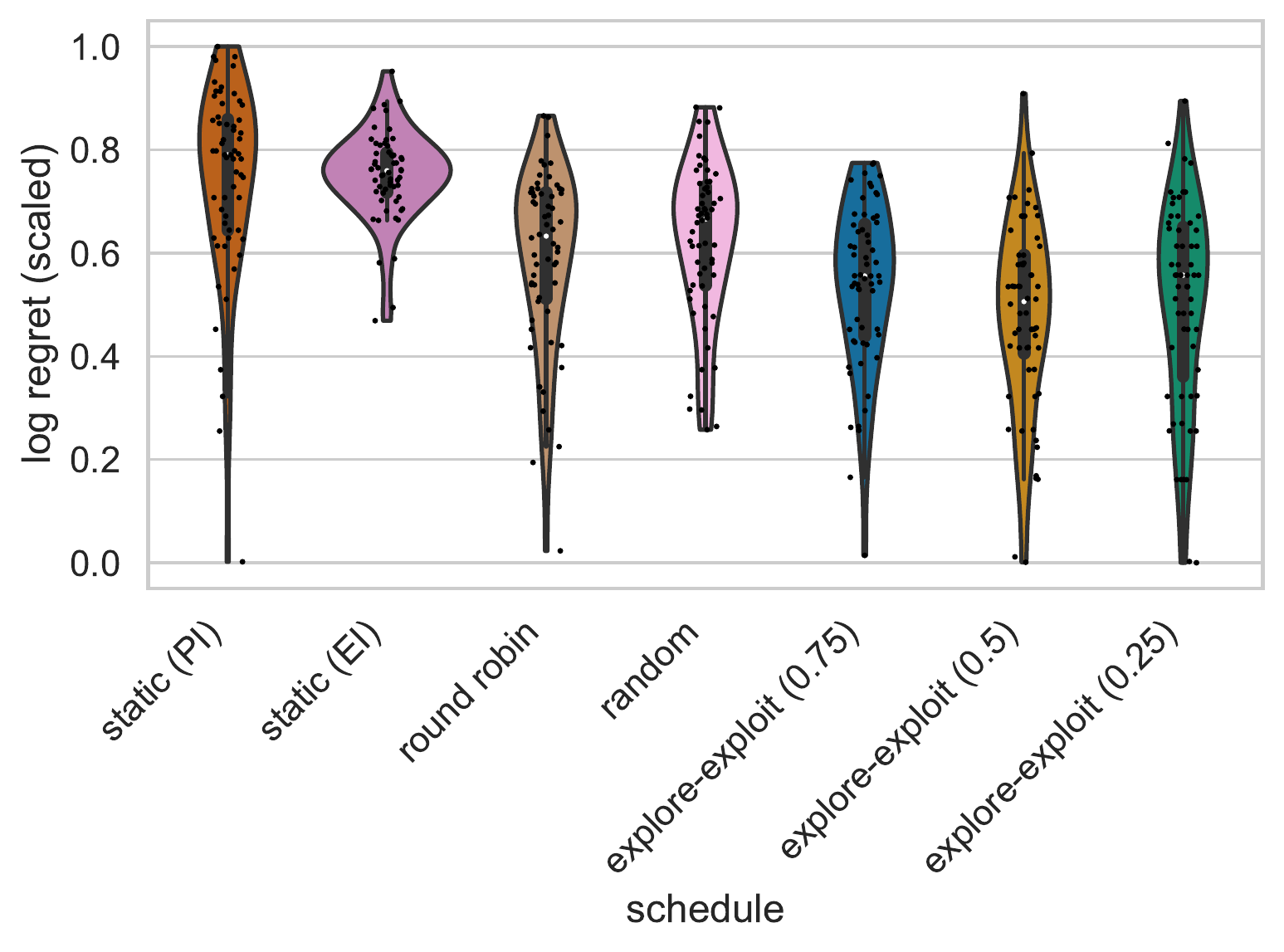}
        \caption{Final Log Regret (Scaled)}
        \label{subfig:boxplot_3}
    \end{subfigure}
    \hfill
    \begin{subfigure}[b]{0.45\textwidth}
        \centering
        \includegraphics[width=\textwidth]{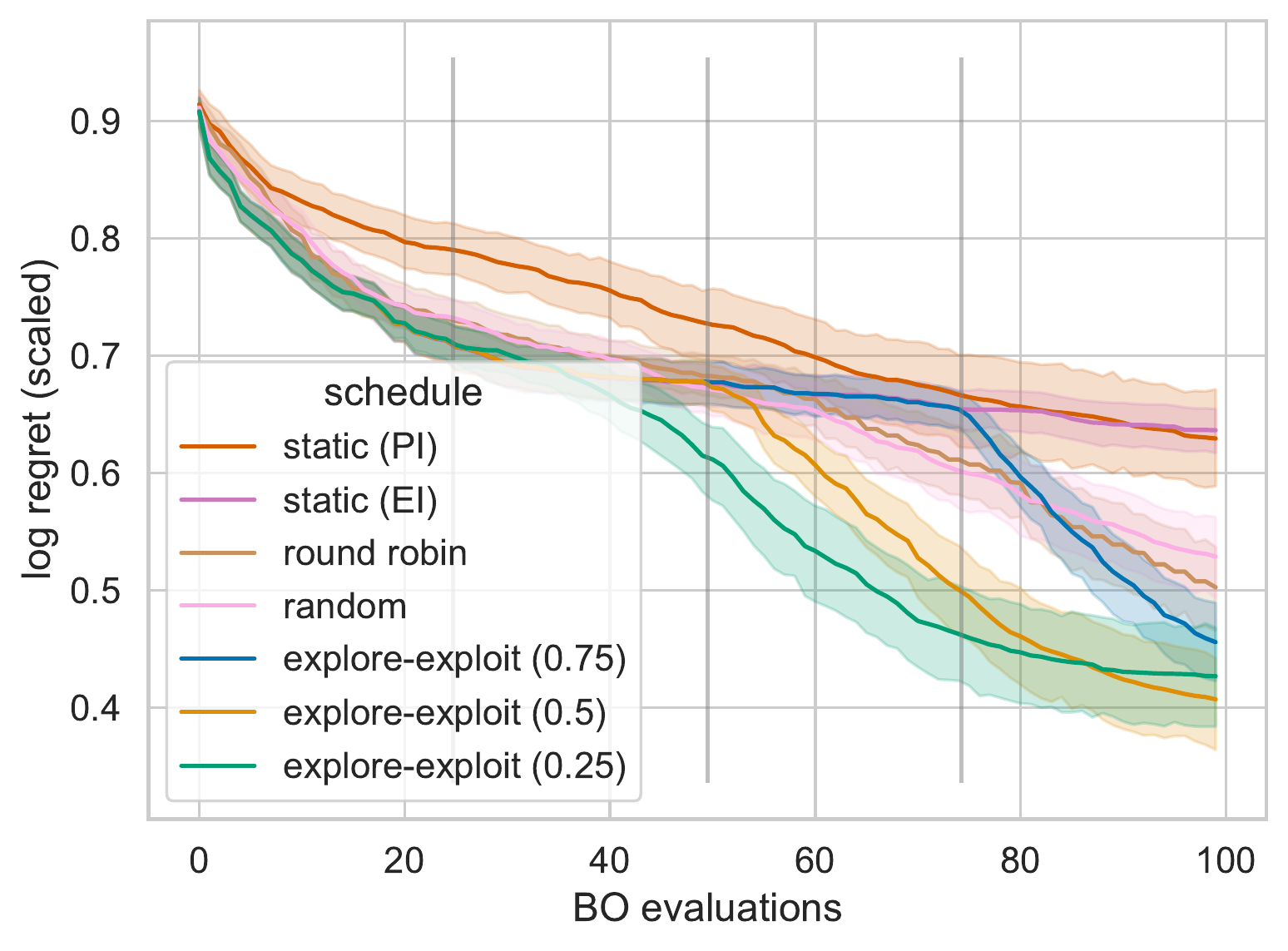}
        \caption{Log-Regret (Scaled) per Step}
        \label{subfig:convergence_3}
    \end{subfigure}\\
    \vspace*{3mm}
    \centering
    \begin{subfigure}[b]{\textwidth}
        \centering
        \includegraphics[width=\textwidth]{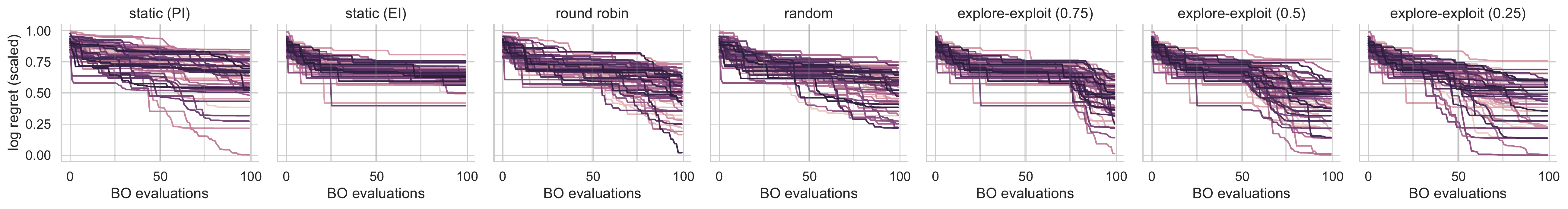}
        \caption{Log-Regret (Scaled) per Step and Seed}
        \label{subfig:convergence_perseed_3}
    \end{subfigure}
    \caption{BBOB Function 3}
    \label{fig:bbob_function_3}
\end{figure}

\begin{figure}[h]
    \centering
    \begin{subfigure}[b]{0.45\textwidth}
        \centering
        \includegraphics[width=\textwidth]{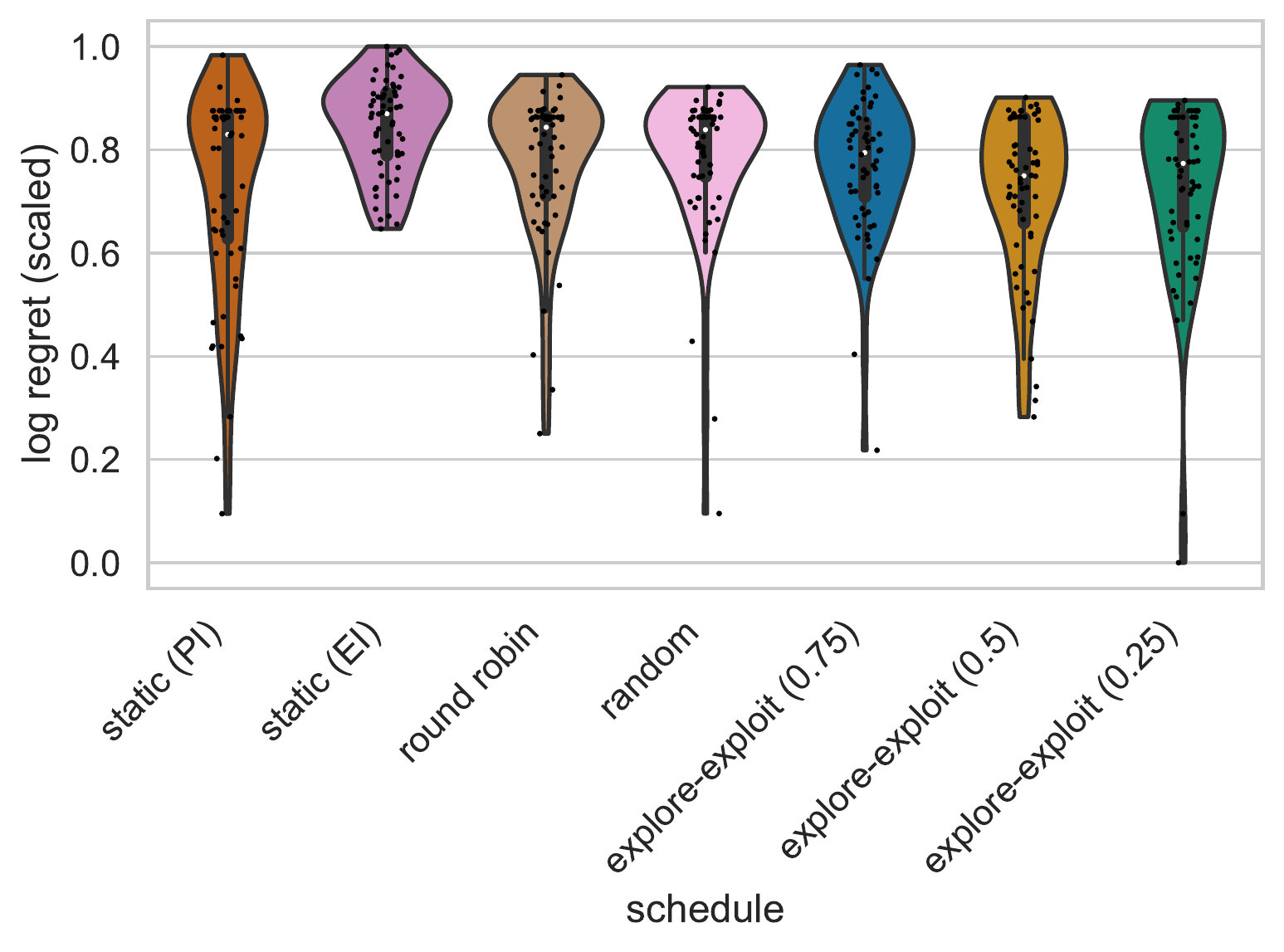}
        \caption{Final Log Regret (Scaled)}
        \label{subfig:boxplot_4}
    \end{subfigure}
    \hfill
    \begin{subfigure}[b]{0.45\textwidth}
        \centering
        \includegraphics[width=\textwidth]{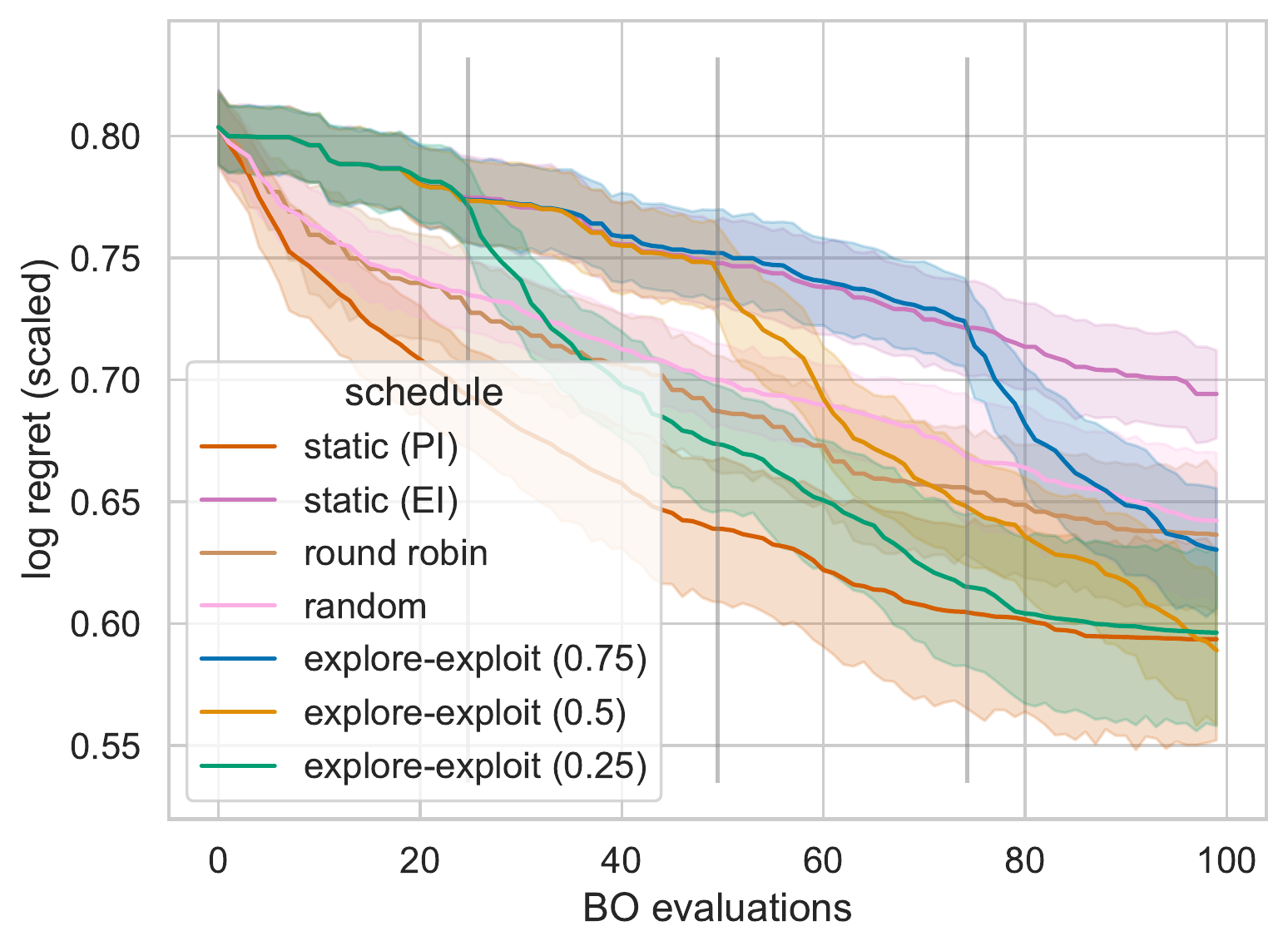}
        \caption{Log-Regret (Scaled) per Step}
        \label{subfig:convergence_4}
    \end{subfigure}\\
    \vspace*{3mm}
    \centering
    \begin{subfigure}[b]{\textwidth}
        \centering
        \includegraphics[width=\textwidth]{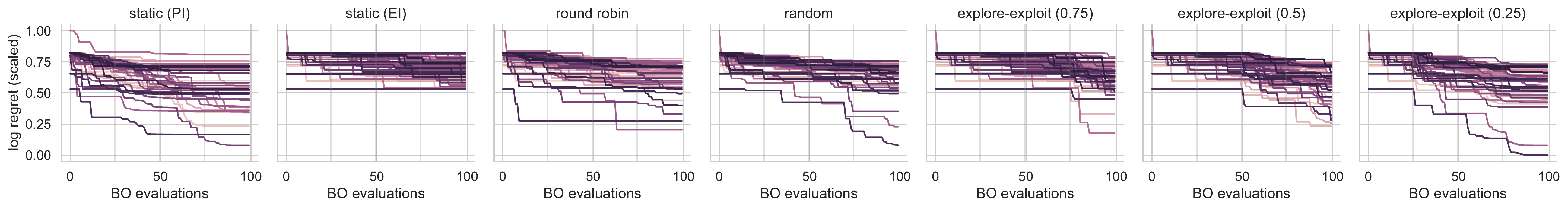}
        \caption{Log-Regret (Scaled) per Step and Seed}
        \label{subfig:convergence_perseed_4}
    \end{subfigure}
    \caption{BBOB Function 4}
    \label{fig:bbob_function_4}
\end{figure}

\begin{figure}[h]
    \centering
    \begin{subfigure}[b]{0.45\textwidth}
        \centering
        \includegraphics[width=\textwidth]{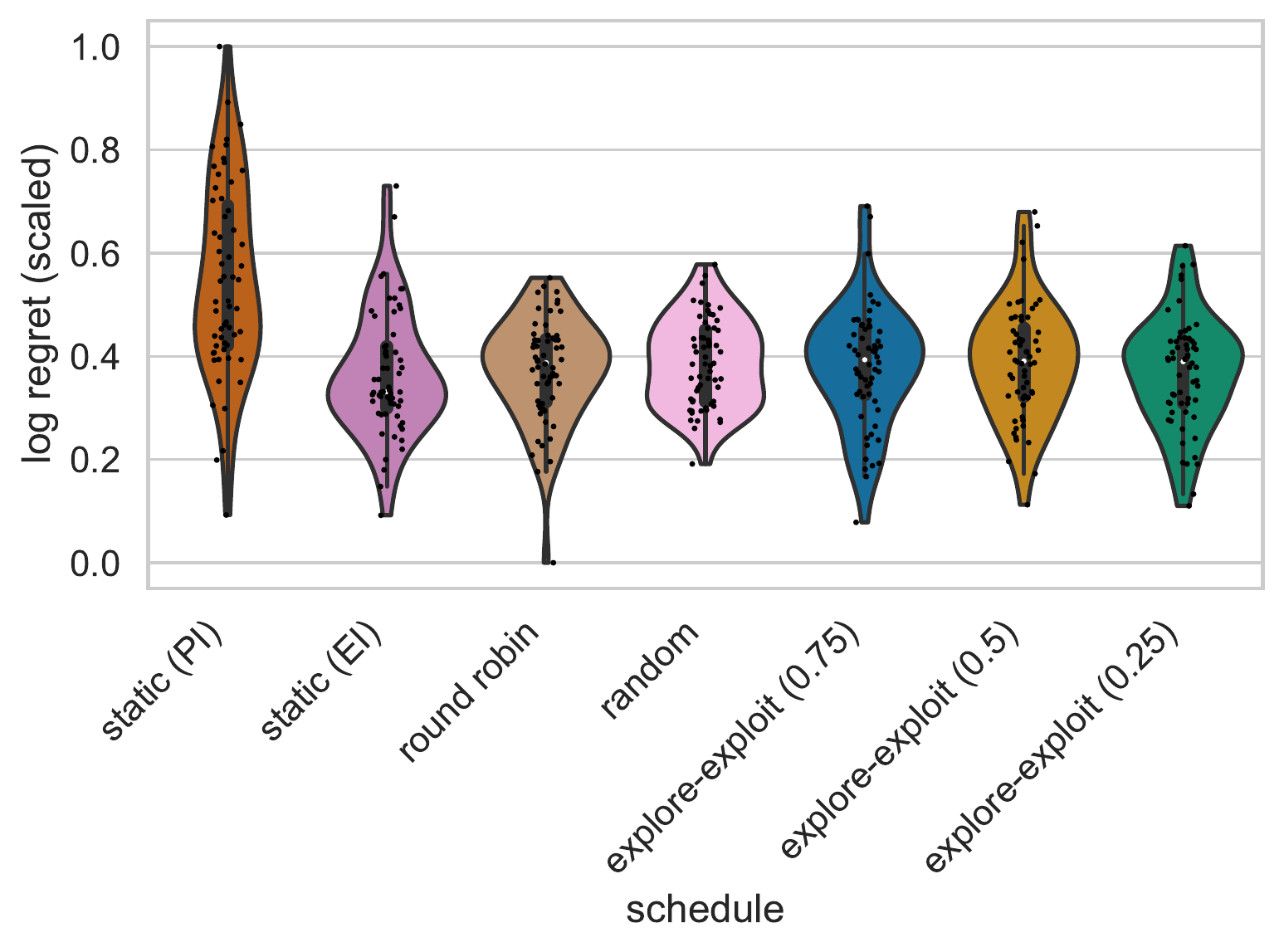}
        \caption{Final Log Regret (Scaled)}
        \label{subfig:boxplot_5}
    \end{subfigure}
    \hfill
    \begin{subfigure}[b]{0.45\textwidth}
        \centering
        \includegraphics[width=\textwidth]{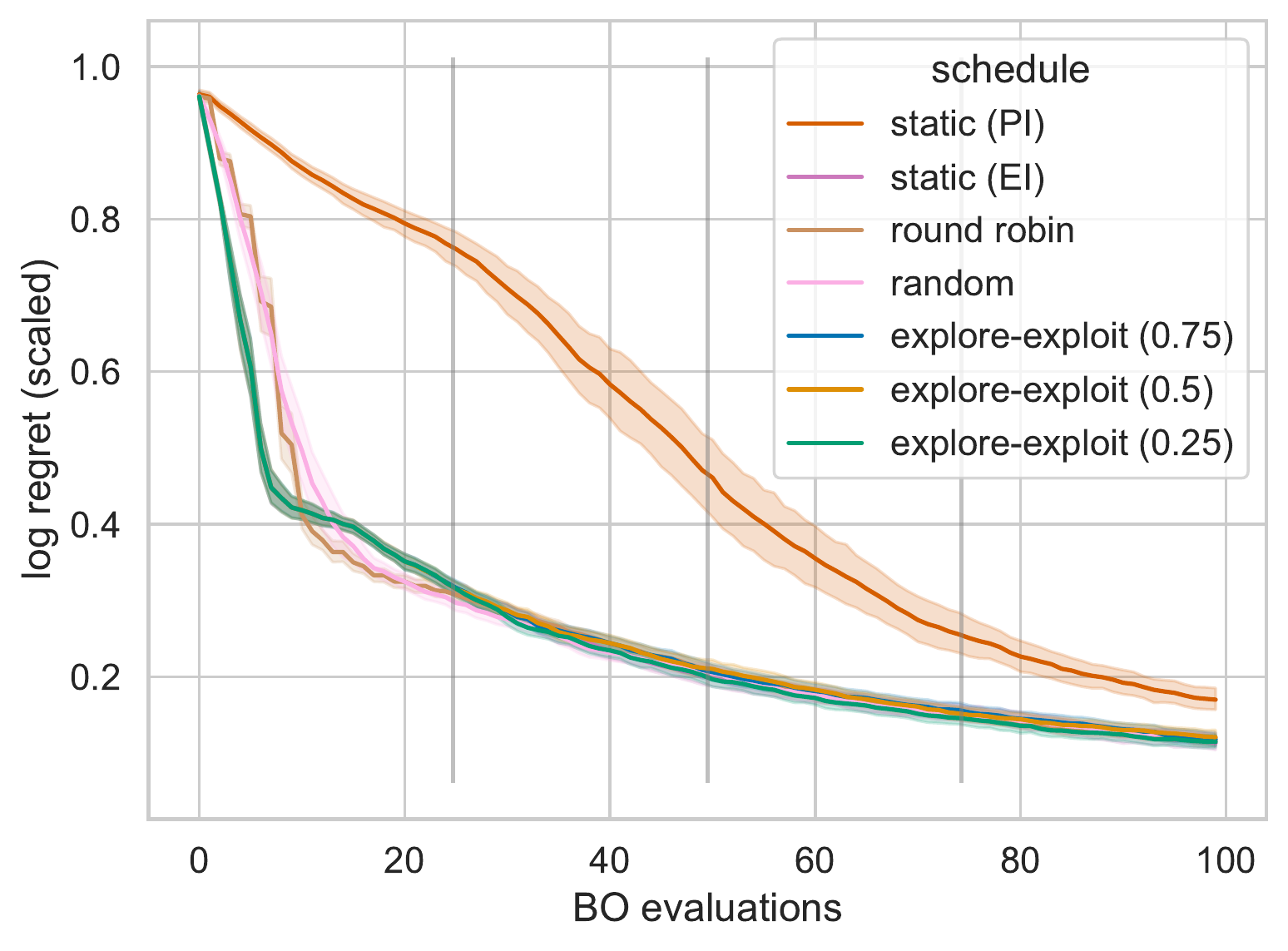}
        \caption{Log-Regret (Scaled) per Step}
        \label{subfig:convergence_5}
    \end{subfigure}\\
    \vspace*{3mm}
    \centering
    \begin{subfigure}[b]{\textwidth}
        \centering
        \includegraphics[width=\textwidth]{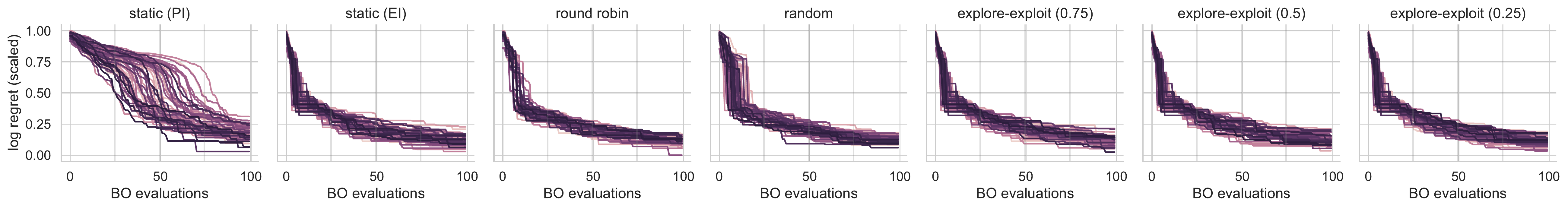}
        \caption{Log-Regret (Scaled) per Step and Seed}
        \label{subfig:convergence_perseed_5}
    \end{subfigure}
    \caption{BBOB Function 5}
    \label{fig:bbob_function_5}
\end{figure}

\begin{figure}[h]
    \centering
    \begin{subfigure}[b]{0.45\textwidth}
        \centering
        \includegraphics[width=\textwidth]{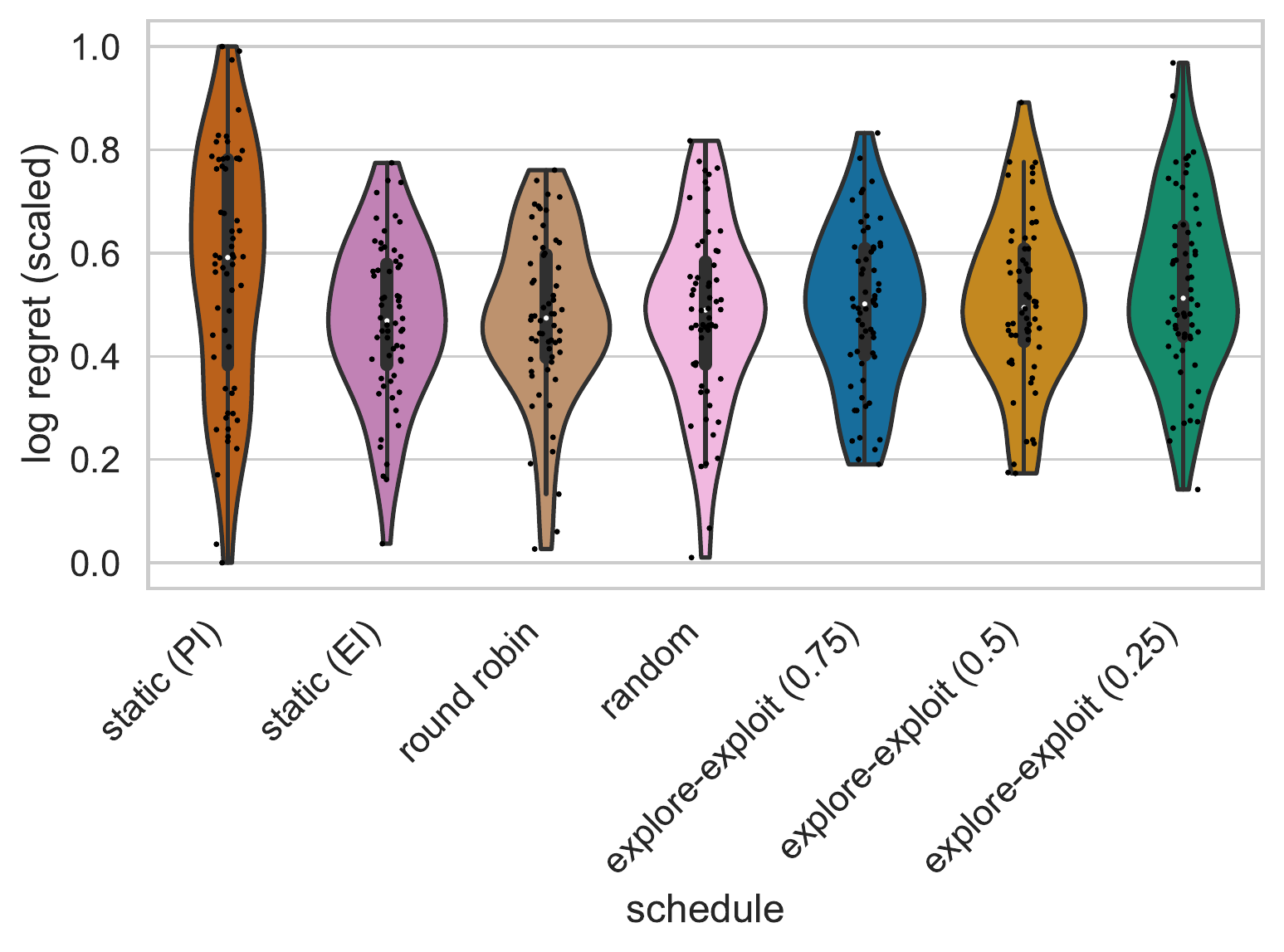}
        \caption{Final Log Regret (Scaled)}
        \label{subfig:boxplot_6}
    \end{subfigure}
    \hfill
    \begin{subfigure}[b]{0.45\textwidth}
        \centering
        \includegraphics[width=\textwidth]{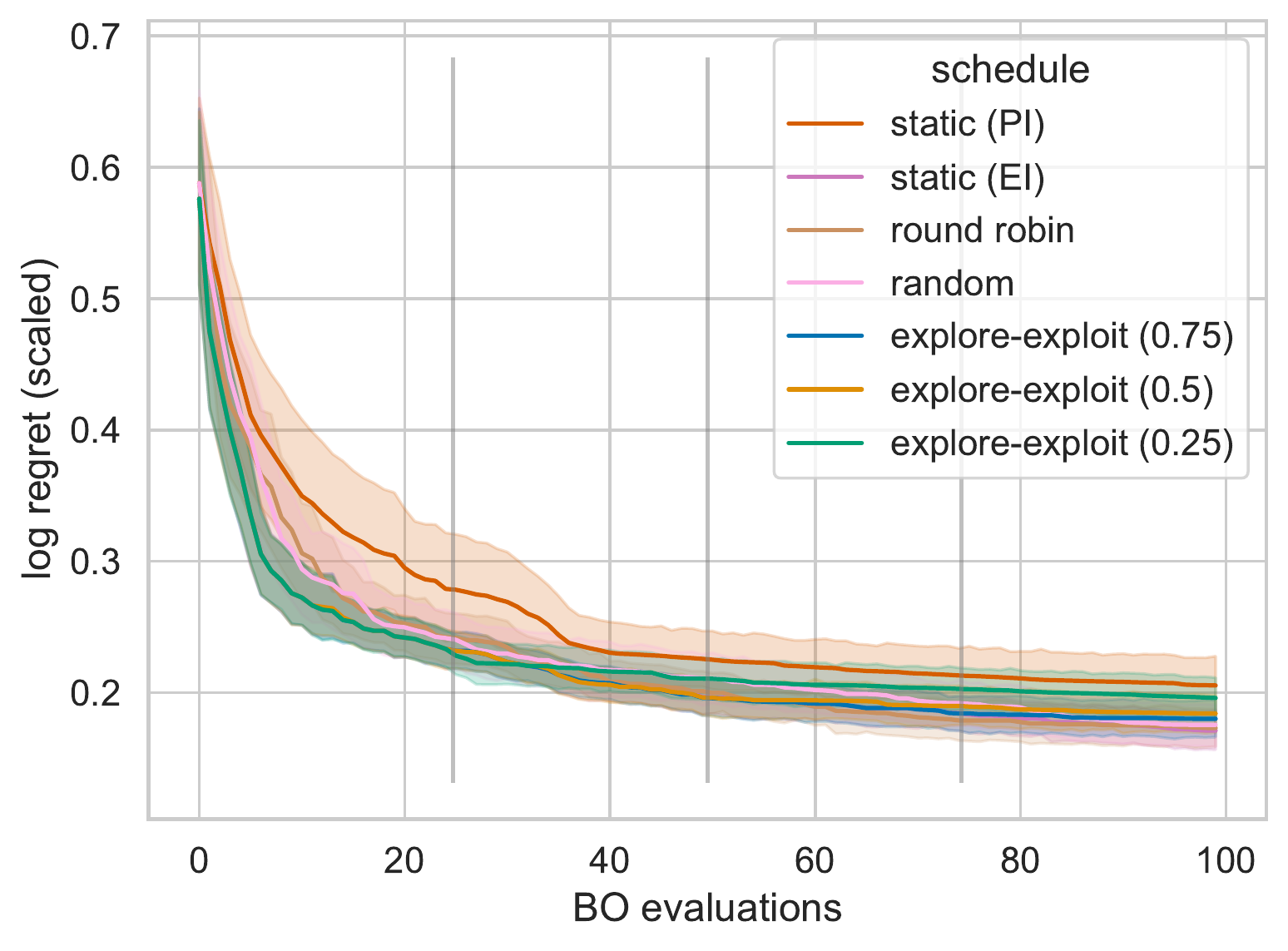}
        \caption{Log-Regret (Scaled) per Step}
        \label{subfig:convergence_6}
    \end{subfigure}\\
    \vspace*{3mm}
    \centering
    \begin{subfigure}[b]{\textwidth}
        \centering
        \includegraphics[width=\textwidth]{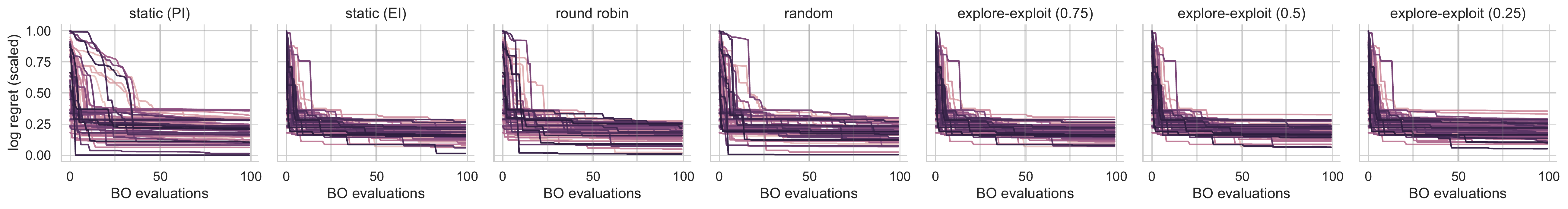}
        \caption{Log-Regret (Scaled) per Step and Seed}
        \label{subfig:convergence_perseed_6}
    \end{subfigure}
    \caption{BBOB Function 6}
    \label{fig:bbob_function_6}
\end{figure}

\begin{figure}[h]
    \centering
    \begin{subfigure}[b]{0.45\textwidth}
        \centering
        \includegraphics[width=\textwidth]{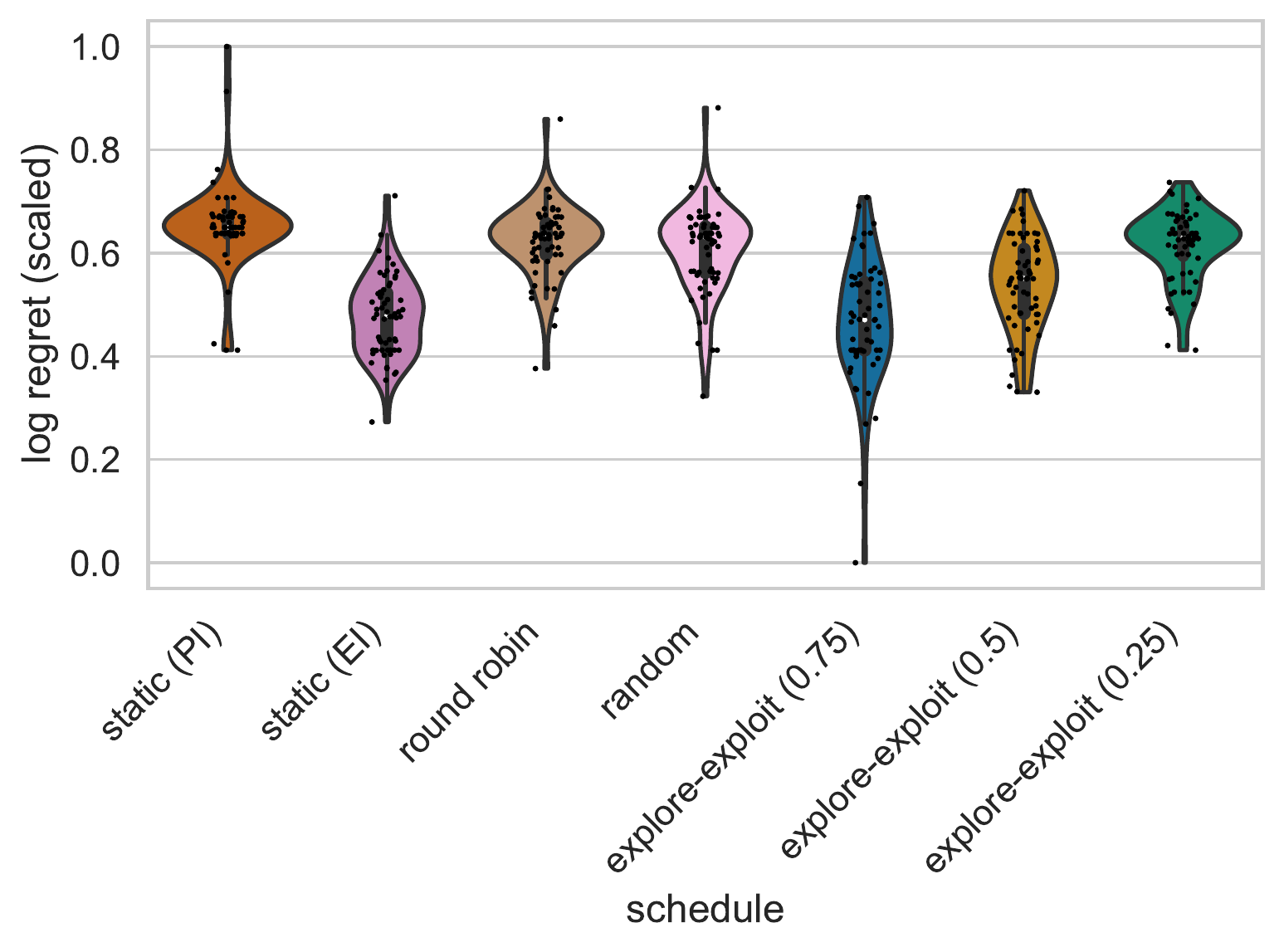}
        \caption{Final Log Regret (Scaled)}
        \label{subfig:boxplot_7}
    \end{subfigure}
    \hfill
    \begin{subfigure}[b]{0.45\textwidth}
        \centering
        \includegraphics[width=\textwidth]{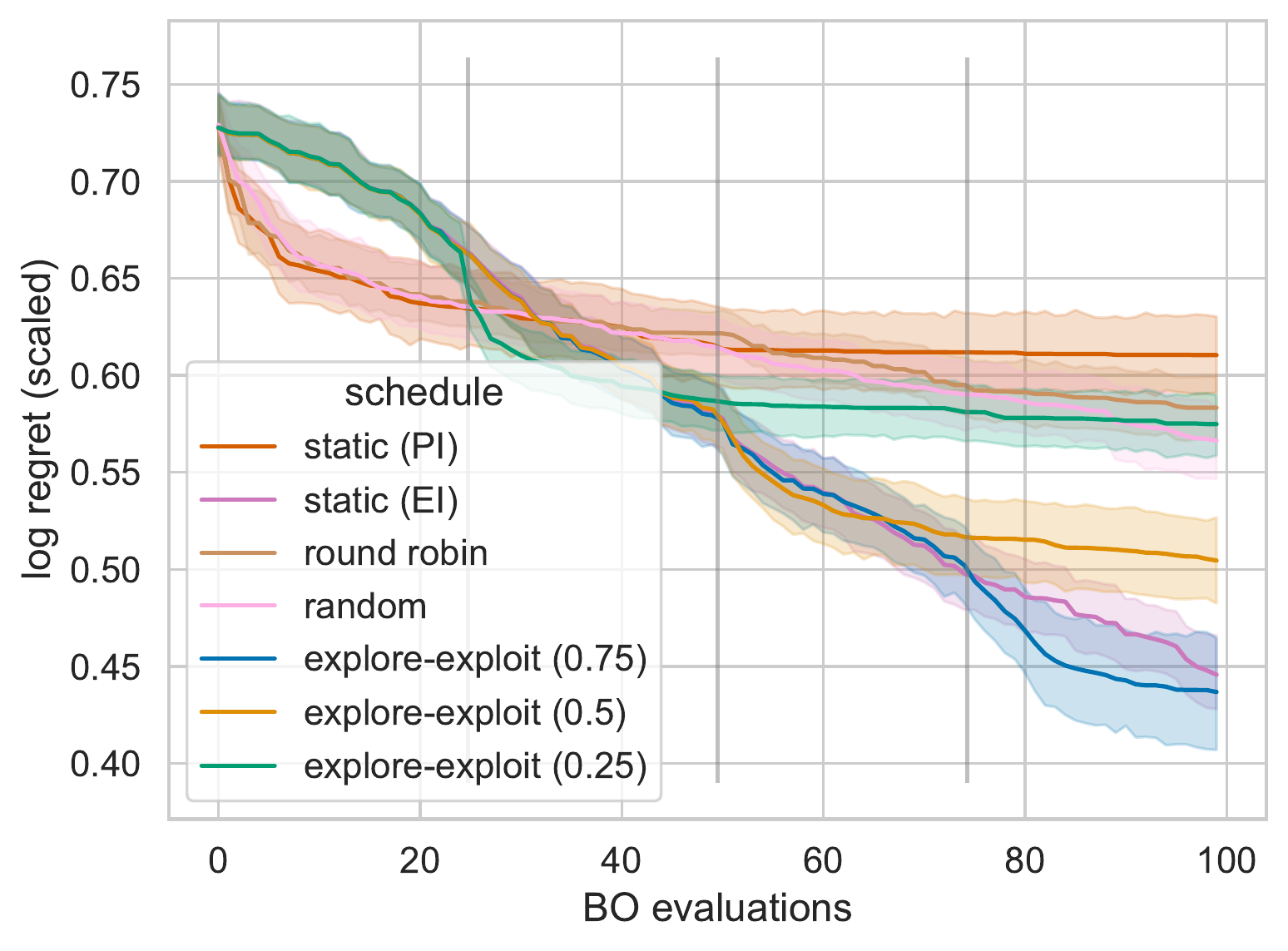}
        \caption{Log-Regret (Scaled) per Step}
        \label{subfig:convergence_7}
    \end{subfigure}\\
    \vspace*{3mm}
    \centering
    \begin{subfigure}[b]{\textwidth}
        \centering
        \includegraphics[width=\textwidth]{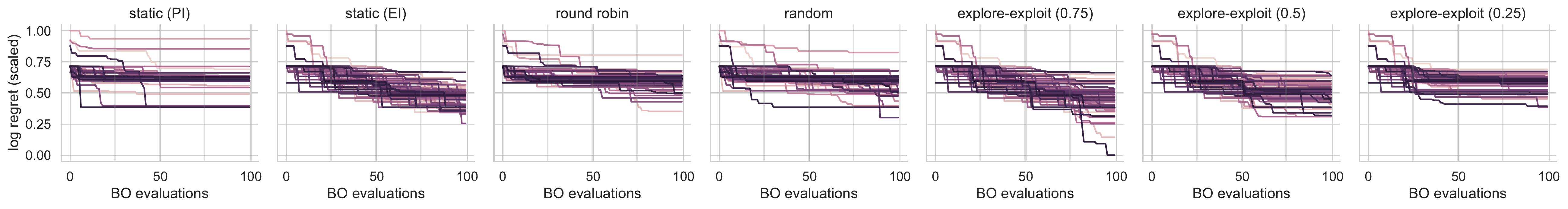}
        \caption{Log-Regret (Scaled) per Step and Seed}
        \label{subfig:convergence_perseed_7}
    \end{subfigure}
    \caption{BBOB Function 7}
    \label{fig:bbob_function_7}
\end{figure}

\begin{figure}[h]
    \centering
    \begin{subfigure}[b]{0.45\textwidth}
        \centering
        \includegraphics[width=\textwidth]{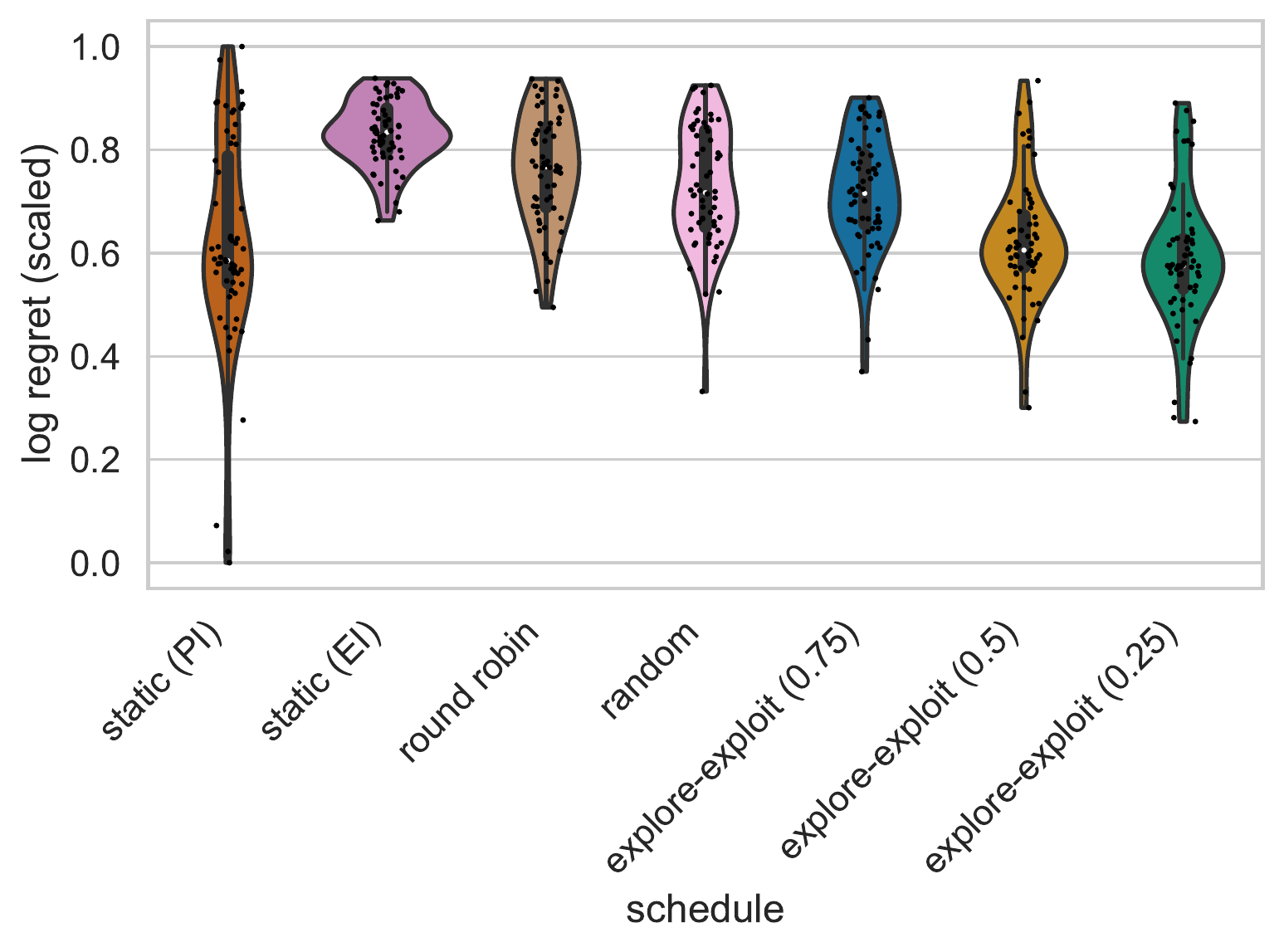}
        \caption{Final Log Regret (Scaled)}
        \label{subfig:boxplot_8}
    \end{subfigure}
    \hfill
    \begin{subfigure}[b]{0.45\textwidth}
        \centering
        \includegraphics[width=\textwidth]{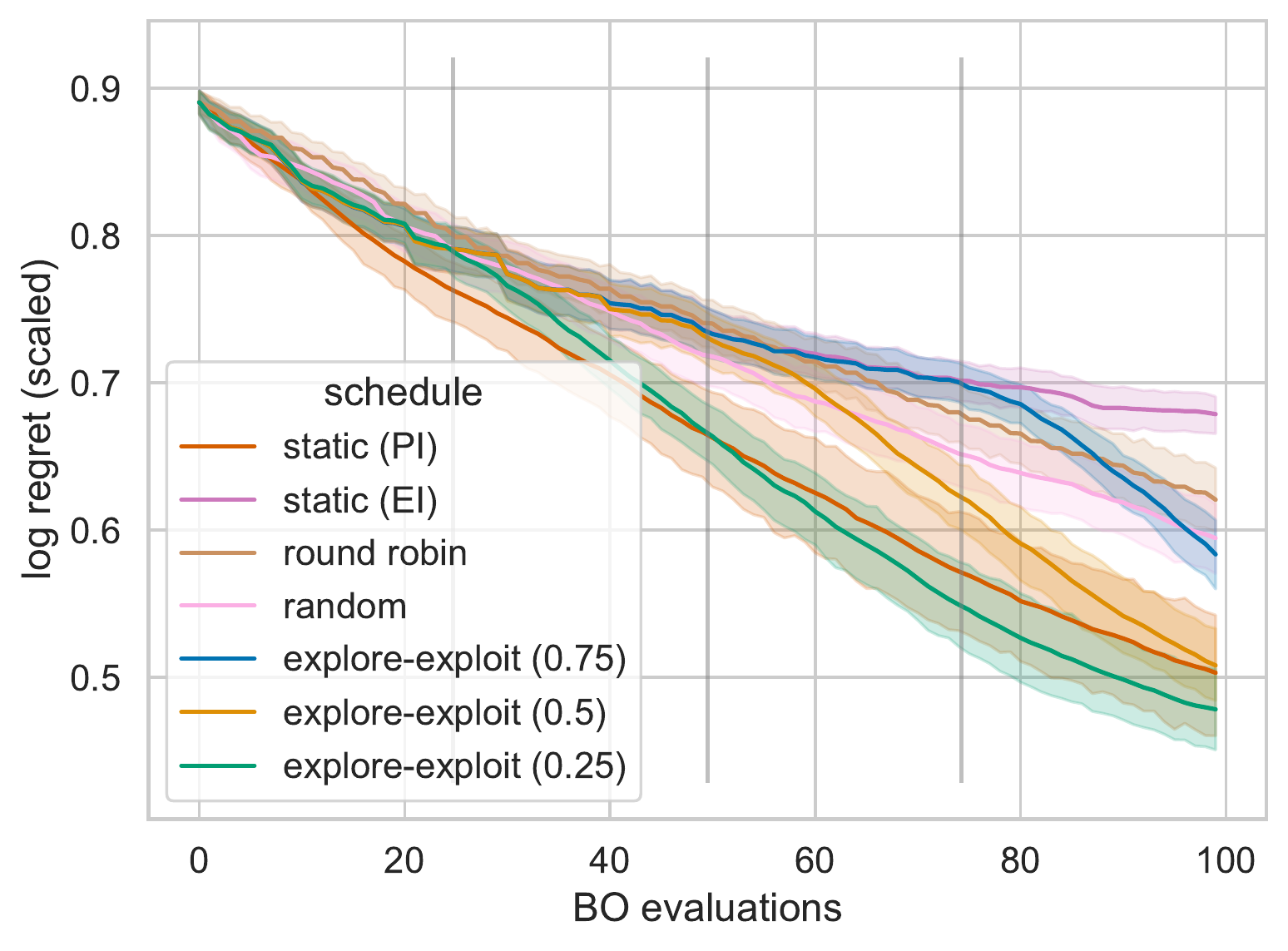}
        \caption{Log-Regret (Scaled) per Step}
        \label{subfig:convergence_8}
    \end{subfigure}\\
    \vspace*{3mm}
    \centering
    \begin{subfigure}[b]{\textwidth}
        \centering
        \includegraphics[width=\textwidth]{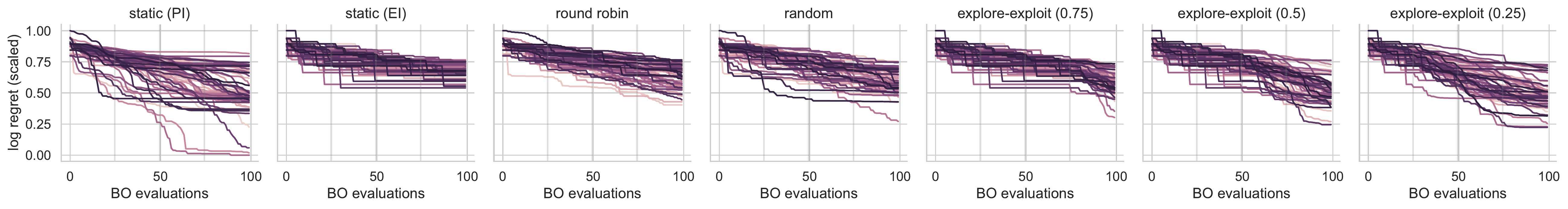}
        \caption{Log-Regret (Scaled) per Step and Seed}
        \label{subfig:convergence_perseed_8}
    \end{subfigure}
    \caption{BBOB Function 8}
    \label{fig:bbob_function_8}
\end{figure}

\begin{figure}[h]
    \centering
    \begin{subfigure}[b]{0.45\textwidth}
        \centering
        \includegraphics[width=\textwidth]{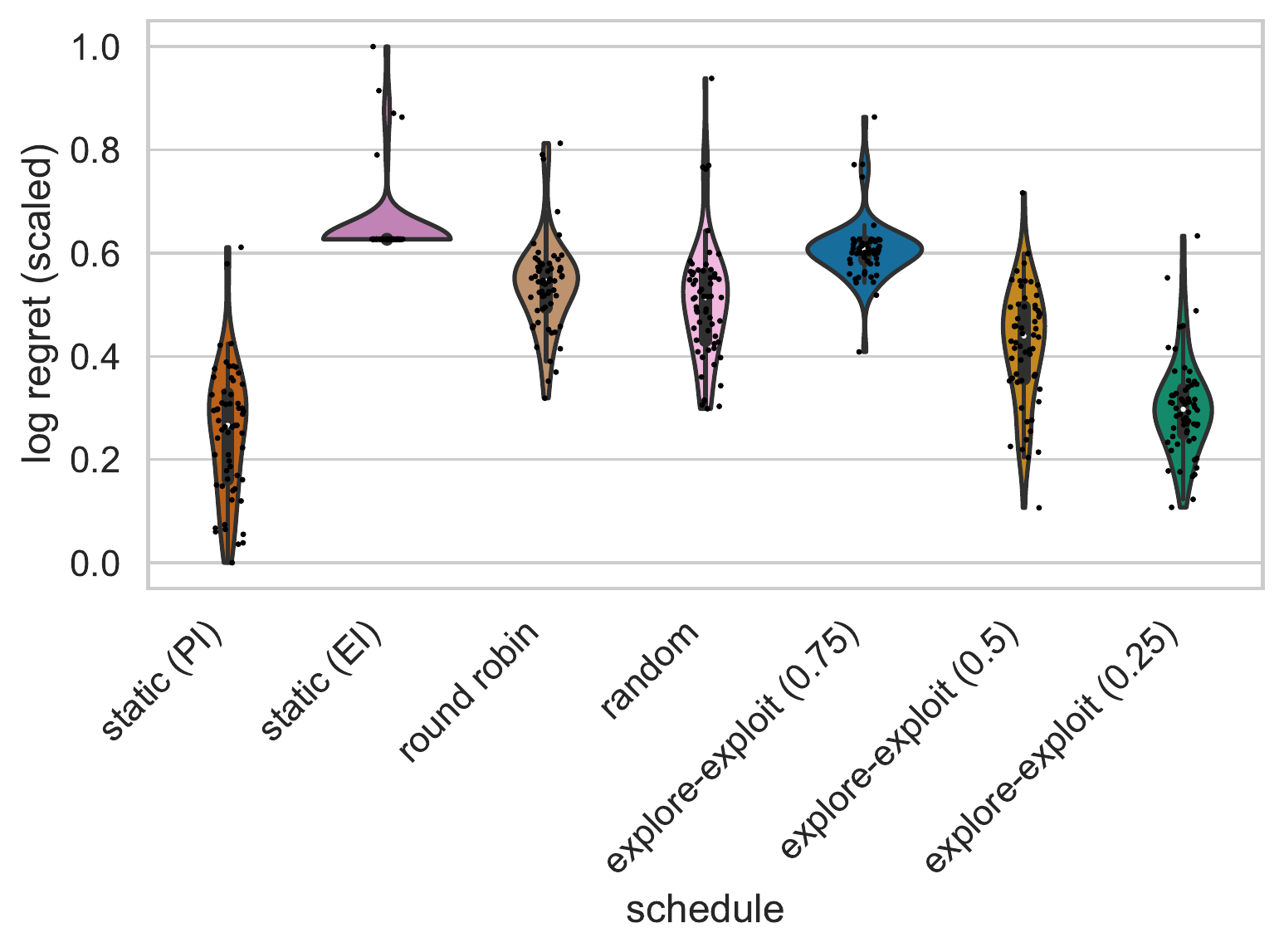}
        \caption{Final Log Regret (Scaled)}
        \label{subfig:boxplot_9}
    \end{subfigure}
    \hfill
    \begin{subfigure}[b]{0.45\textwidth}
        \centering
        \includegraphics[width=\textwidth]{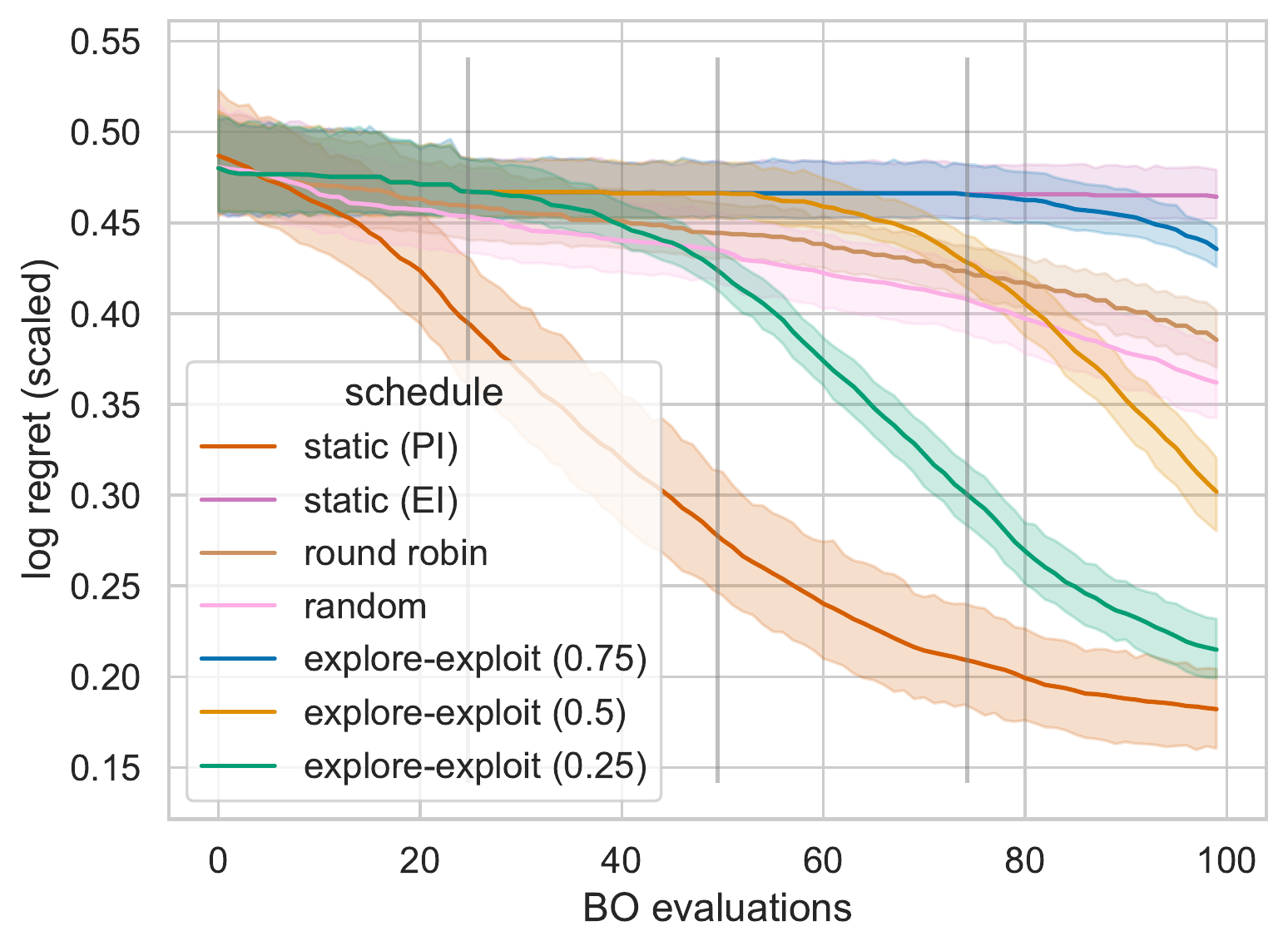}
        \caption{Log-Regret (Scaled) per Step}
        \label{subfig:convergence_9}
    \end{subfigure}\\
    \vspace*{3mm}
    \centering
    \begin{subfigure}[b]{\textwidth}
        \centering
        \includegraphics[width=\textwidth]{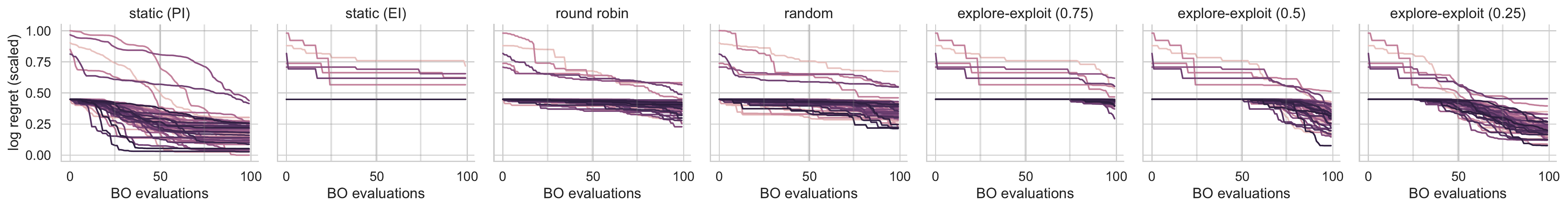}
        \caption{Log-Regret (Scaled) per Step and Seed}
        \label{subfig:convergence_perseed_9}
    \end{subfigure}
    \caption{BBOB Function 9}
    \label{fig:bbob_function_9}
\end{figure}

\begin{figure}[h]
    \centering
    \begin{subfigure}[b]{0.45\textwidth}
        \centering
        \includegraphics[width=\textwidth]{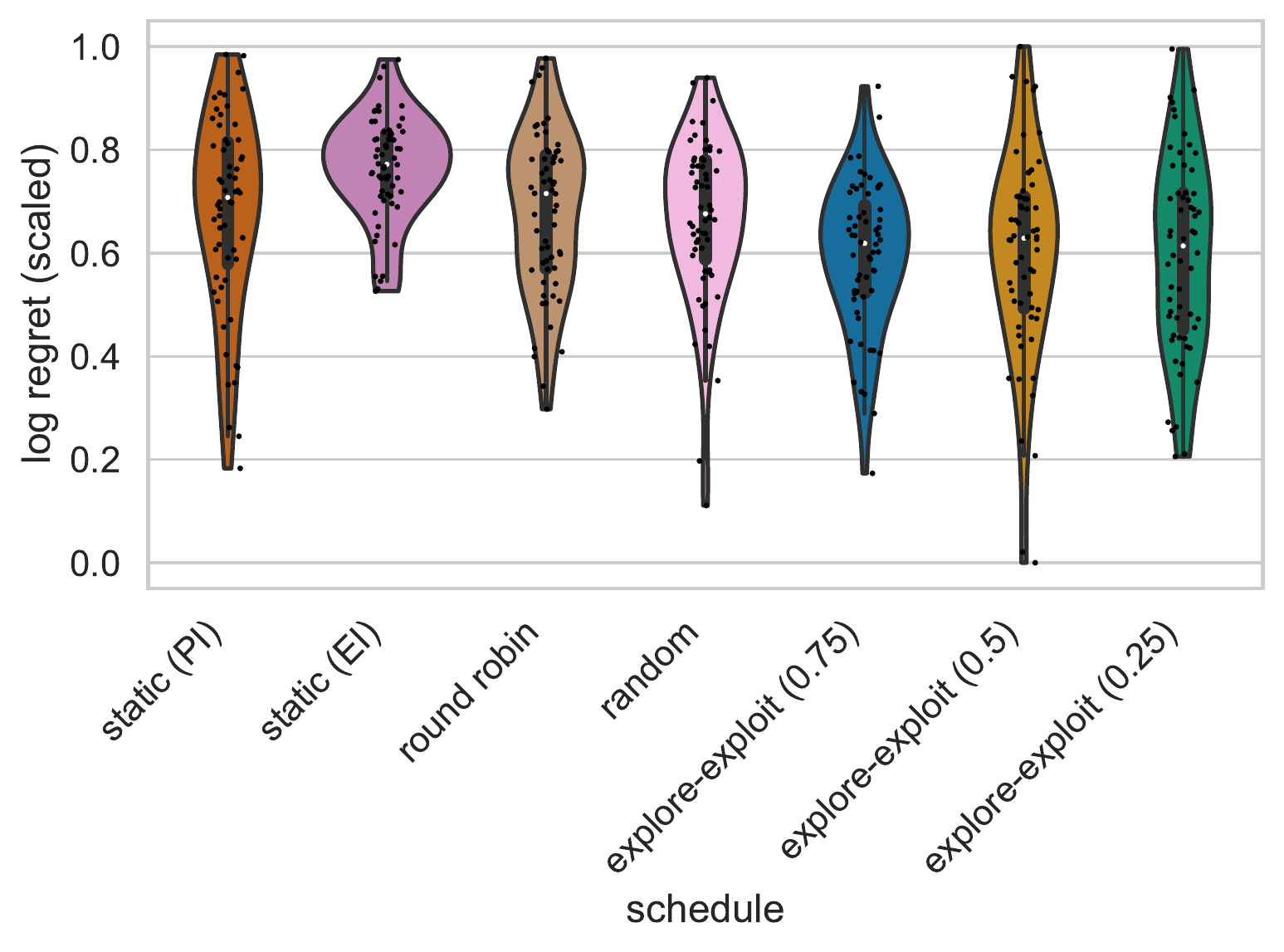}
        \caption{Final Log Regret (Scaled)}
        \label{subfig:boxplot_10}
    \end{subfigure}
    \hfill
    \begin{subfigure}[b]{0.45\textwidth}
        \centering
        \includegraphics[width=\textwidth]{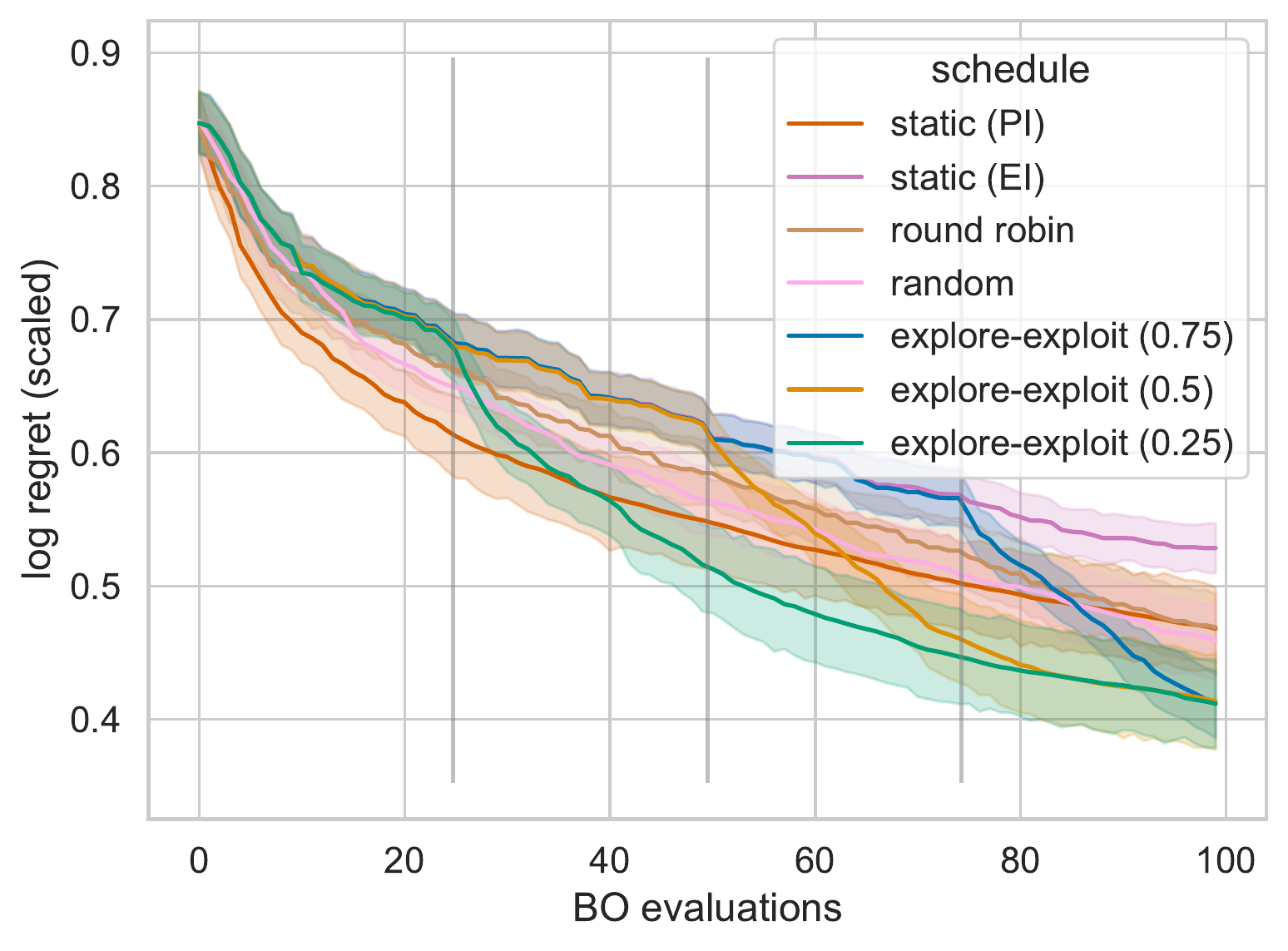}
        \caption{Log-Regret (Scaled) per Step}
        \label{subfig:convergence_10}
    \end{subfigure}\\
    \vspace*{3mm}
    \centering
    \begin{subfigure}[b]{\textwidth}
        \centering
        \includegraphics[width=\textwidth]{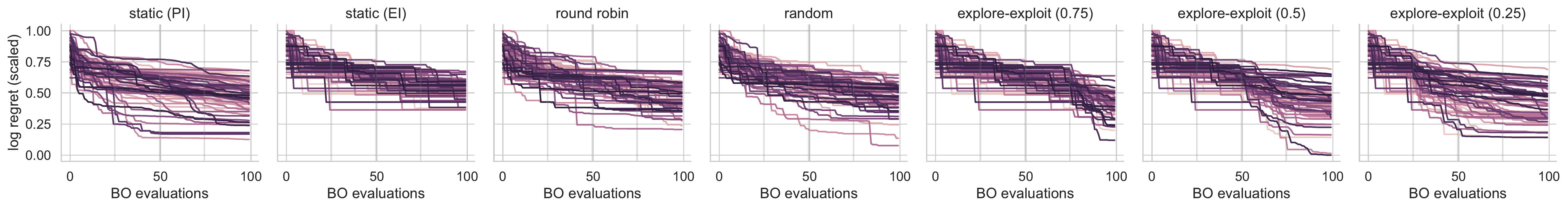}
        \caption{Log-Regret (Scaled) per Step and Seed}
        \label{subfig:convergence_perseed_10}
    \end{subfigure}
    \caption{BBOB Function 10}
    \label{fig:bbob_function_10}
\end{figure}

\begin{figure}[h]
    \centering
    \begin{subfigure}[b]{0.45\textwidth}
        \centering
        \includegraphics[width=\textwidth]{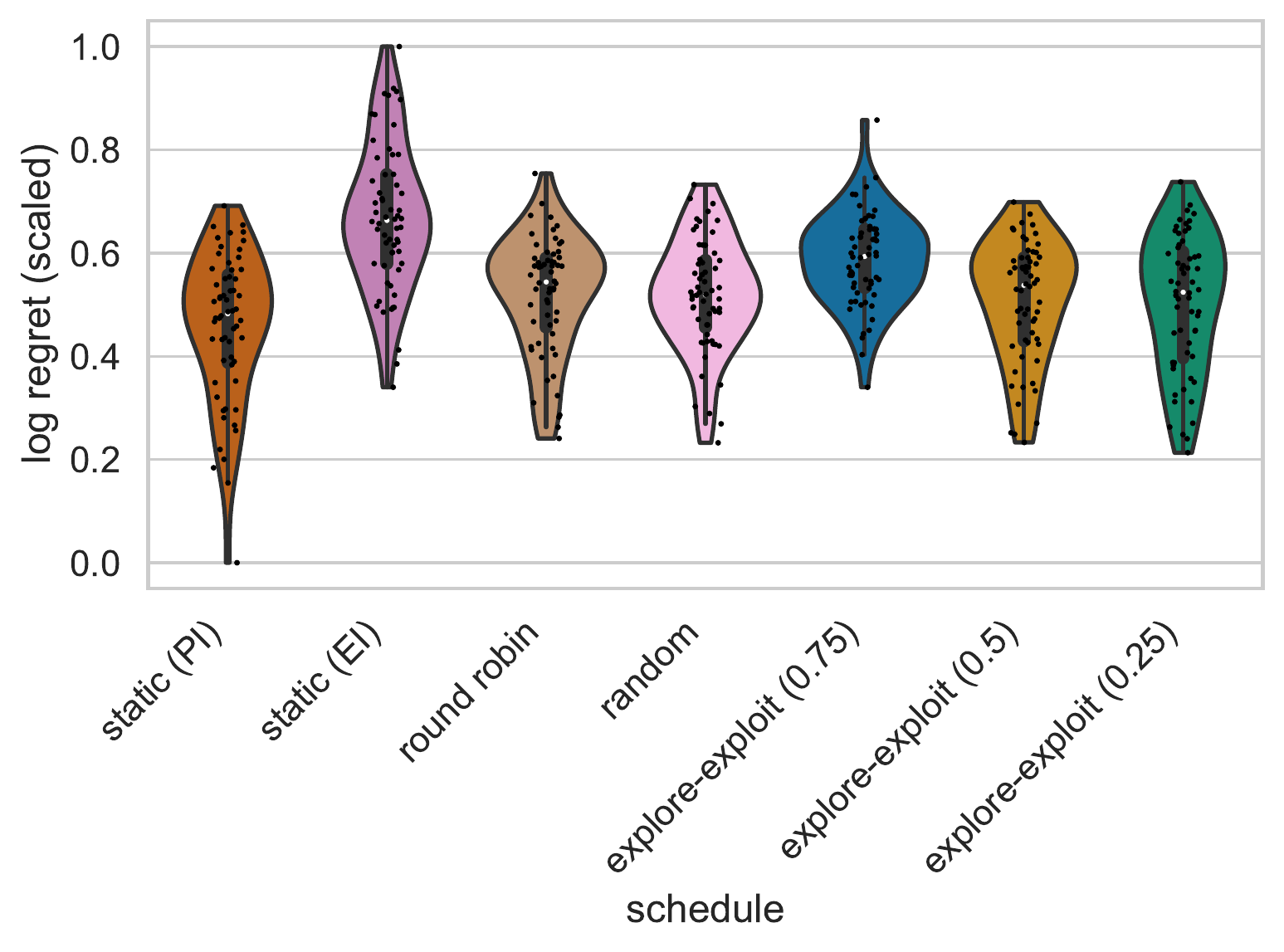}
        \caption{Final Log Regret (Scaled)}
        \label{subfig:boxplot_11}
    \end{subfigure}
    \hfill
    \begin{subfigure}[b]{0.45\textwidth}
        \centering
        \includegraphics[width=\textwidth]{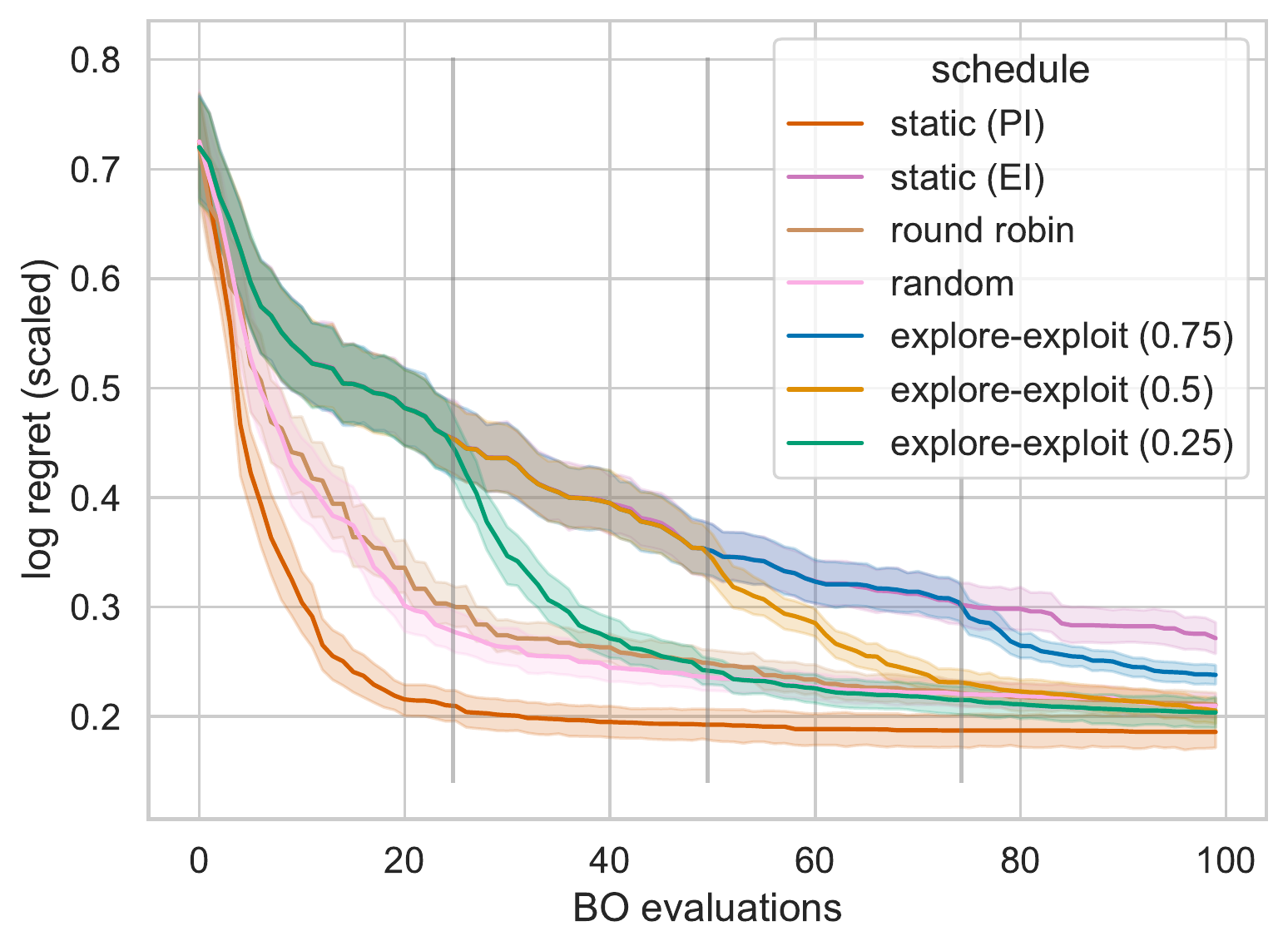}
        \caption{Log-Regret (Scaled) per Step}
        \label{subfig:convergence_11}
    \end{subfigure}\\
    \vspace*{3mm}
    \centering
    \begin{subfigure}[b]{\textwidth}
        \centering
        \includegraphics[width=\textwidth]{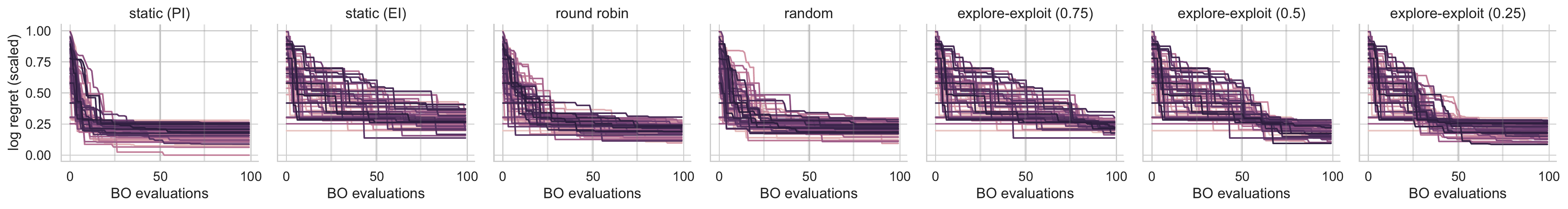}
        \caption{Log-Regret (Scaled) per Step and Seed}
        \label{subfig:convergence_perseed_11}
    \end{subfigure}
    \caption{BBOB Function 11}
    \label{fig:bbob_function_11}
\end{figure}

\begin{figure}[h]
    \centering
    \begin{subfigure}[b]{0.45\textwidth}
        \centering
        \includegraphics[width=\textwidth]{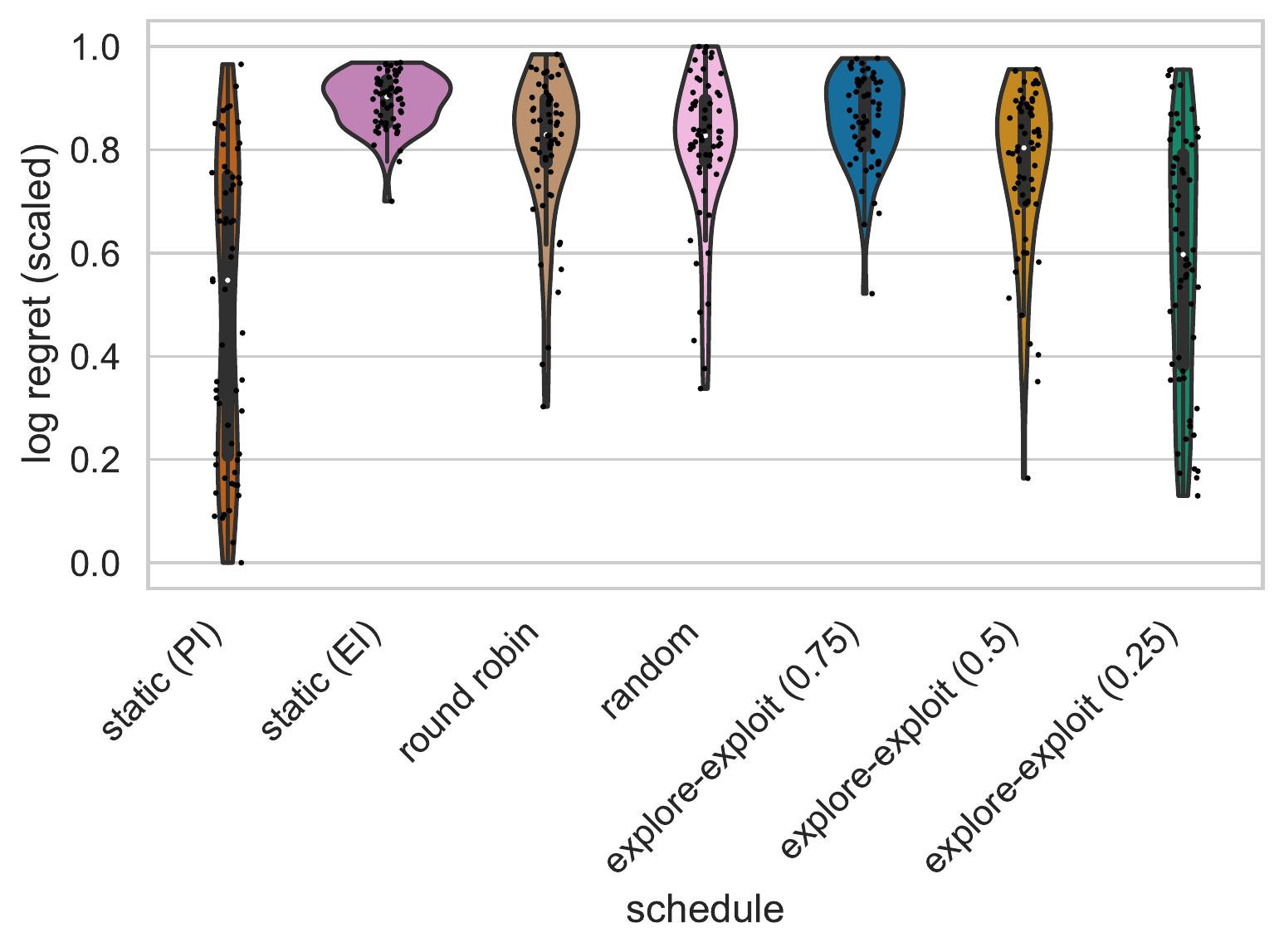}
        \caption{Final Log Regret (Scaled)}
        \label{subfig:boxplot_12}
    \end{subfigure}
    \hfill
    \begin{subfigure}[b]{0.45\textwidth}
        \centering
        \includegraphics[width=\textwidth]{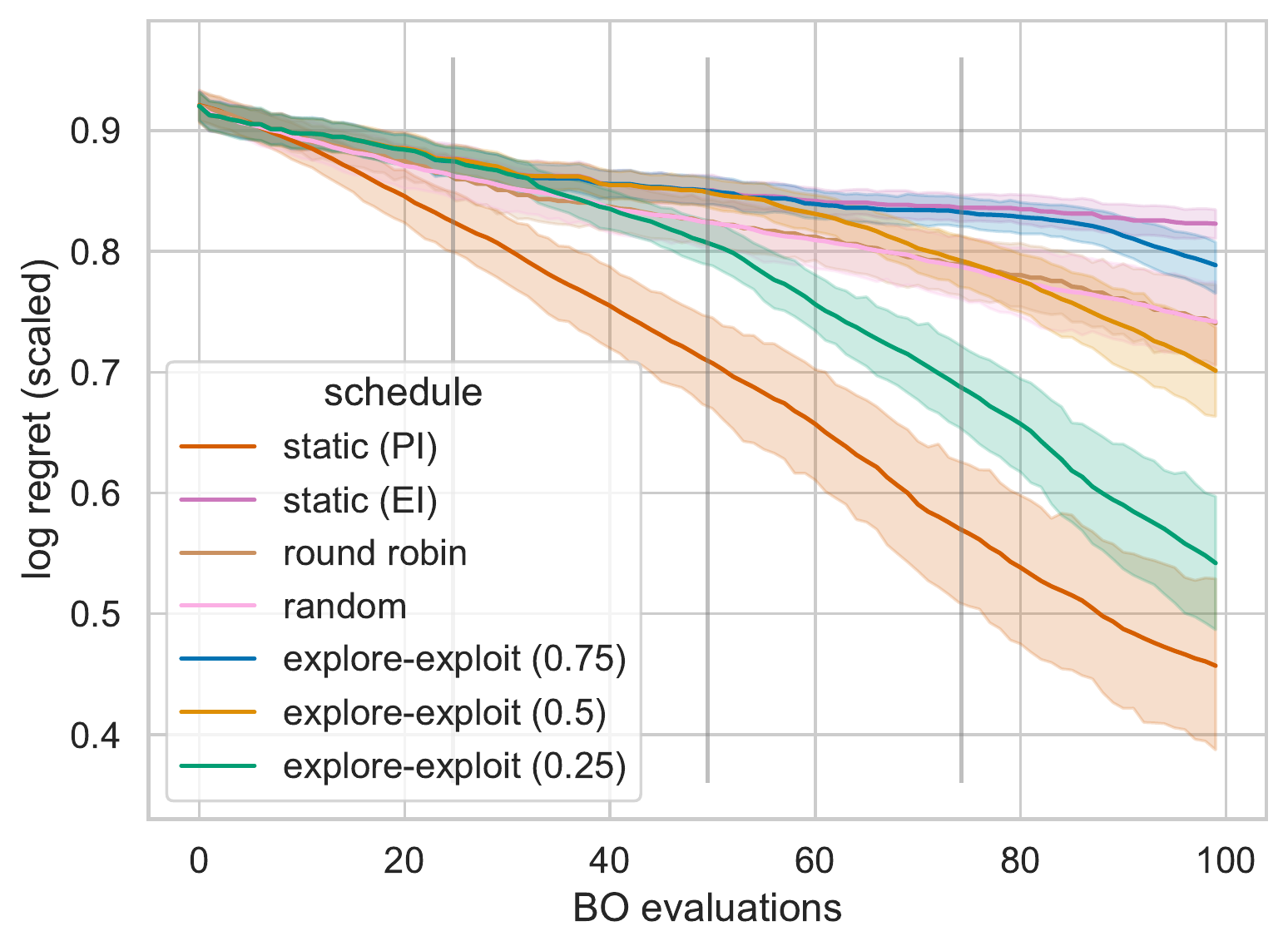}
        \caption{Log-Regret (Scaled) per Step}
        \label{subfig:convergence_12}
    \end{subfigure}\\
    \vspace*{3mm}
    \centering
    \begin{subfigure}[b]{\textwidth}
        \centering
        \includegraphics[width=\textwidth]{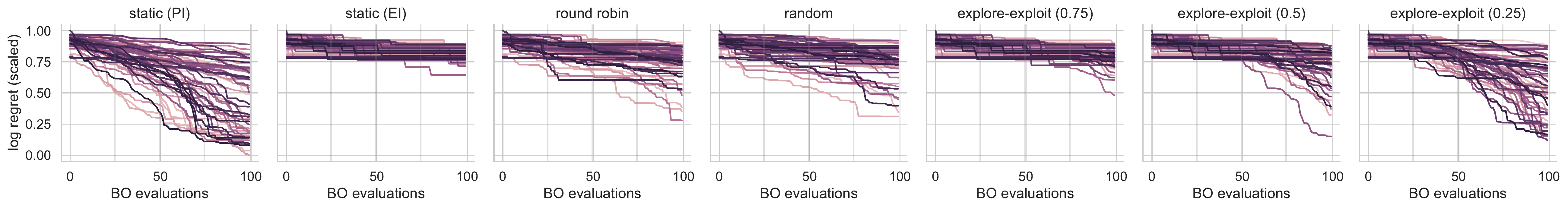}
        \caption{Log-Regret (Scaled) per Step and Seed}
        \label{subfig:convergence_perseed_12}
    \end{subfigure}
    \caption{BBOB Function 12}
    \label{fig:bbob_function_12}
\end{figure}

\begin{figure}[h]
    \centering
    \begin{subfigure}[b]{0.45\textwidth}
        \centering
        \includegraphics[width=\textwidth]{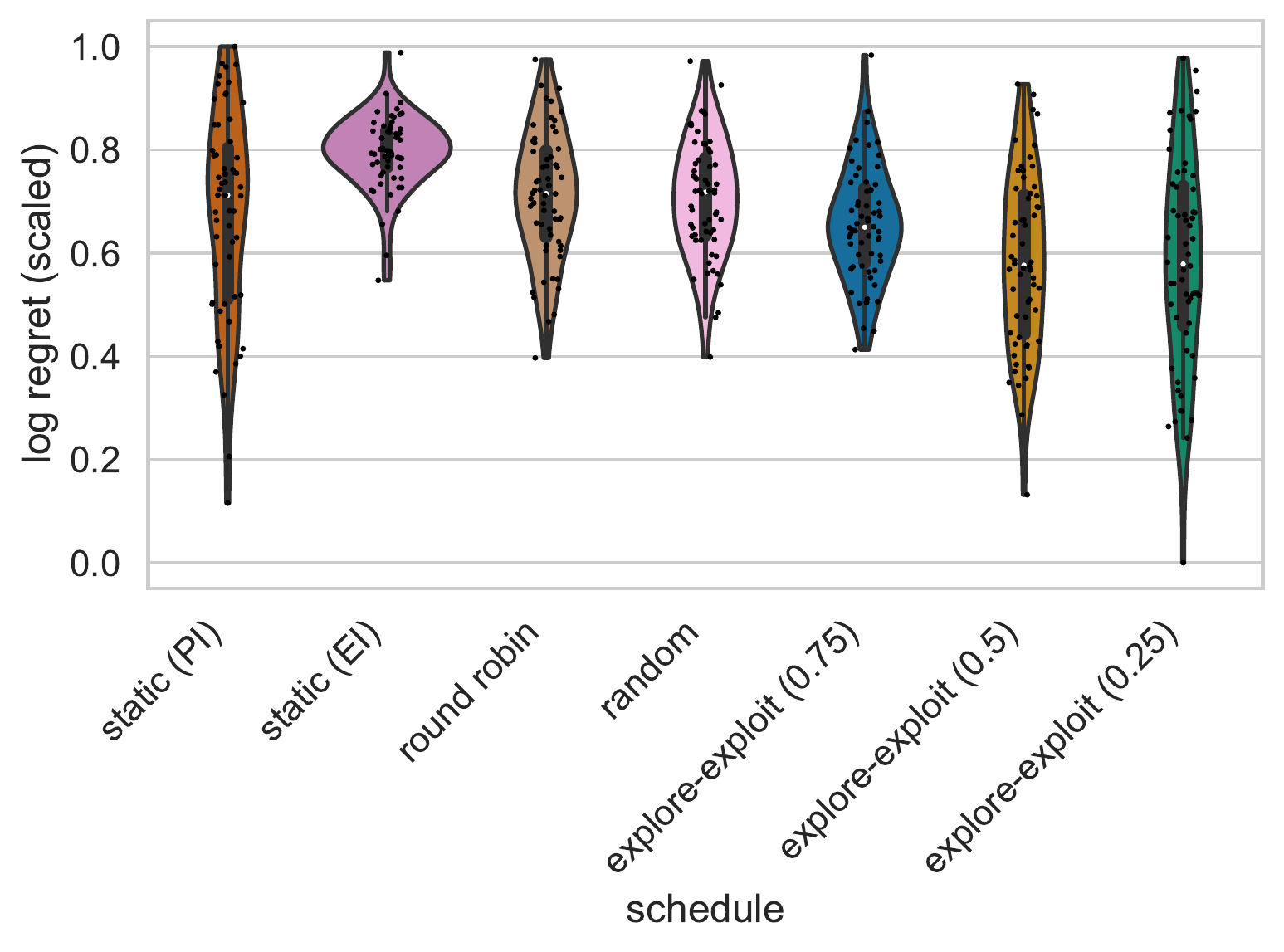}
        \caption{Final Log Regret (Scaled)}
        \label{subfig:boxplot_13}
    \end{subfigure}
    \hfill
    \begin{subfigure}[b]{0.45\textwidth}
        \centering
        \includegraphics[width=\textwidth]{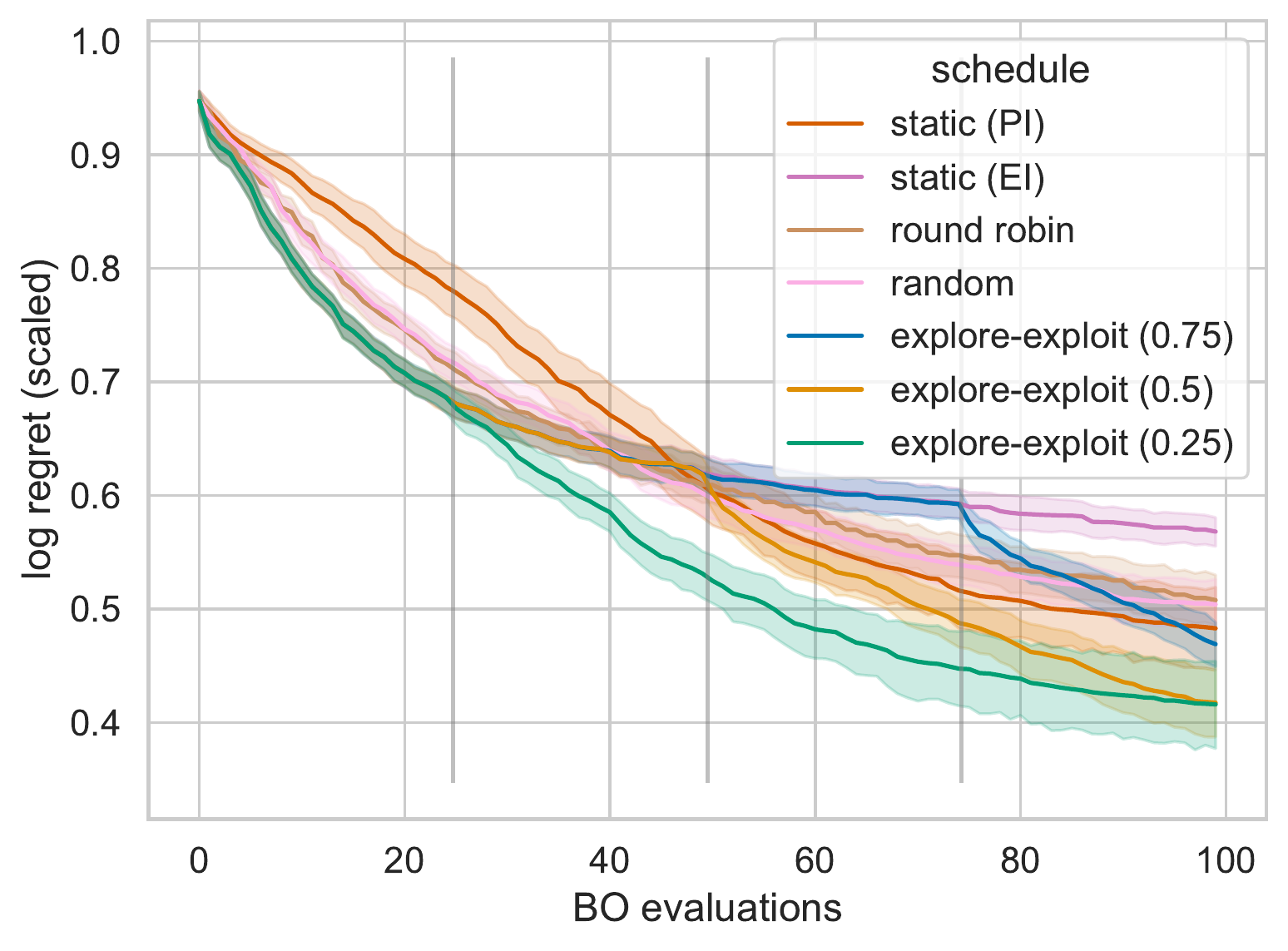}
        \caption{Log-Regret (Scaled) per Step}
        \label{subfig:convergence_13}
    \end{subfigure}\\
    \vspace*{3mm}
    \centering
    \begin{subfigure}[b]{\textwidth}
        \centering
        \includegraphics[width=\textwidth]{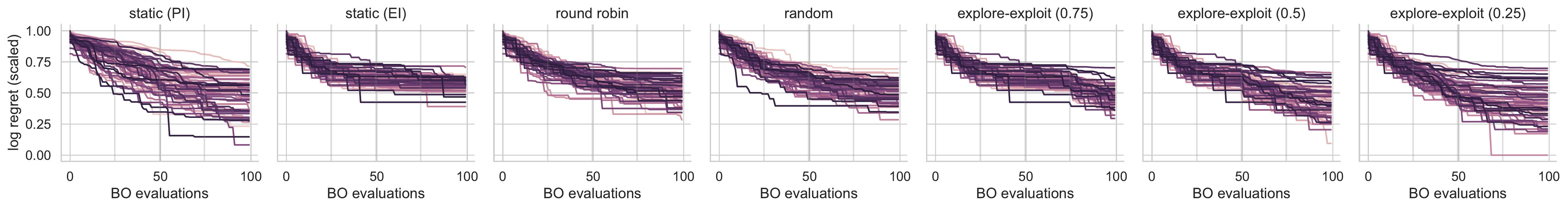}
        \caption{Log-Regret (Scaled) per Step and Seed}
        \label{subfig:convergence_perseed_13}
    \end{subfigure}
    \caption{BBOB Function 13}
    \label{fig:bbob_function_13}
\end{figure}

\begin{figure}[h]
    \centering
    \begin{subfigure}[b]{0.45\textwidth}
        \centering
        \includegraphics[width=\textwidth]{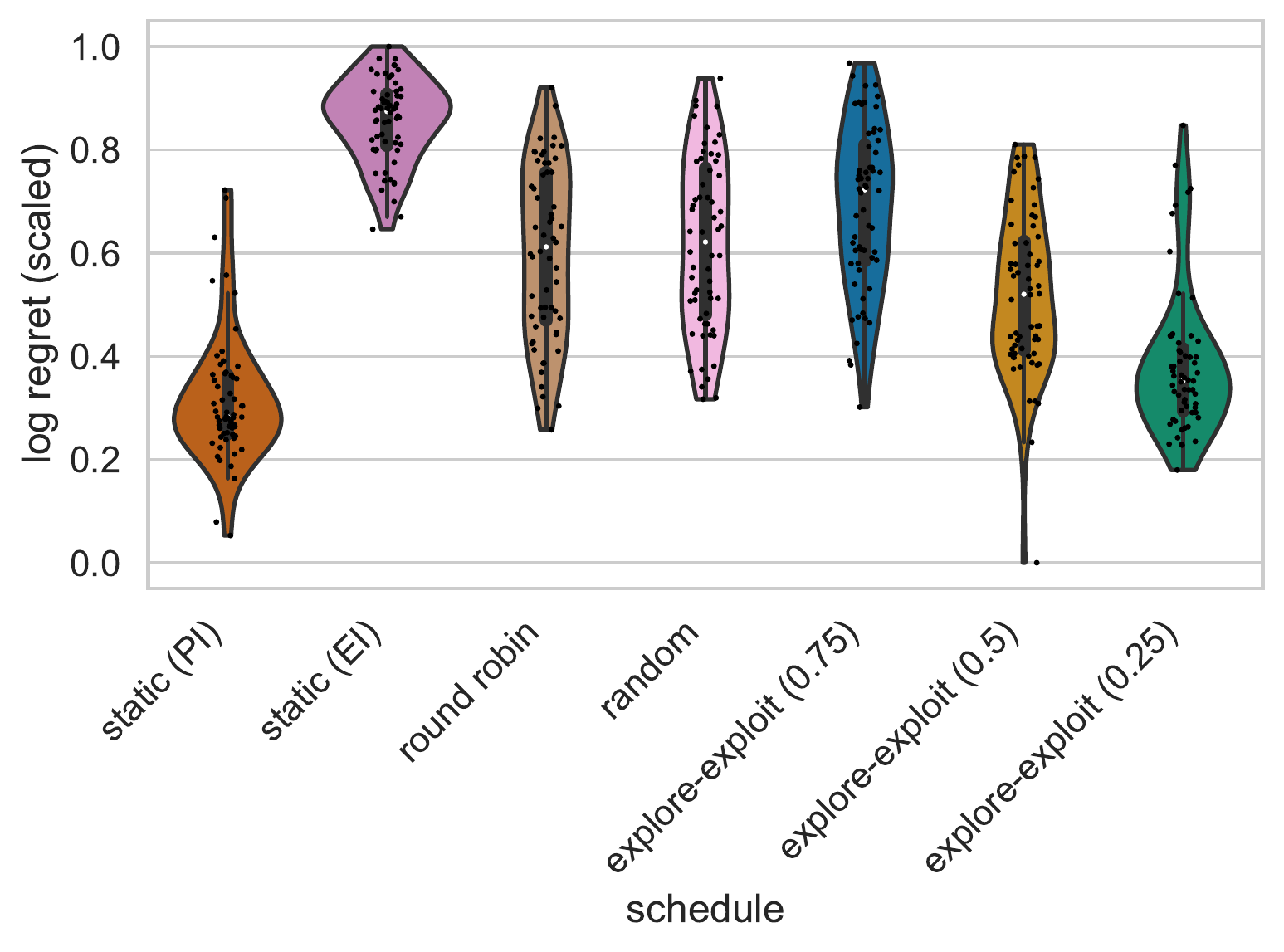}
        \caption{Final Log Regret (Scaled)}
        \label{subfig:boxplot_14}
    \end{subfigure}
    \hfill
    \begin{subfigure}[b]{0.45\textwidth}
        \centering
        \includegraphics[width=\textwidth]{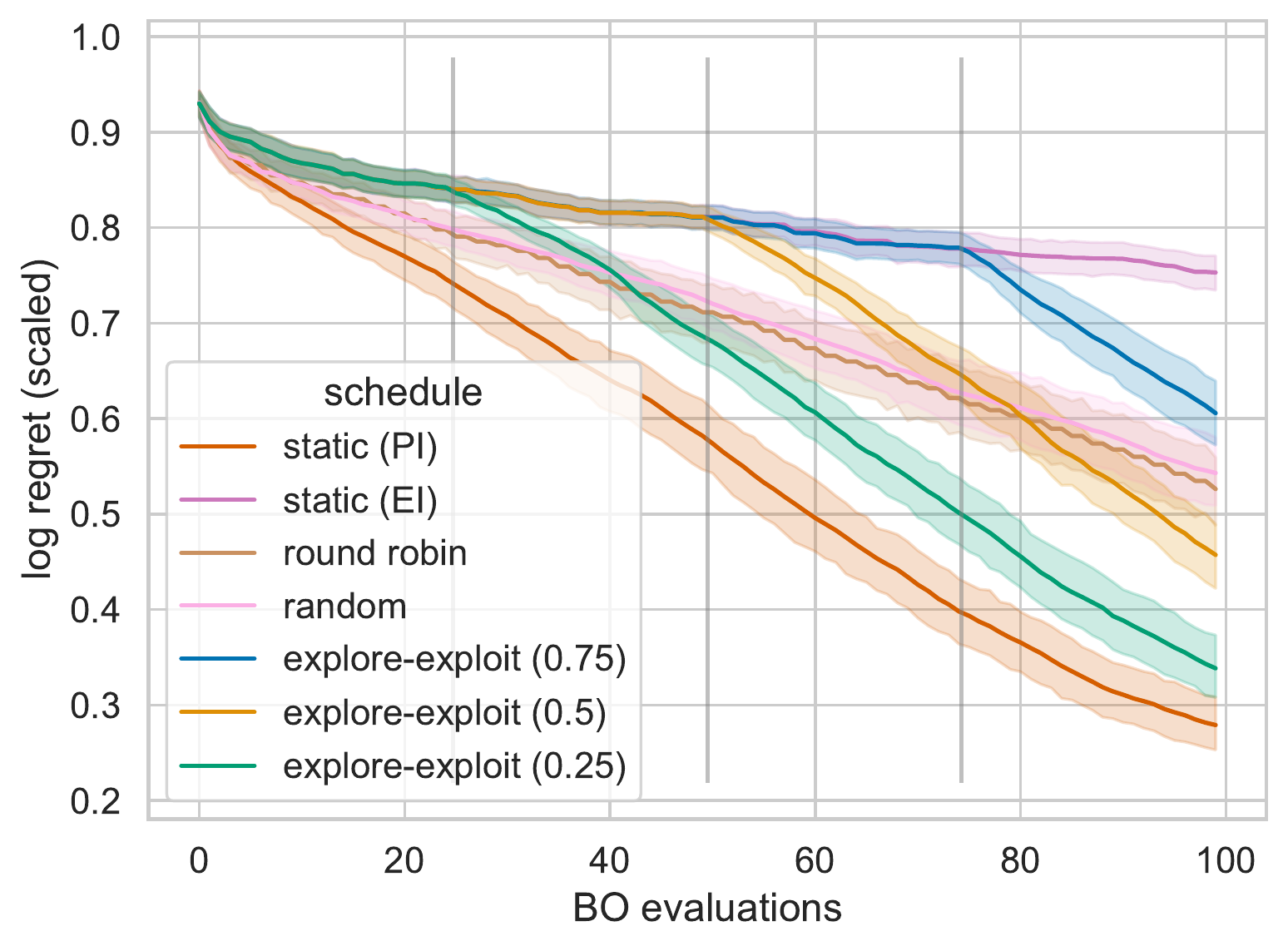}
        \caption{Log-Regret (Scaled) per Step}
        \label{subfig:convergence_14}
    \end{subfigure}\\
    \vspace*{3mm}
    \centering
    \begin{subfigure}[b]{\textwidth}
        \centering
        \includegraphics[width=\textwidth]{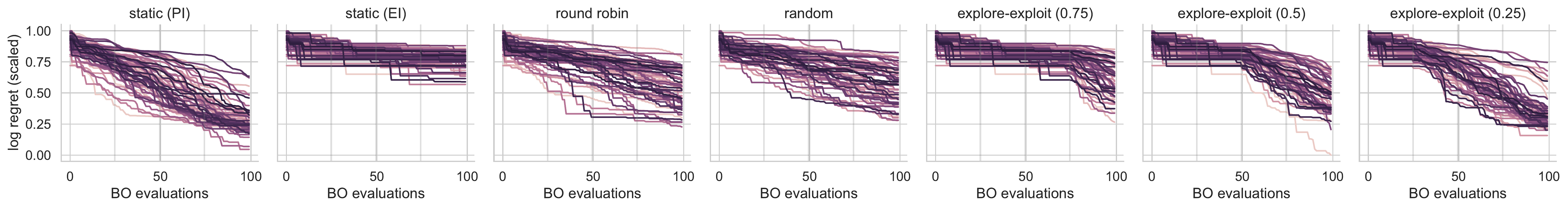}
        \caption{Log-Regret (Scaled) per Step and Seed}
        \label{subfig:convergence_perseed_14}
    \end{subfigure}
    \caption{BBOB Function 14}
    \label{fig:bbob_function_14}
\end{figure}

\begin{figure}[h]
    \centering
    \begin{subfigure}[b]{0.45\textwidth}
        \centering
        \includegraphics[width=\textwidth]{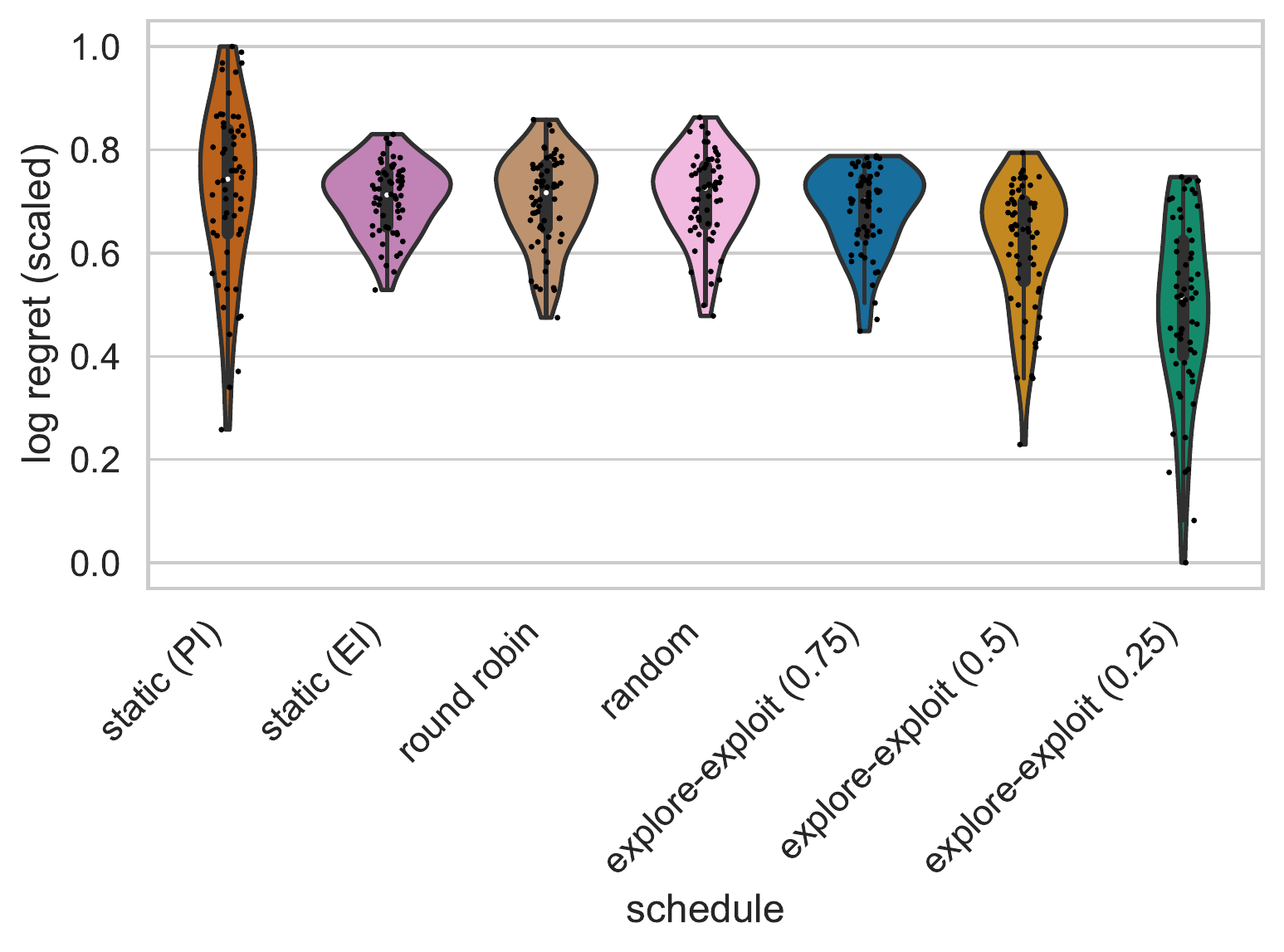}
        \caption{Final Log Regret (Scaled)}
        \label{subfig:boxplot_15}
    \end{subfigure}
    \hfill
    \begin{subfigure}[b]{0.45\textwidth}
        \centering
        \includegraphics[width=\textwidth]{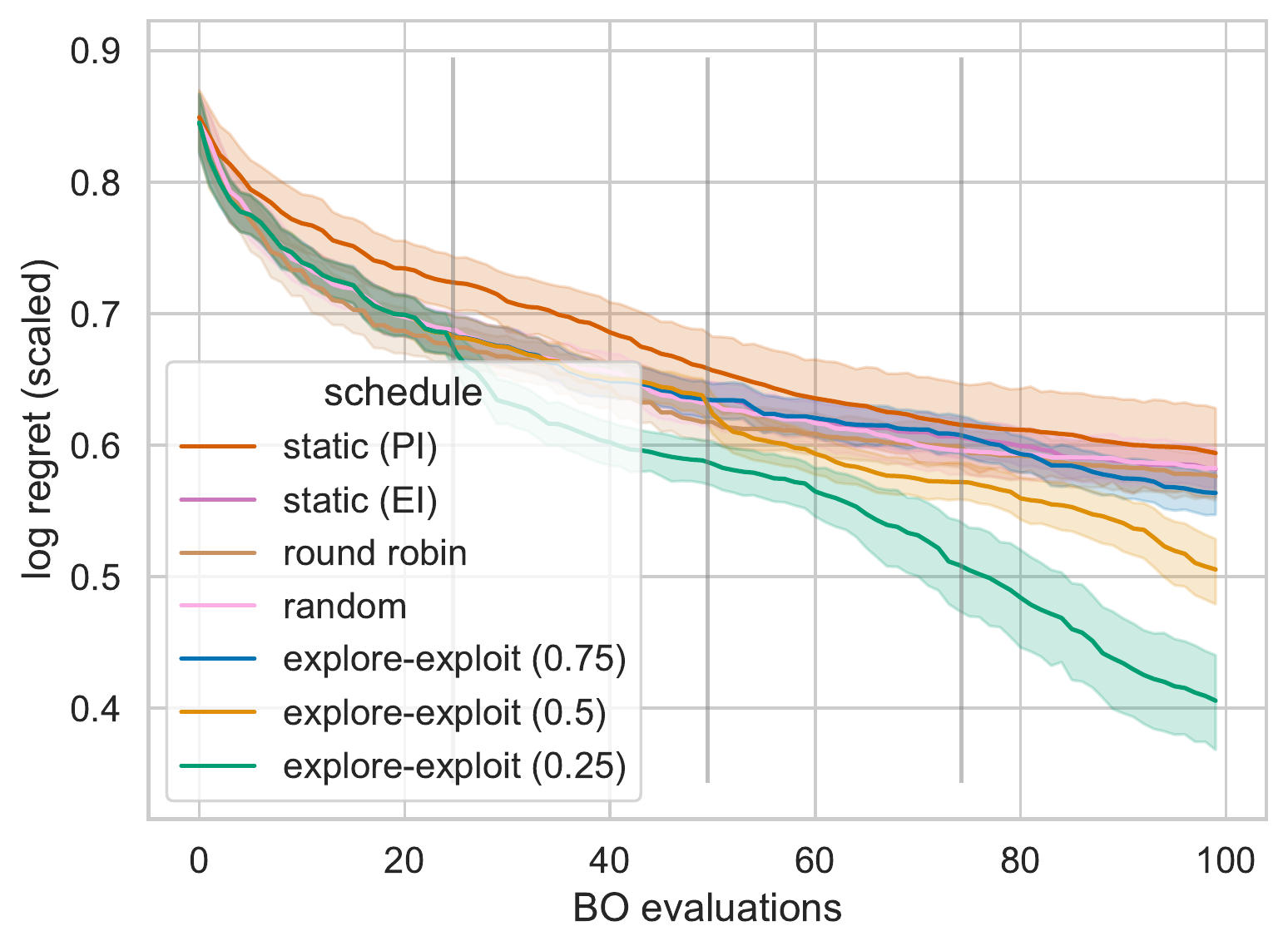}
        \caption{Log-Regret (Scaled) per Step}
        \label{subfig:convergence_15}
    \end{subfigure}\\
    \vspace*{3mm}
    \centering
    \begin{subfigure}[b]{\textwidth}
        \centering
        \includegraphics[width=\textwidth]{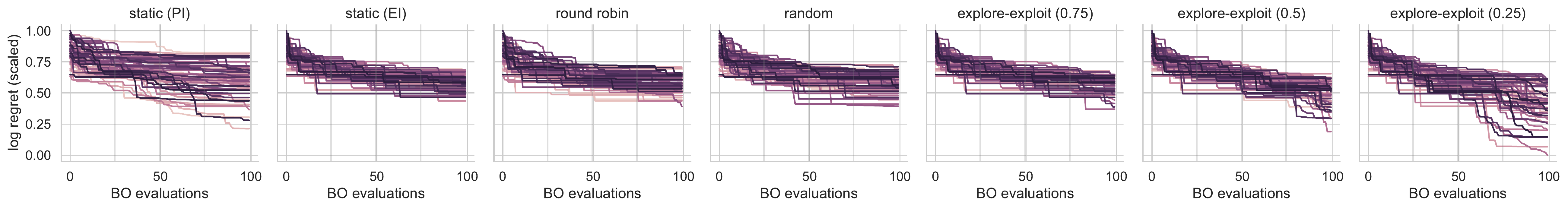}
        \caption{Log-Regret (Scaled) per Step and Seed}
        \label{subfig:convergence_perseed_15}
    \end{subfigure}
    \caption{BBOB Function 15}
    \label{fig:bbob_function_15}
\end{figure}

\begin{figure}[h]
    \centering
    \begin{subfigure}[b]{0.45\textwidth}
        \centering
        \includegraphics[width=\textwidth]{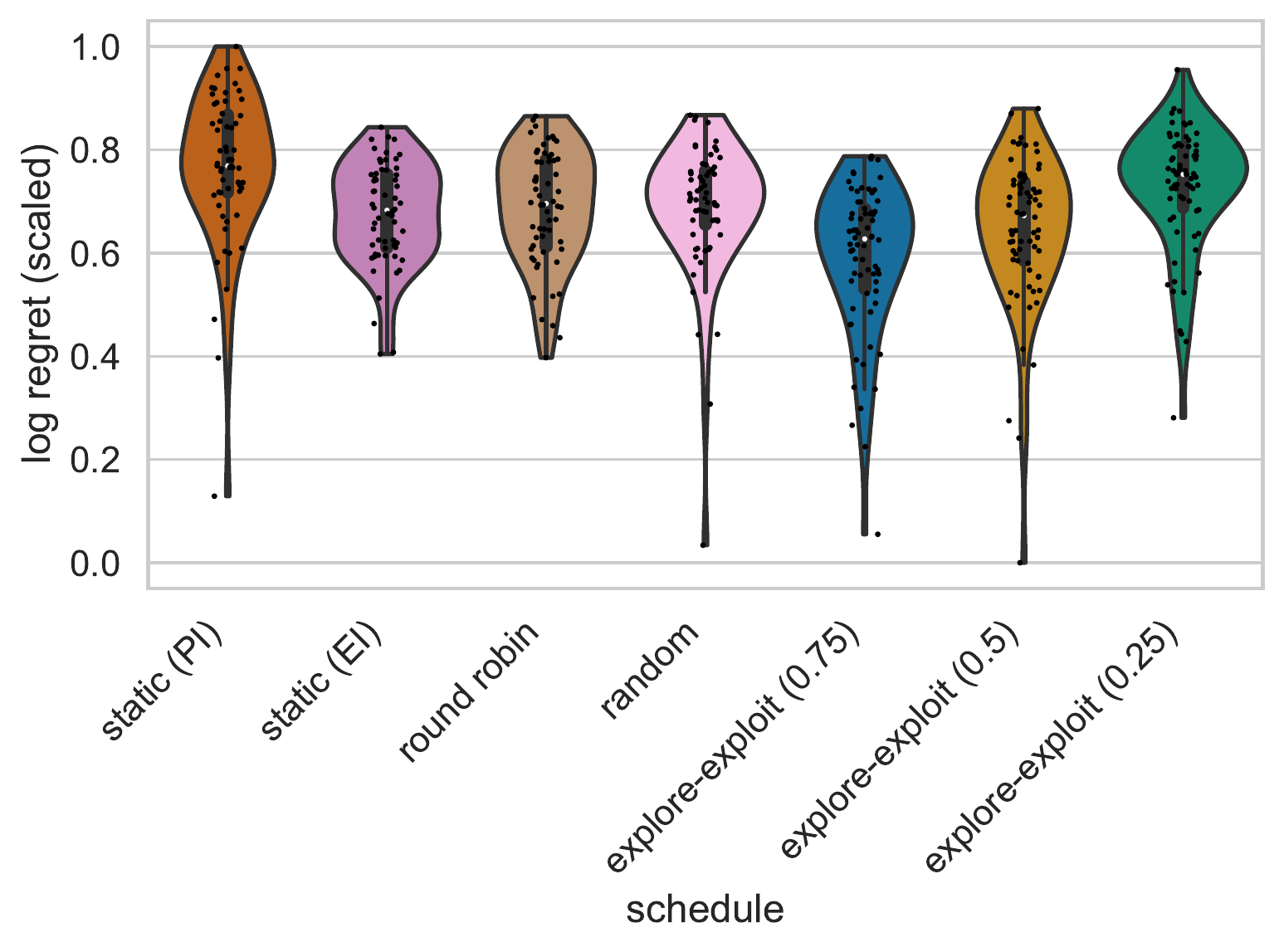}
        \caption{Final Log Regret (Scaled)}
        \label{subfig:boxplot_16}
    \end{subfigure}
    \hfill
    \begin{subfigure}[b]{0.45\textwidth}
        \centering
        \includegraphics[width=\textwidth]{figures/convergence/regret_over_steps_dimension_5-function_16.pdf}
        \caption{Log-Regret (Scaled) per Step}
        \label{subfig:convergence_16}
    \end{subfigure}\\
    \vspace*{3mm}
    \centering
    \begin{subfigure}[b]{\textwidth}
        \centering
        \includegraphics[width=\textwidth]{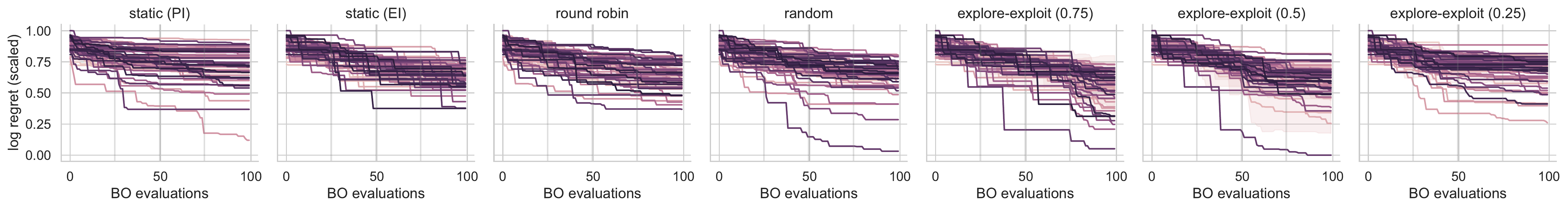}
        \caption{Log-Regret (Scaled) per Step and Seed}
        \label{subfig:convergence_perseed_16}
    \end{subfigure}
    \caption{BBOB Function 16}
    \label{fig:bbob_function_16}
\end{figure}

\begin{figure}[h]
    \centering
    \begin{subfigure}[b]{0.45\textwidth}
        \centering
        \includegraphics[width=\textwidth]{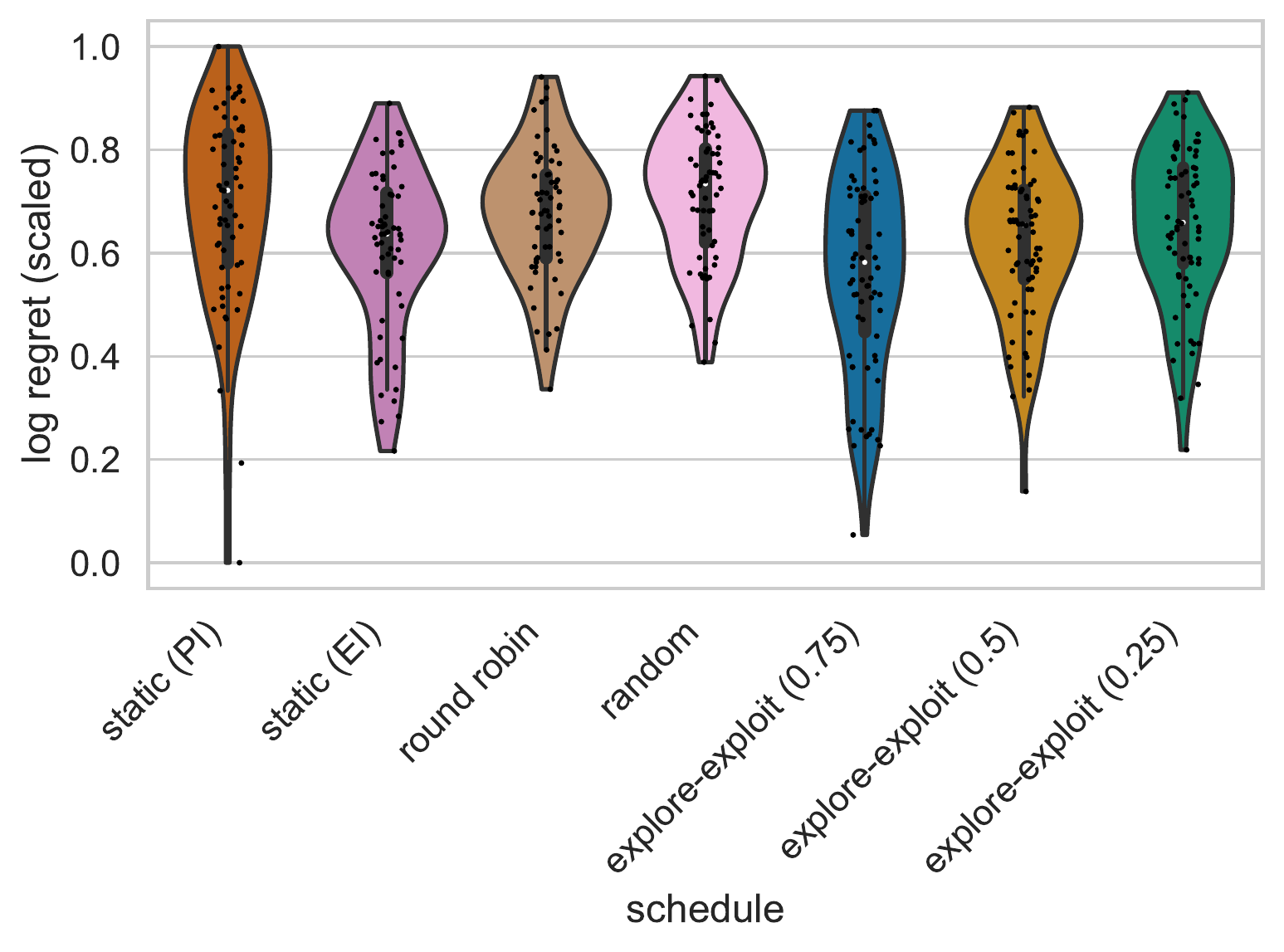}
        \caption{Final Log Regret (Scaled)}
        \label{subfig:boxplot_17}
    \end{subfigure}
    \hfill
    \begin{subfigure}[b]{0.45\textwidth}
        \centering
        \includegraphics[width=\textwidth]{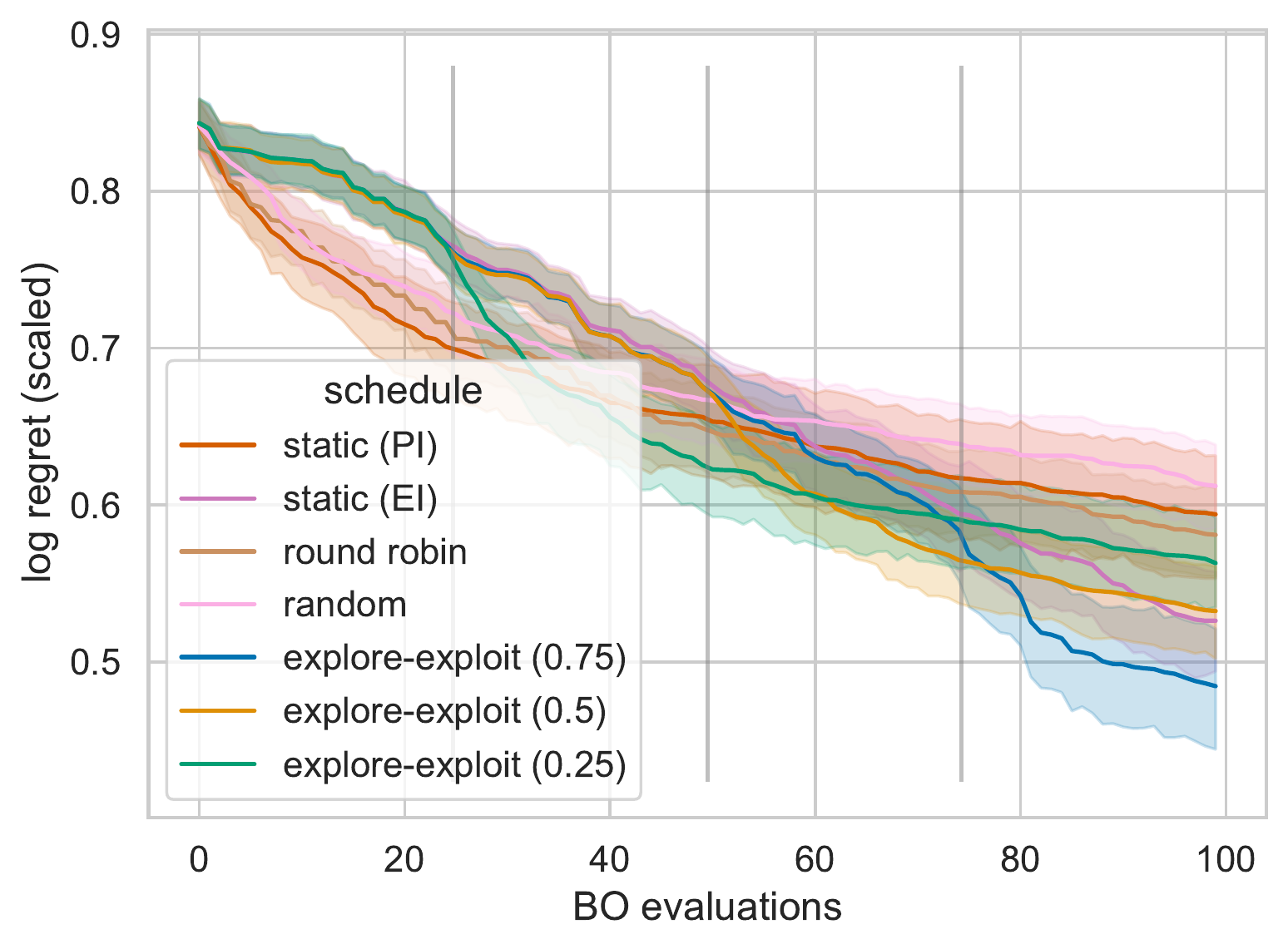}
        \caption{Log-Regret (Scaled) per Step}
        \label{subfig:convergence_17}
    \end{subfigure}\\
    \vspace*{3mm}
    \centering
    \begin{subfigure}[b]{\textwidth}
        \centering
        \includegraphics[width=\textwidth]{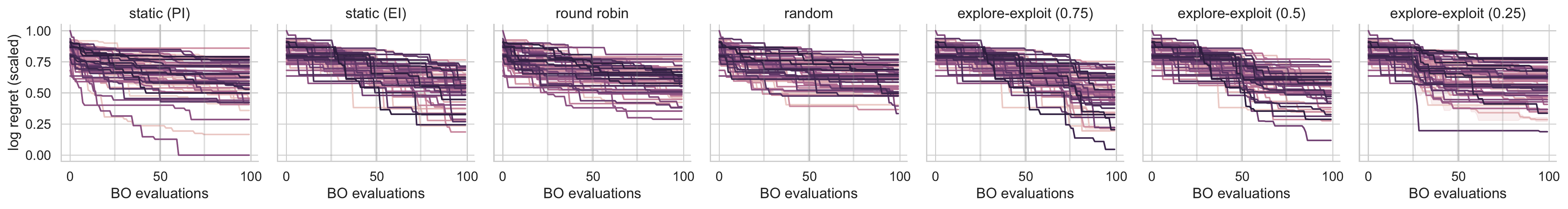}
        \caption{Log-Regret (Scaled) per Step and Seed}
        \label{subfig:convergence_perseed_17}
    \end{subfigure}
    \caption{BBOB Function 17}
    \label{fig:bbob_function_17}
\end{figure}

\begin{figure}[h]
    \centering
    \begin{subfigure}[b]{0.45\textwidth}
        \centering
        \includegraphics[width=\textwidth]{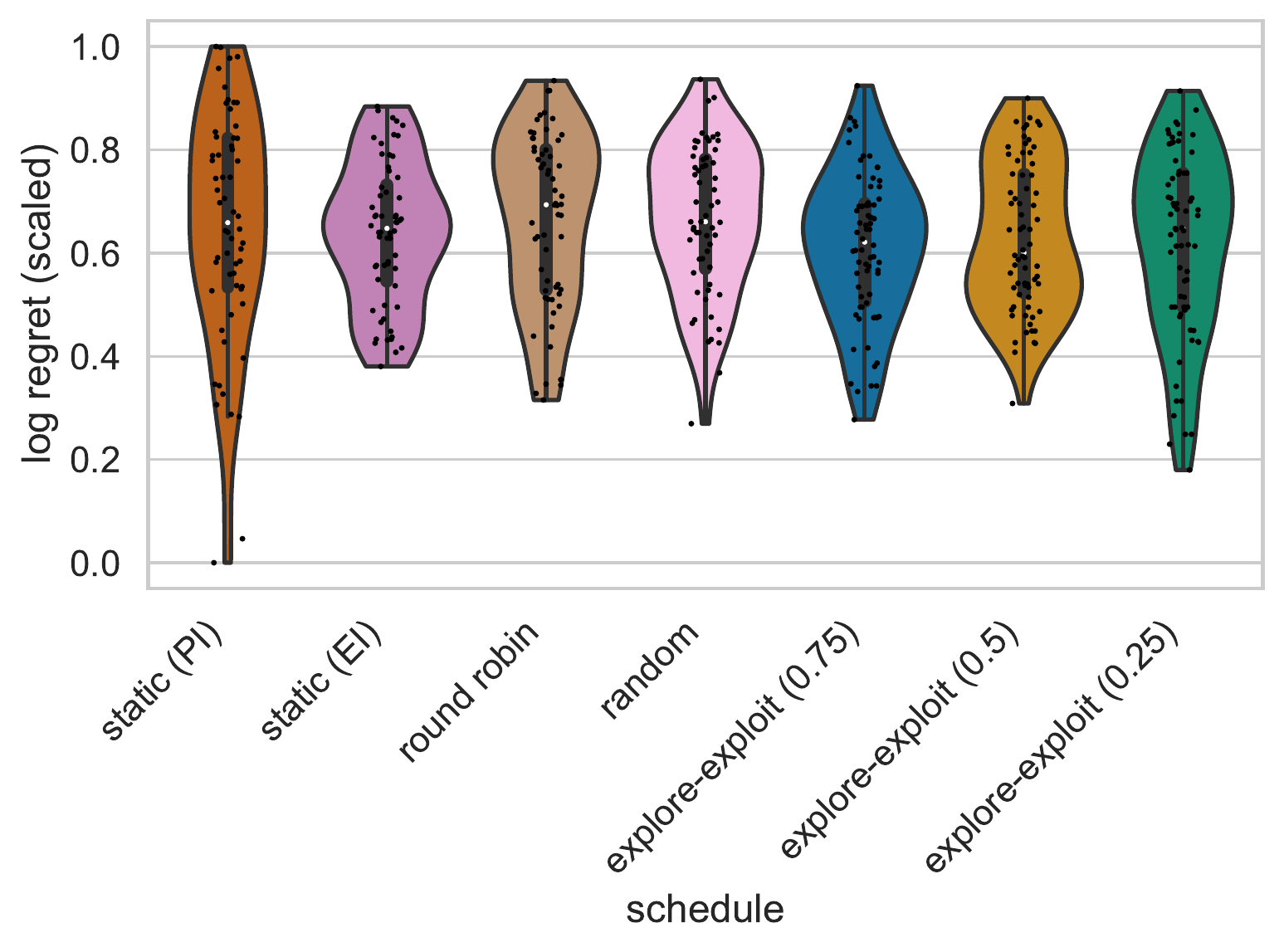}
        \caption{Final Log Regret (Scaled)}
        \label{subfig:boxplot_18}
    \end{subfigure}
    \hfill
    \begin{subfigure}[b]{0.45\textwidth}
        \centering
        \includegraphics[width=\textwidth]{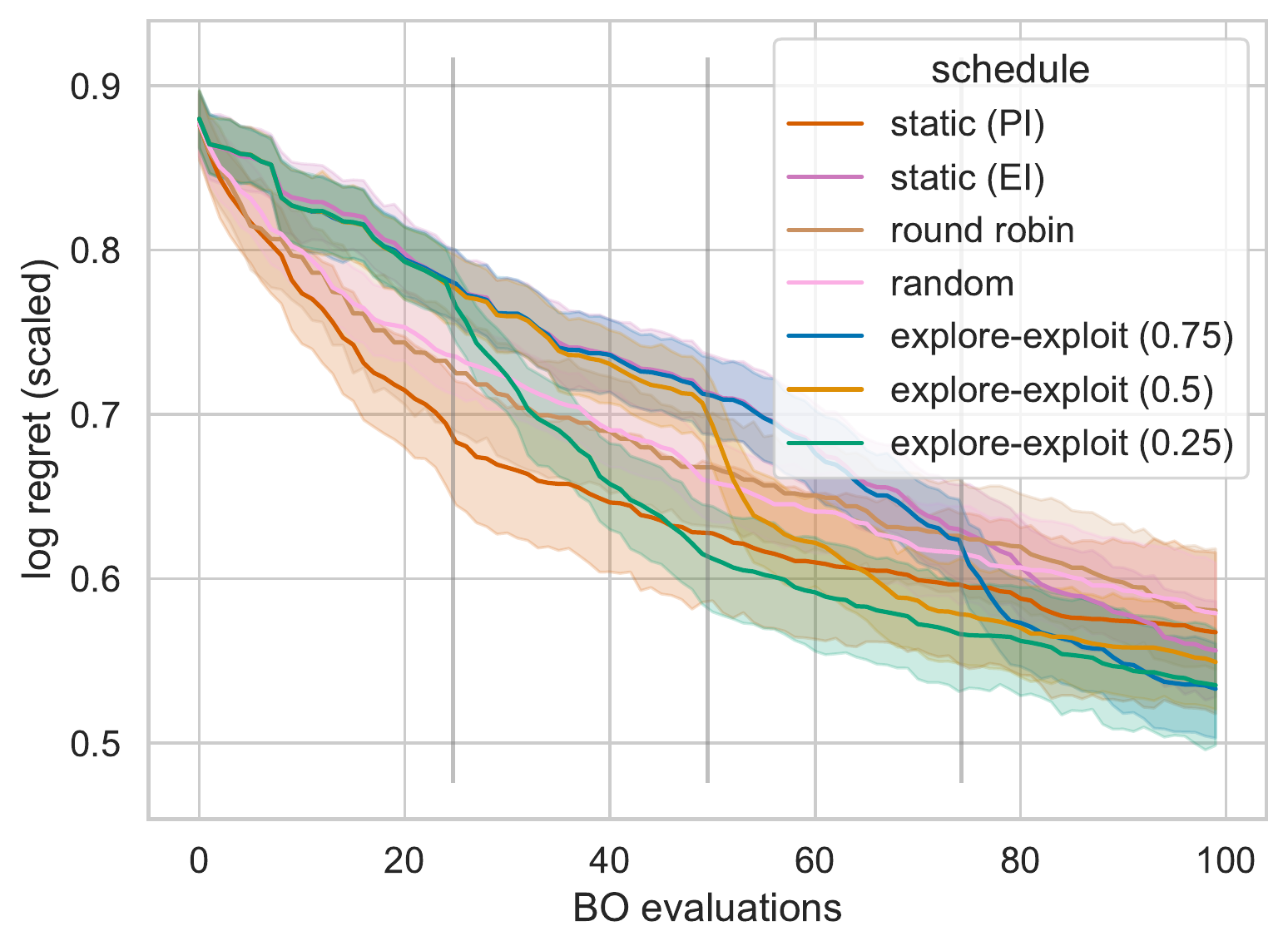}
        \caption{Log-Regret (Scaled) per Step}
        \label{subfig:convergence_18}
    \end{subfigure}\\
    \vspace*{3mm}
    \centering
    \begin{subfigure}[b]{\textwidth}
        \centering
        \includegraphics[width=\textwidth]{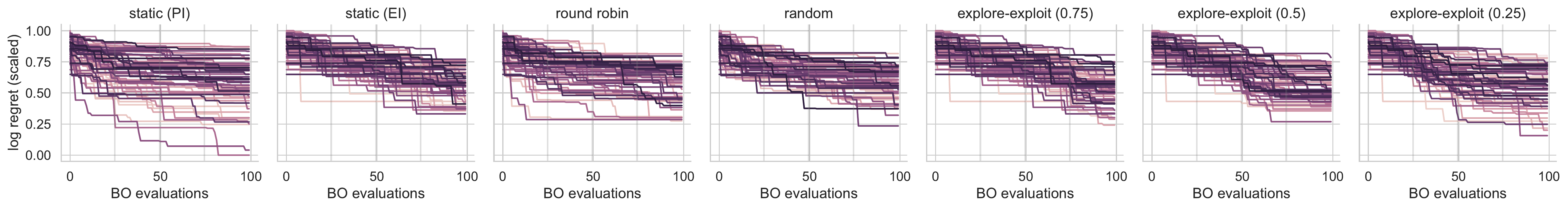}
        \caption{Log-Regret (Scaled) per Step and Seed}
        \label{subfig:convergence_perseed_18}
    \end{subfigure}
    \caption{BBOB Function 18}
    \label{fig:bbob_function_18}
\end{figure}

\begin{figure}[h]
    \centering
    \begin{subfigure}[b]{0.45\textwidth}
        \centering
        \includegraphics[width=\textwidth]{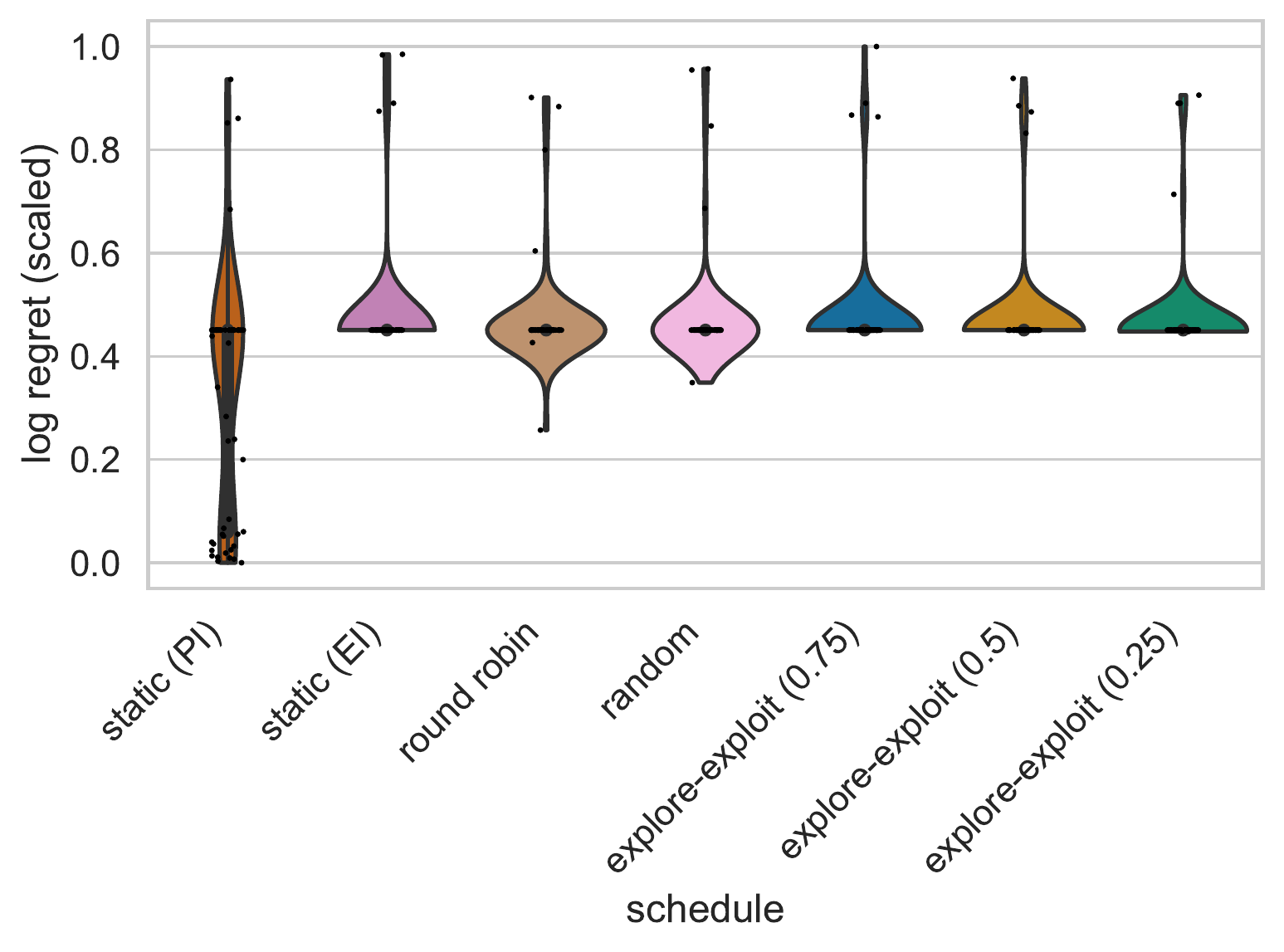}
        \caption{Final Log Regret (Scaled)}
        \label{subfig:boxplot_19}
    \end{subfigure}
    \hfill
    \begin{subfigure}[b]{0.45\textwidth}
        \centering
        \includegraphics[width=\textwidth]{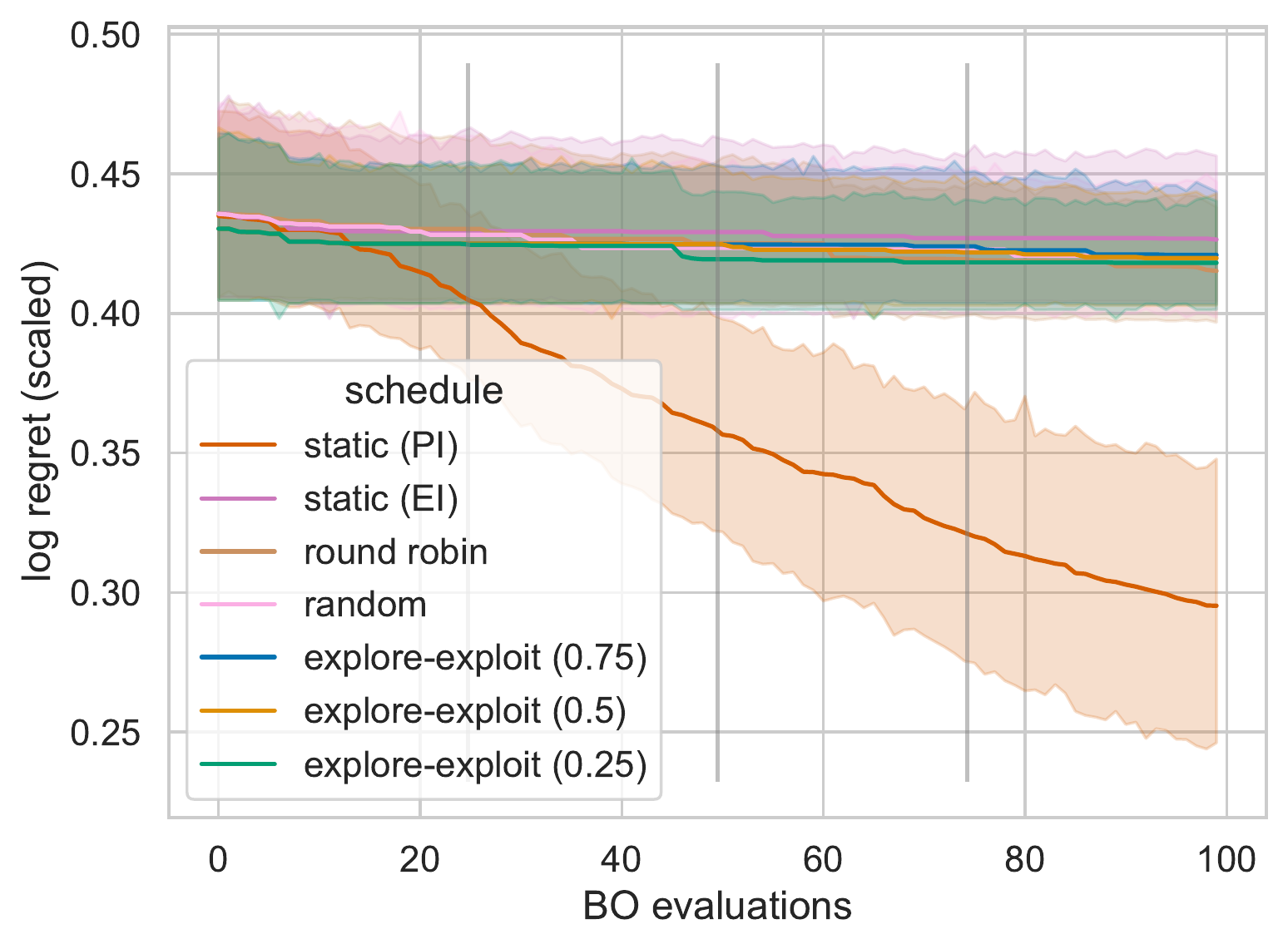}
        \caption{Log-Regret (Scaled) per Step}
        \label{subfig:convergence_19}
    \end{subfigure}\\
    \vspace*{3mm}
    \centering
    \begin{subfigure}[b]{\textwidth}
        \centering
        \includegraphics[width=\textwidth]{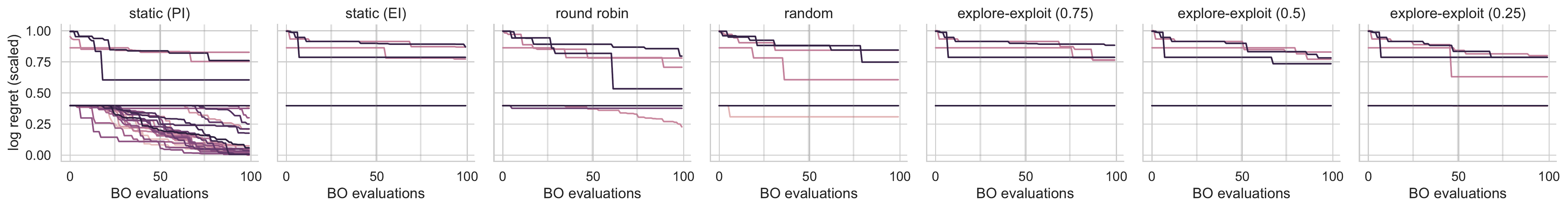}
        \caption{Log-Regret (Scaled) per Step and Seed}
        \label{subfig:convergence_perseed_19}
    \end{subfigure}
    \caption{BBOB Function 19}
    \label{fig:bbob_function_19}
\end{figure}

\begin{figure}[h]
    \centering
    \begin{subfigure}[b]{0.45\textwidth}
        \centering
        \includegraphics[width=\textwidth]{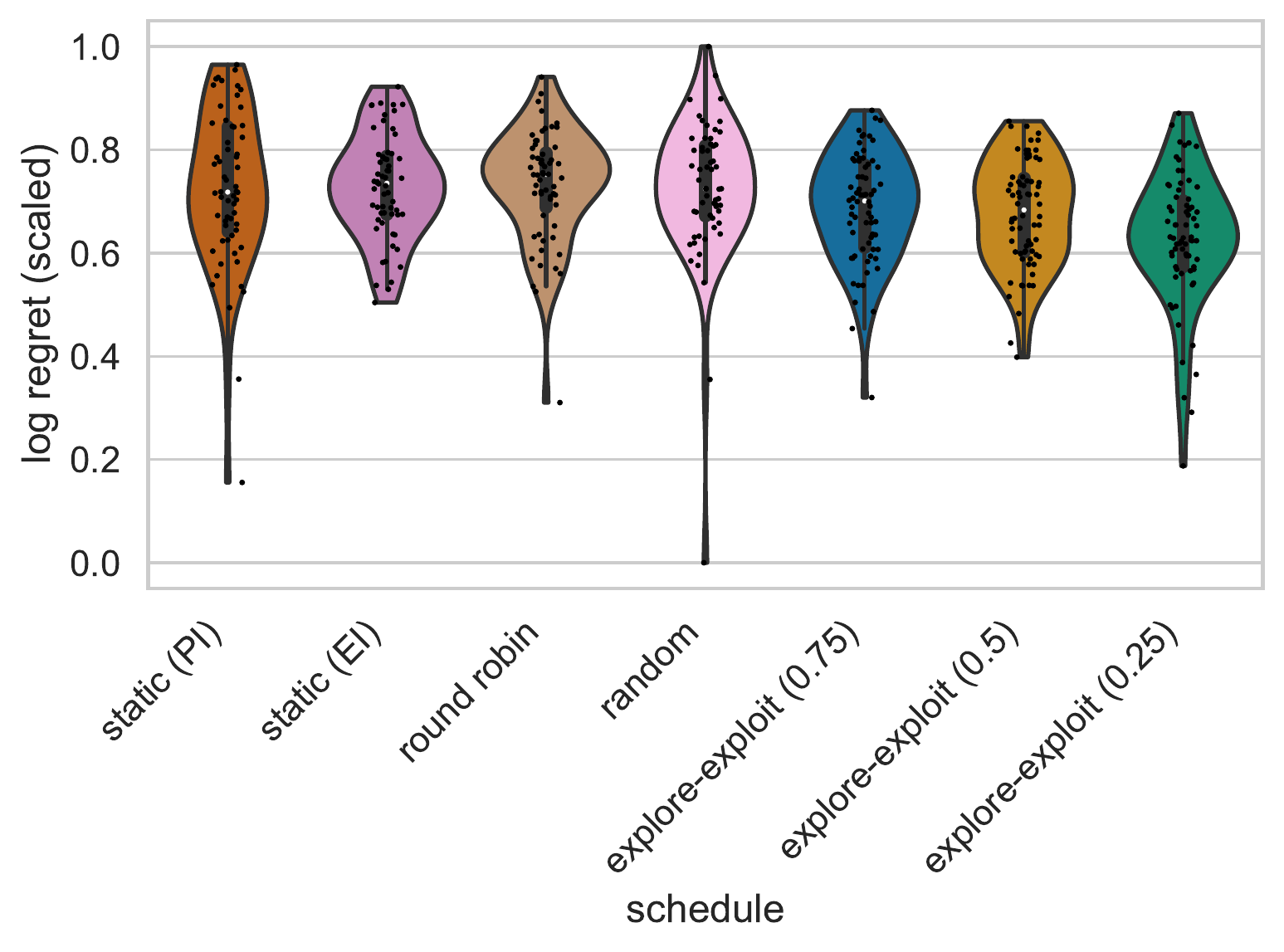}
        \caption{Final Log Regret (Scaled)}
        \label{subfig:boxplot_20}
    \end{subfigure}
    \hfill
    \begin{subfigure}[b]{0.45\textwidth}
        \centering
        \includegraphics[width=\textwidth]{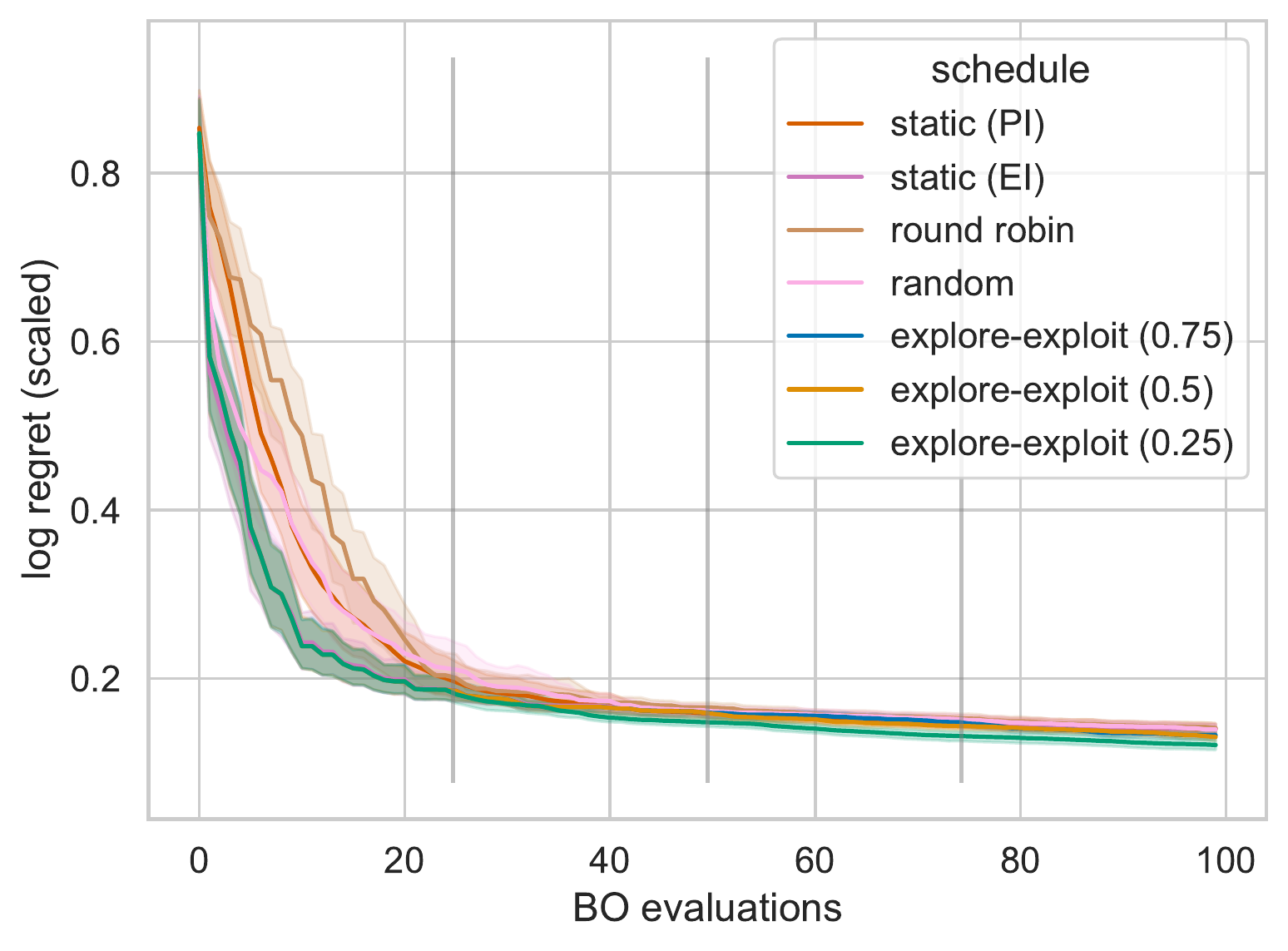}
        \caption{Log-Regret (Scaled) per Step}
        \label{subfig:convergence_20}
    \end{subfigure}\\
    \vspace*{3mm}
    \centering
    \begin{subfigure}[b]{\textwidth}
        \centering
        \includegraphics[width=\textwidth]{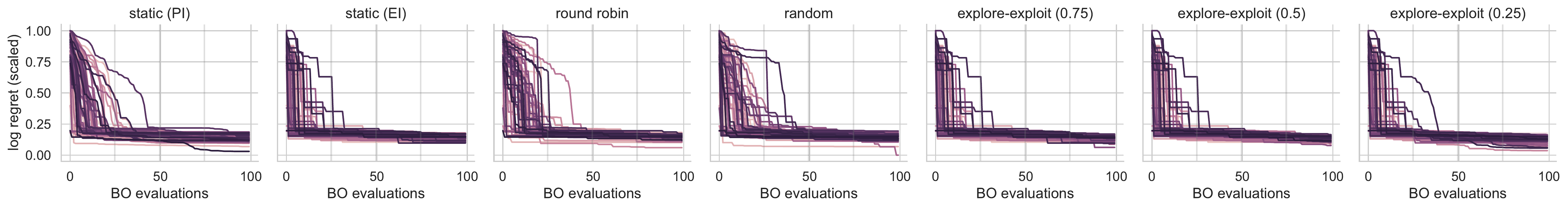}
        \caption{Log-Regret (Scaled) per Step and Seed}
        \label{subfig:convergence_perseed_20}
    \end{subfigure}
    \caption{BBOB Function 20}
    \label{fig:bbob_function_20}
\end{figure}

\begin{figure}[h]
    \centering
    \begin{subfigure}[b]{0.45\textwidth}
        \centering
        \includegraphics[width=\textwidth]{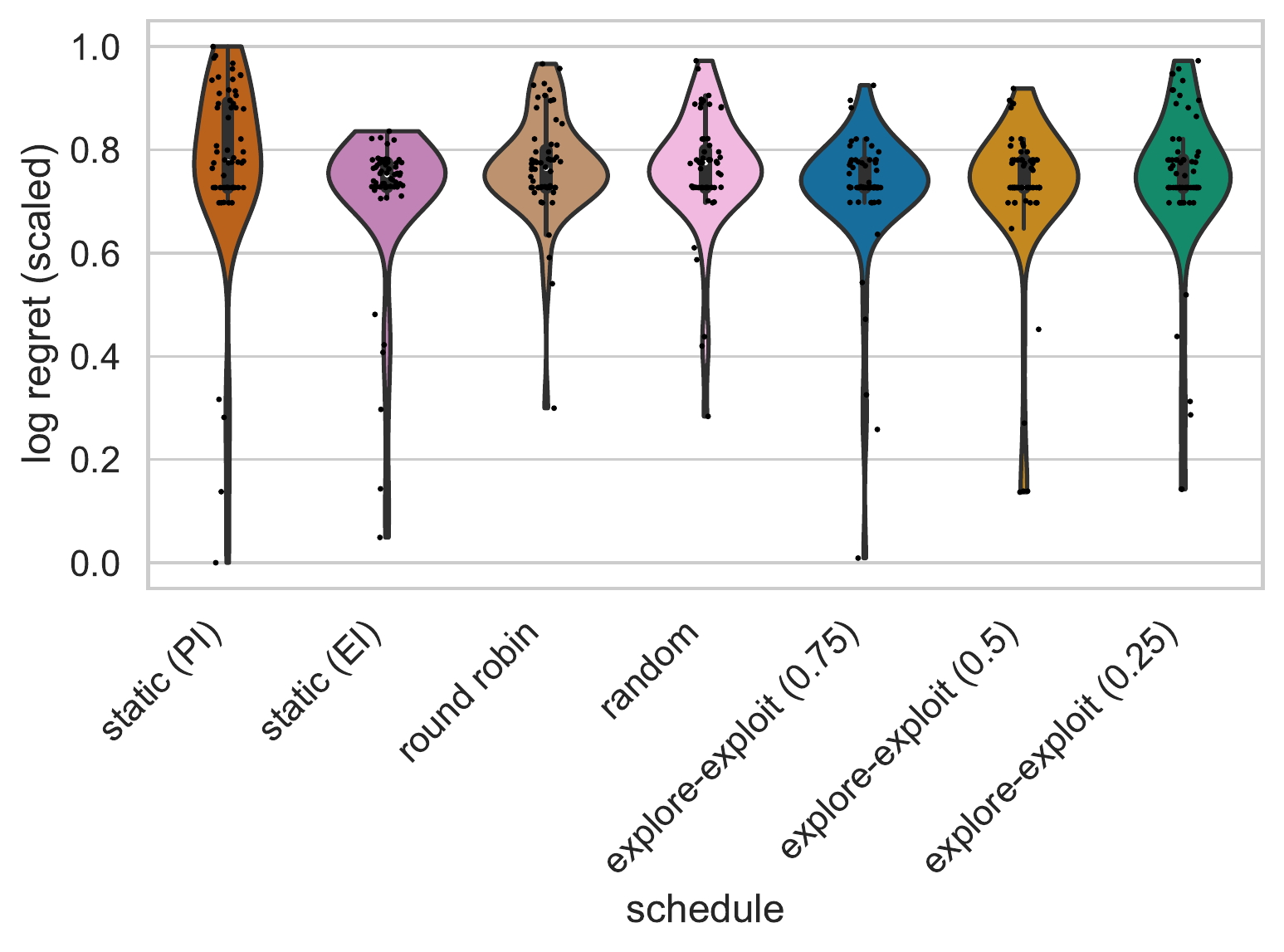}
        \caption{Final Log Regret (Scaled)}
        \label{subfig:boxplot_21}
    \end{subfigure}
    \hfill
    \begin{subfigure}[b]{0.45\textwidth}
        \centering
        \includegraphics[width=\textwidth]{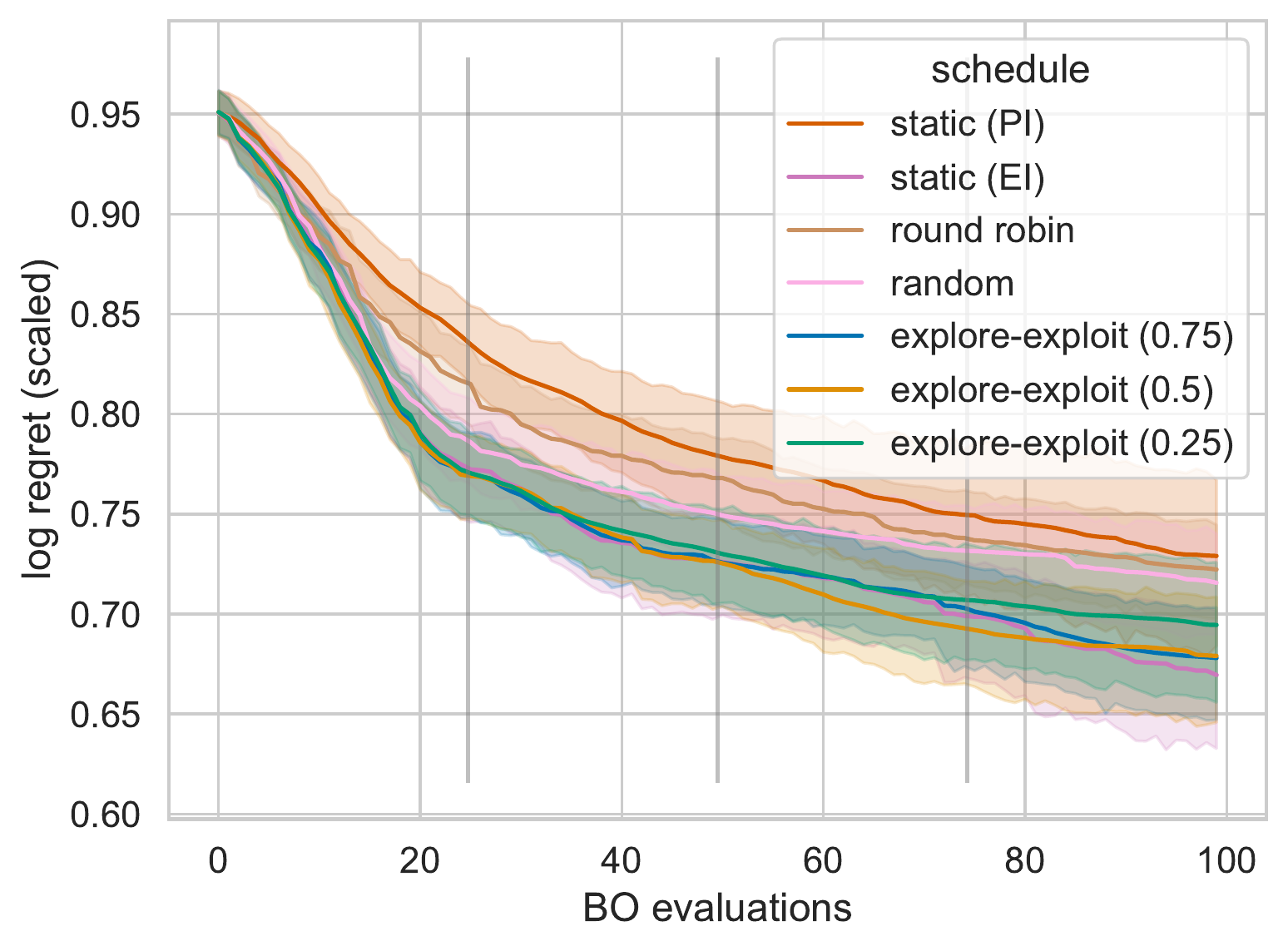}
        \caption{Log-Regret (Scaled) per Step}
        \label{subfig:convergence_21}
    \end{subfigure}\\
    \vspace*{3mm}
    \centering
    \begin{subfigure}[b]{\textwidth}
        \centering
        \includegraphics[width=\textwidth]{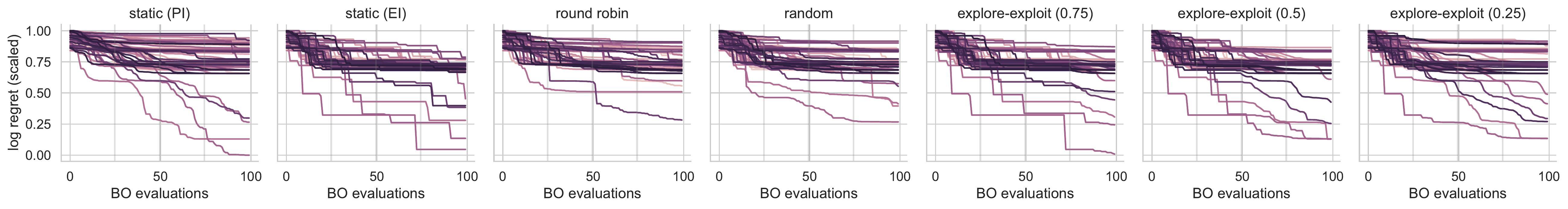}
        \caption{Log-Regret (Scaled) per Step and Seed}
        \label{subfig:convergence_perseed_21}
    \end{subfigure}
    \caption{BBOB Function 21}
    \label{fig:bbob_function_21}
\end{figure}

\begin{figure}[h]
    \centering
    \begin{subfigure}[b]{0.45\textwidth}
        \centering
        \includegraphics[width=\textwidth]{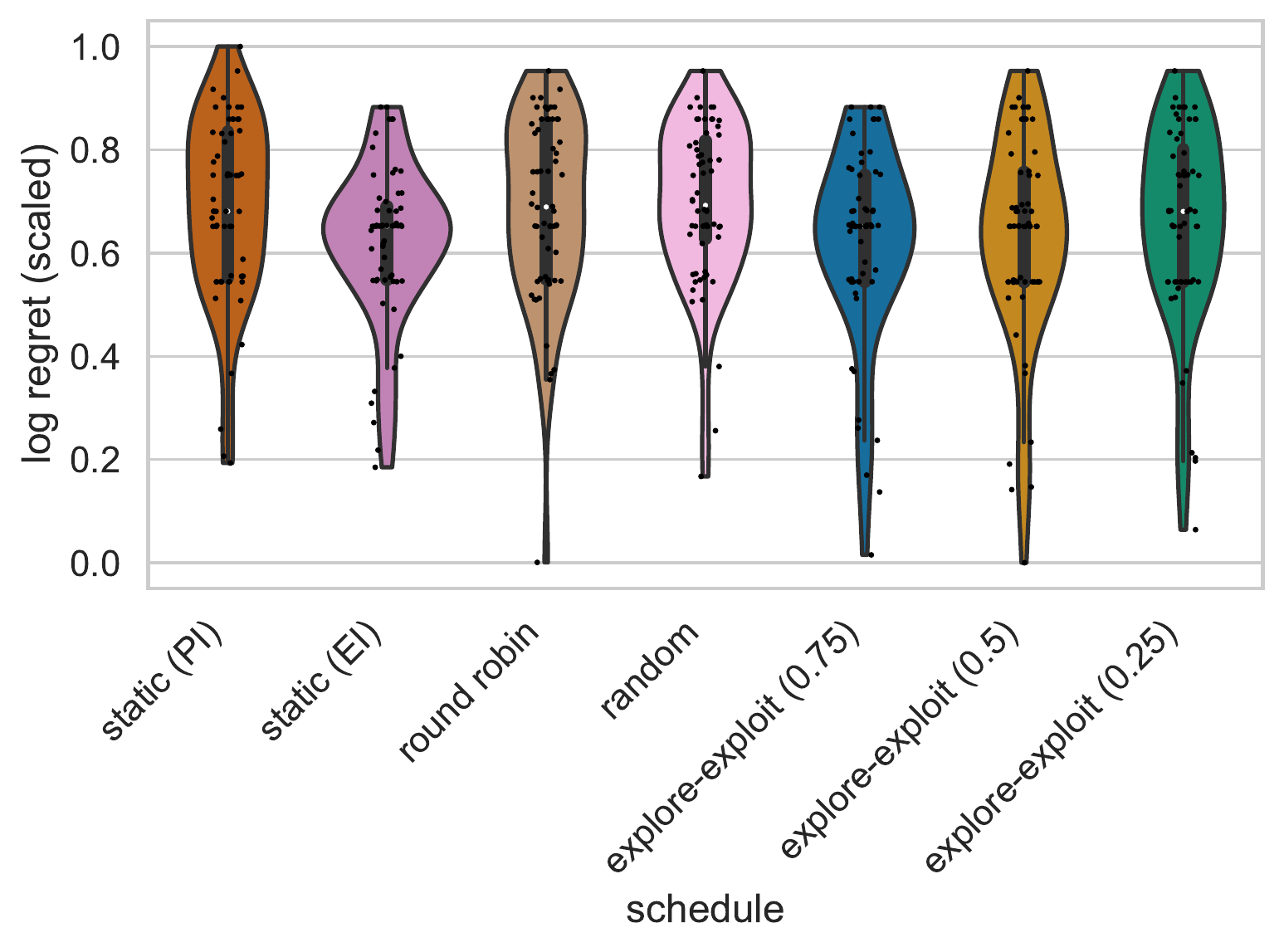}
        \caption{Final Log Regret (Scaled)}
        \label{subfig:boxplot_22}
    \end{subfigure}
    \hfill
    \begin{subfigure}[b]{0.45\textwidth}
        \centering
        \includegraphics[width=\textwidth]{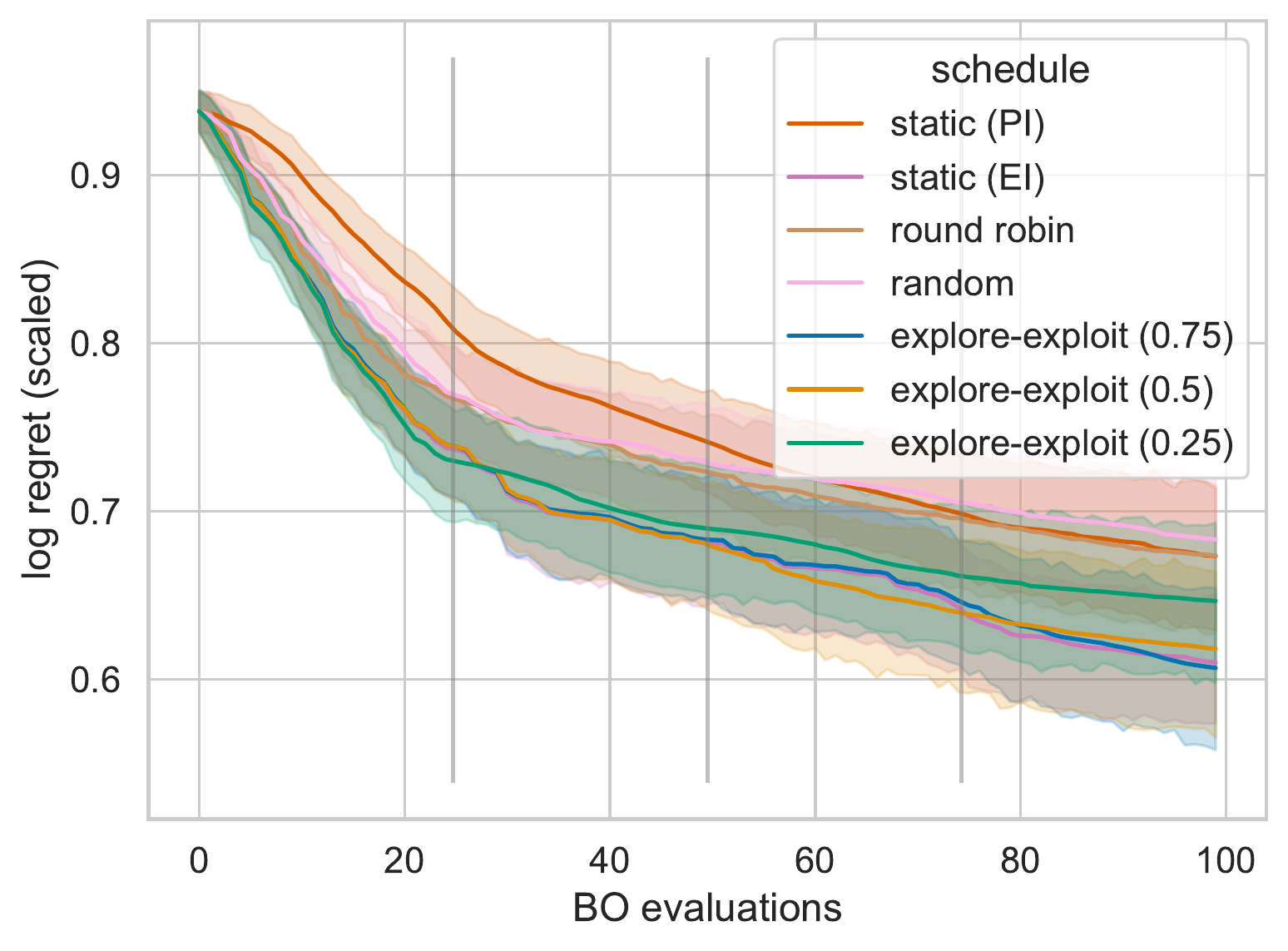}
        \caption{Log-Regret (Scaled) per Step}
        \label{subfig:convergence_22}
    \end{subfigure}\\
    \vspace*{3mm}
    \centering
    \begin{subfigure}[b]{\textwidth}
        \centering
        \includegraphics[width=\textwidth]{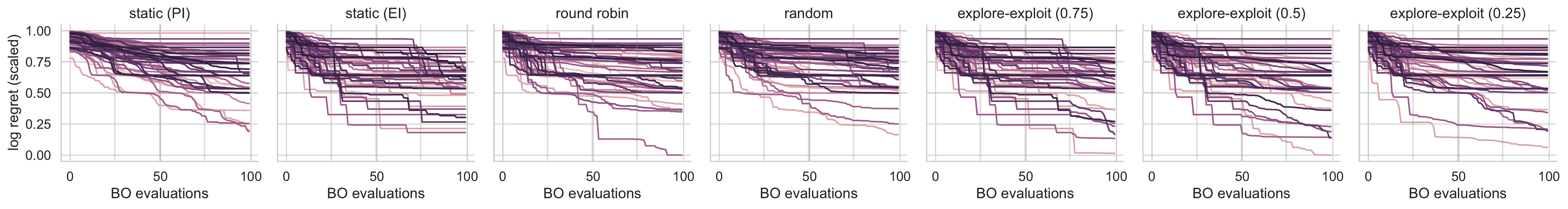}
        \caption{Log-Regret (Scaled) per Step and Seed}
        \label{subfig:convergence_perseed_22}
    \end{subfigure}
    \caption{BBOB Function 22}
    \label{fig:bbob_function_22}
\end{figure}

\begin{figure}[h]
    \centering
    \begin{subfigure}[b]{0.45\textwidth}
        \centering
        \includegraphics[width=\textwidth]{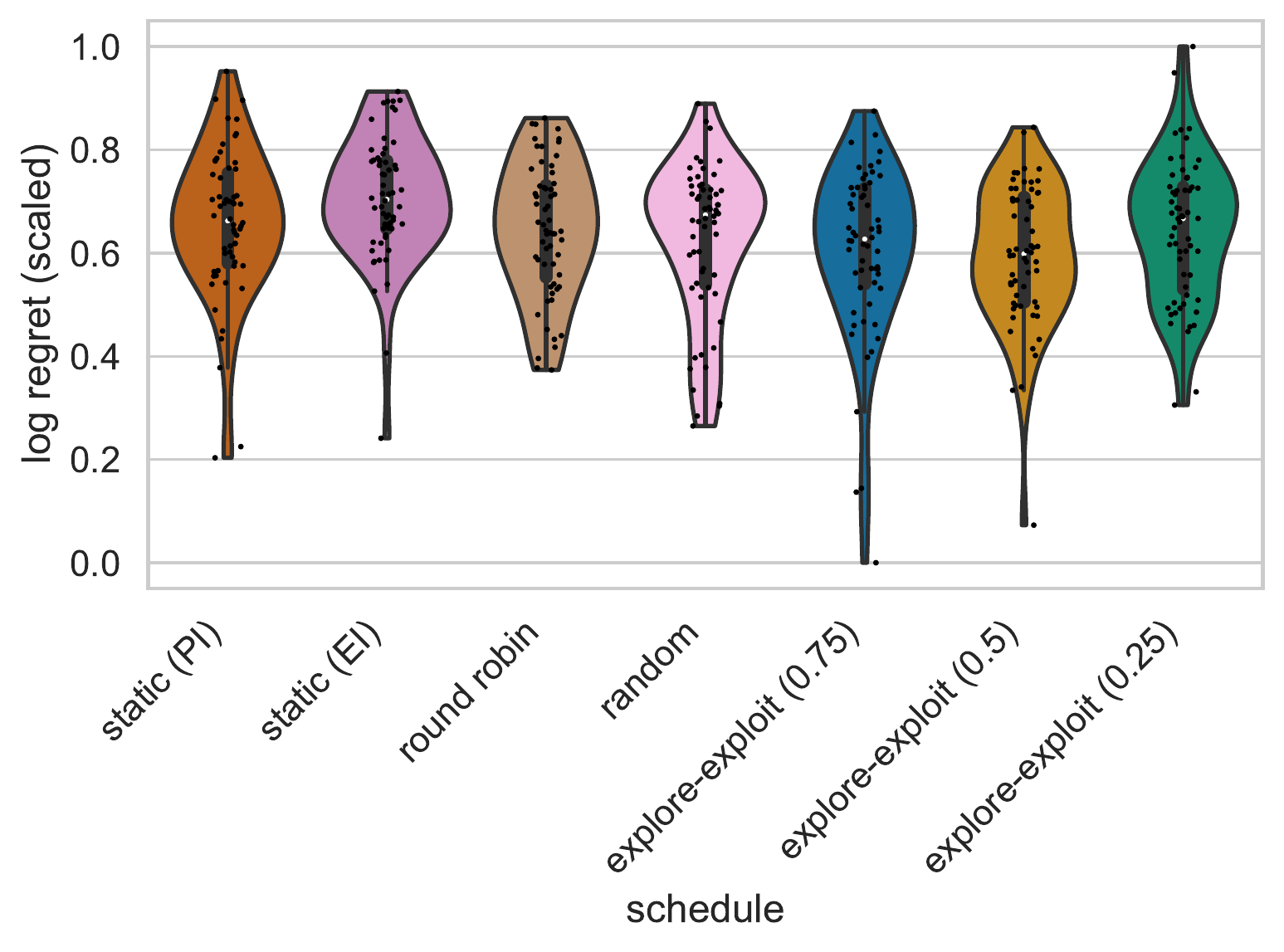}
        \caption{Final Log Regret (Scaled)}
        \label{subfig:boxplot_23}
    \end{subfigure}
    \hfill
    \begin{subfigure}[b]{0.45\textwidth}
        \centering
        \includegraphics[width=\textwidth]{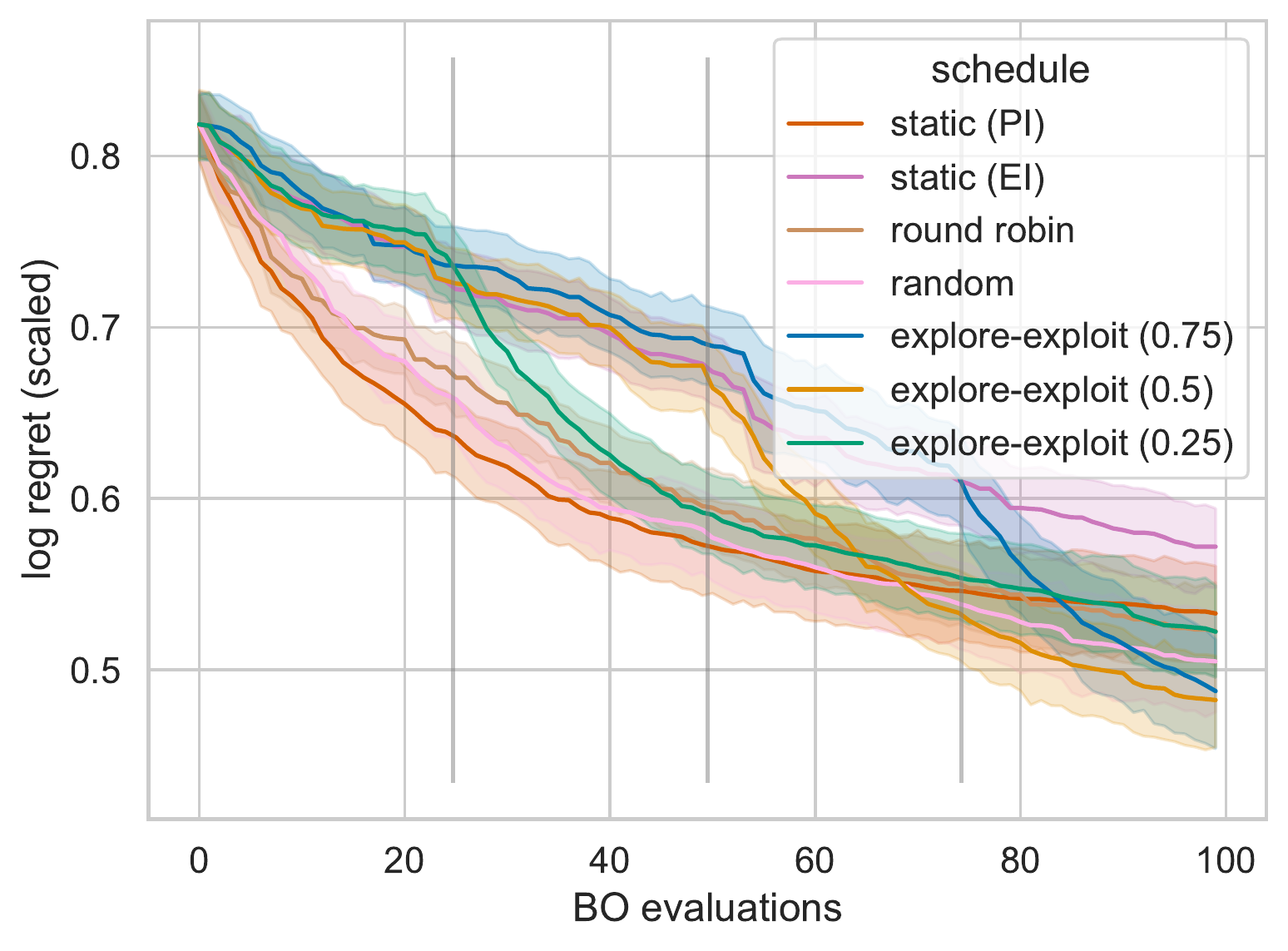}
        \caption{Log-Regret (Scaled) per Step}
        \label{subfig:convergence_23}
    \end{subfigure}\\
    \vspace*{3mm}
    \centering
    \begin{subfigure}[b]{\textwidth}
        \centering
        \includegraphics[width=\textwidth]{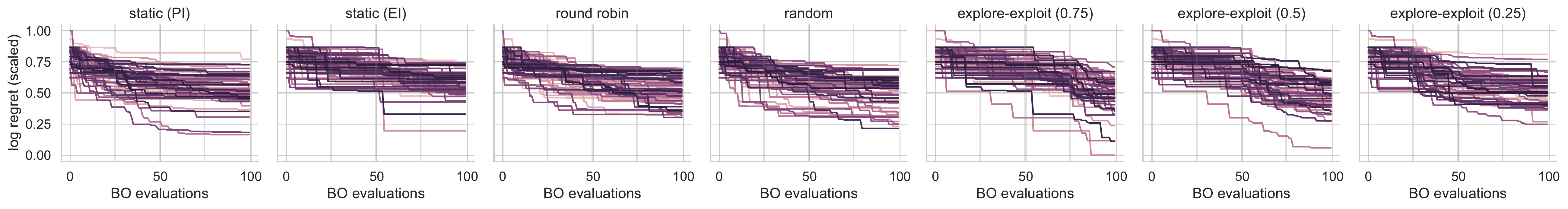}
        \caption{Log-Regret (Scaled) per Step and Seed}
        \label{subfig:convergence_perseed_23}
    \end{subfigure}
    \caption{BBOB Function 23}
    \label{fig:bbob_function_23}
\end{figure}

\begin{figure}[h]
    \centering
    \begin{subfigure}[b]{0.45\textwidth}
        \centering
        \includegraphics[width=\textwidth]{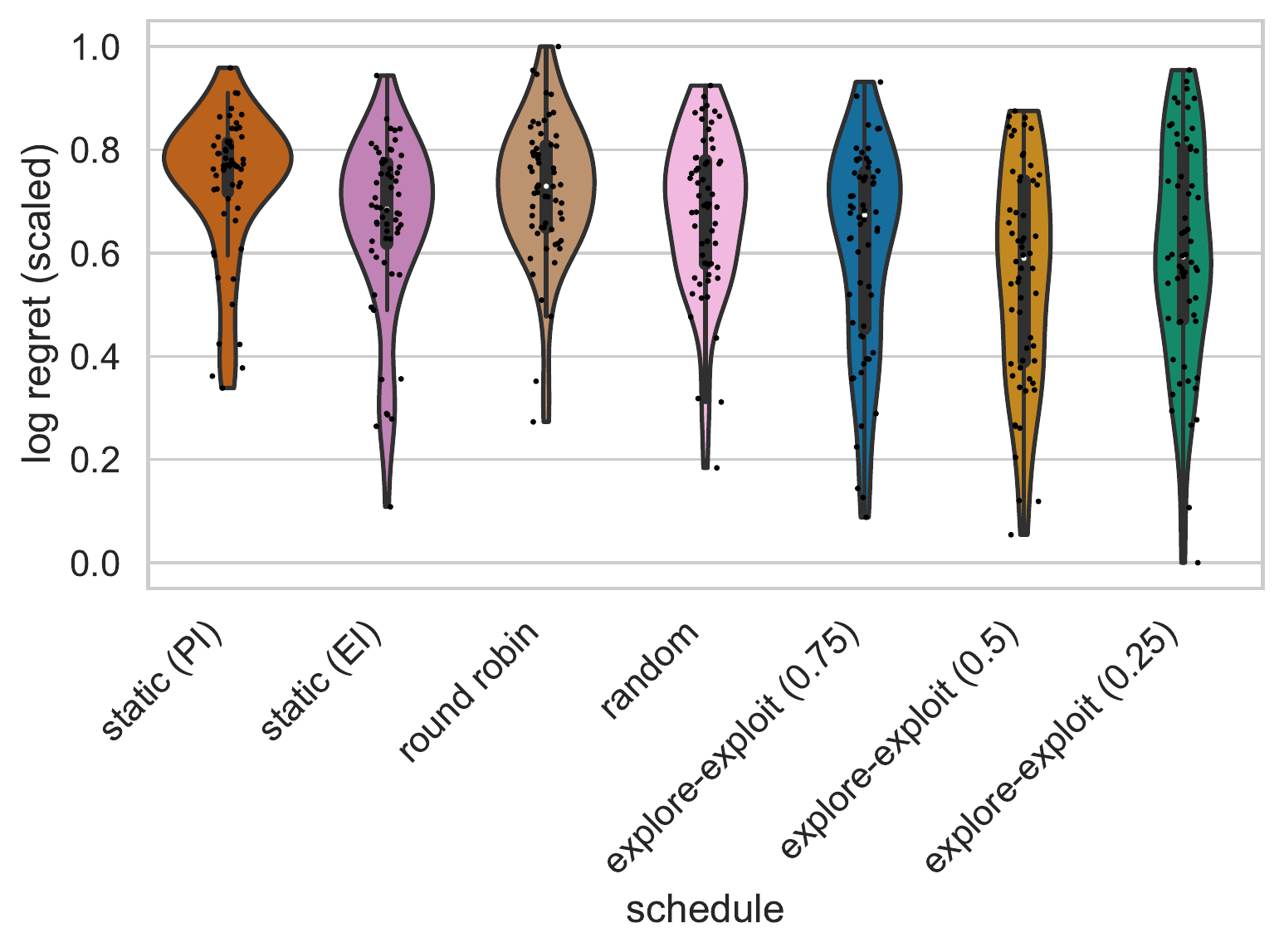}
        \caption{Final Log Regret (Scaled)}
        \label{subfig:boxplot_24}
    \end{subfigure}
    \hfill
    \begin{subfigure}[b]{0.45\textwidth}
        \centering
        \includegraphics[width=\textwidth]{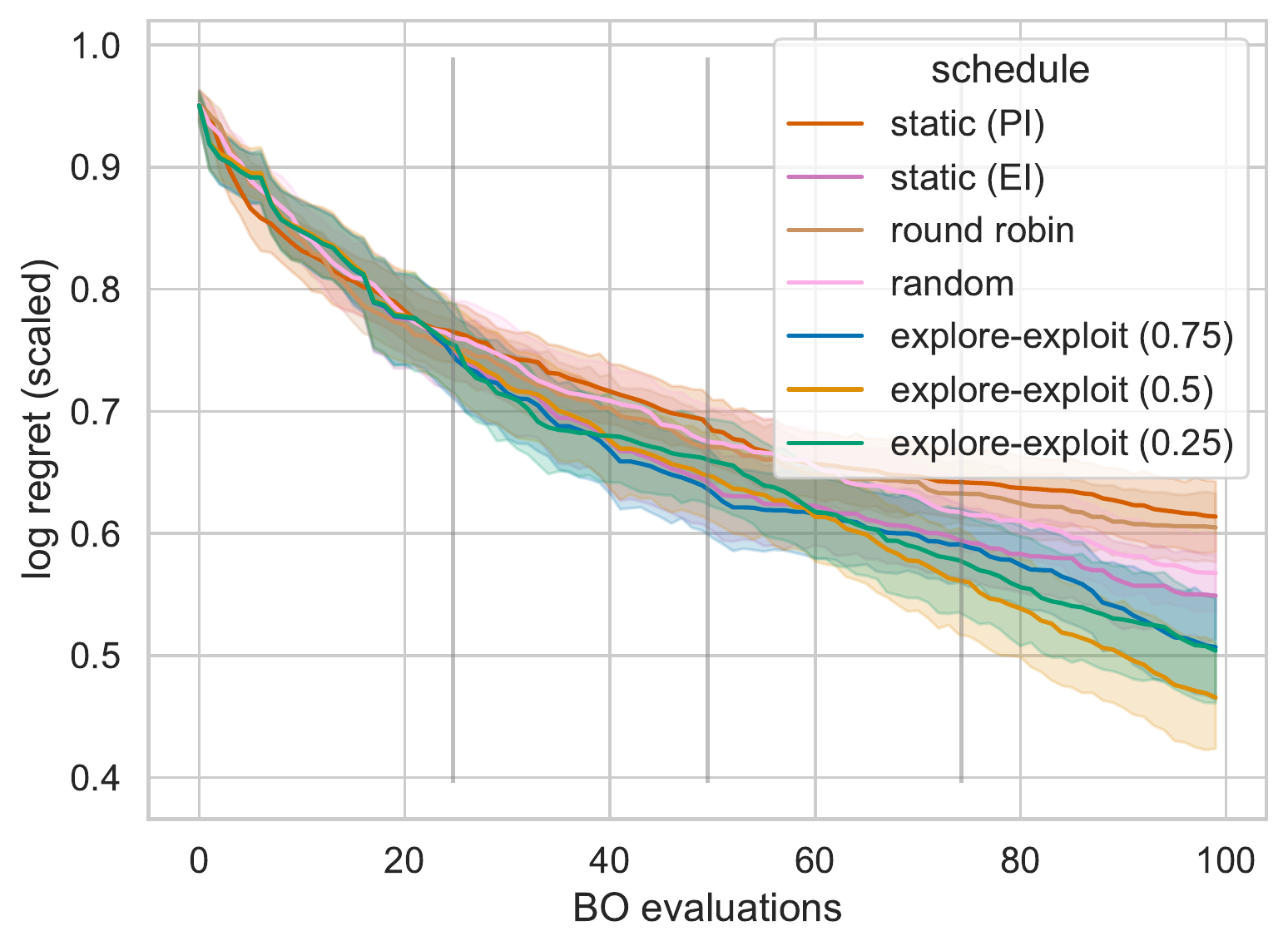}
        \caption{Log-Regret (Scaled) per Step}
        \label{subfig:convergence_24}
    \end{subfigure}\\
    \vspace*{3mm}
    \centering
    \begin{subfigure}[b]{\textwidth}
        \centering
        \includegraphics[width=\textwidth]{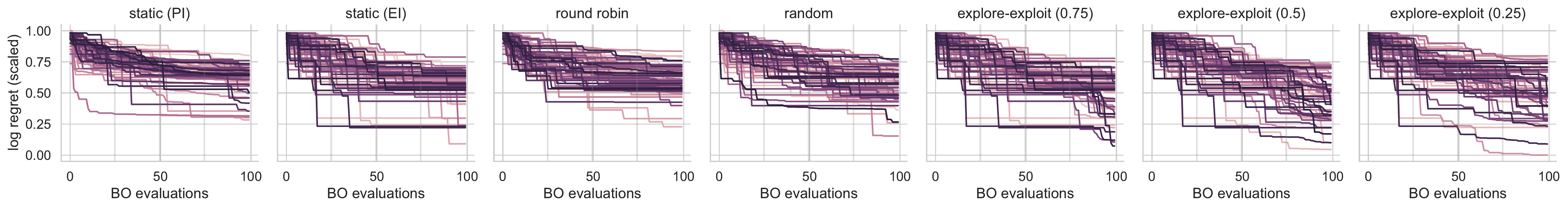}
        \caption{Log-Regret (Scaled) per Step and Seed}
        \label{subfig:convergence_perseed_24}
    \end{subfigure}
    \caption{BBOB Function 24}
    \label{fig:bbob_function_24}
\end{figure}

\end{document}